\documentclass[lettersize,journal]{IEEEtran}
\usepackage{amsmath,amsfonts}
\usepackage{algorithmic}
\usepackage{array}
\usepackage[caption=false,font=normalsize,labelfont=sf,textfont=sf]{subfig}
\usepackage{textcomp}
\usepackage{url}
\usepackage{bigstrut,multirow}
\usepackage{verbatim}
\usepackage{graphicx}
\usepackage{amsmath,amsfonts}
\usepackage{algorithmic}
\usepackage{algorithm}
\usepackage{array}
\usepackage{verbatim}
\usepackage{caption}
\usepackage{import}
\usepackage{colortbl}
\usepackage{booktabs}
\usepackage{xcolor}
\usepackage{orcidlink} 
\usepackage{balance}
\usepackage{float}
\usepackage{dblfloatfix}
\usepackage{longtable}  
\def\BibTeX{{\rm B\kern-.05em{\sc i\kern-.025em b}\kern-.08em
    T\kern-.1667em\lower.7ex\hbox{E}\kern-.125emX}}

\begin{document}
\title{MFVLR: Multi-domain Fine-grained Vision-Language Reconstruction for Generalizable Diffusion Face Forgery Detection and Localization}
\author{Yaning Zhang \orcidlink{0000-0001-8442-2777}, Tianyi Wang \orcidlink{0000-0003-2920-6099}, \emph{Member, IEEE}, Zan Gao \orcidlink{0000-0003-2182-5741}, \emph{Senior Member, IEEE}, Yibo Zhao \orcidlink{0000-0003-4187-1980}, Chunjie Ma \orcidlink{0000-0002-6348-671X}, and Meng Wang \orcidlink{0000-0002-3094-7735}, \emph{Fellow, IEEE}
	\thanks{This work was supported in part by the National Natural Science Foundation of China (No.U25A20444, No.62372325, No.62402255, No.62502344), Natural Science Foundation of Tianjin Municipality (No.23JCZDJC00280), Shandong Provincial Natural Science Foundation (No.ZR2024QF020), Shandong Province National Talents Supporting Program (No.2023GJJLJRC-070), Young Talent of Lifting engineering for Science and Technology in Shandong (No.SDAST2024QTB001), Shandong Project towards the Integration of Education and Industry (No.2024ZDZX11), the Open Project Program of State Key Laboratory of Virtual Reality Technology and Systems, Beihang University (No.VRLAB2025C05). (Corresponding author: Zan Gao)}
	
	\thanks{Y. Zhang is with the Faculty of Computer Science and Technology, Qilu University of Technology (Shandong Academy of Sciences), Jinan, 250353, China. E-mail: zhangyaning0321@163.com}
	
	\thanks{T. Wang is with the School of Computing, National University of Singapore, 21 Lower Kent Ridge Rd, 11907750, Singapore. E-mail: terry.ai.wang@gmail.com}
	
	\thanks{Z. Gao is with the Shandong Artificial Intelligence Institute, Qilu University of Technology (Shandong Academy of Sciences), Jinan, 250014, China, and also with the Key Laboratory of Computer Vision and System, Ministry of Education, Tianjin University of Technology, Tianjin, 300384, China. E-mail: zangaonsh4522@gmail.com } 
		\thanks{Y. Zhao is with the Key Laboratory of Computer Vision and Systems, Ministry of Education, Tianjin University of Technology, Tianjin, 300384, China. E-mail: zybtjut@163.com } 
		\thanks{C. Ma is with the Shandong Artificial Intelligence Institute, Qilu University of
			Technology (Shandong Academy of Sciences), Jinan, 250014, China. E-mail: mcj@machunjie.com }
	\thanks{M. Wang is with the School of Computer Science and Information
		Engineering, Hefei University of Technology, Hefei, 300384, China. E-mail:
		eric.mengwang@gmail.com} 
}
\markboth{Journal of \LaTeX\ Class Files,~Vol.~18, No.~9, September~2020}%
{How to Use the IEEEtran \LaTeX \ Templates}

\maketitle

\begin{abstract}
The swift advancement in photo-realistic face generation technology has sparked considerable concerns across society and academia, emphasizing the requirement of generalizable face forgery detection and localization methods. Prior works tend to capture face forgery patterns across multiple domains using image modality, other modalities like fine-grained texts are not comprehensively investigated, which restricts the generalization capability of models. Besides, they usually analyze facial images created by GAN, but struggle to identify and localize those synthesized by diffusion. To solve the problems, in this paper, we devise a novel multi-domain fine-grained vision-language reconstruction (MFVLR) model, which explores comprehensive and diverse visual forgery traces via language-guided face forgery representation learning, to achieve generalizable diffusion-synthesized face forgery detection and localization (DFFDL). Specifically, we devise a fine-grained language transformer that studies general fine-grained language embeddings using language reconstruction. We propose a multi-domain vision encoder to capture general and complementary visual forgery patterns across the image and residual domains. A vision decoder is designed to reconstruct image appearance and achieve forgery localization. Besides, we propose an innovative plug-and-play vision injection module to enhance the interaction between the vision and language embeddings. Extensive experiments and visualizations demonstrate that our network outperforms the state of the art on different settings like cross-generator, cross-forgery, and cross-dataset evaluations.
\end{abstract}

\vspace{-2em}

\section{Introduction}

Face forgery detection and localization (FFDL) technologies aim to identify and locate digitally manipulated facial areas using deep learning-based methods such as convolutional neural networks (CNN) \cite{Xception,resnet} and transformers \cite{transformer,vitsur,DTN,TransDFD}. Specifically, face forgery detection refers to determining the authenticity of facial images at either the image level or the video level, and localization intends to achieve pixel-level predictions for facial images further. Photorealistic generative approaches, including generative adversarial networks (GANs)  \cite{NIPS2014_5ca3e9b1, StyleGAN3, StyleGAN2} and denoising diffusion probabilistic models (DDPM) \cite{ddpm}, have attained extraordinary advancement in generating extremely lifelike facial images, such that human fails to distinguish their reliability \cite{DeepfakeSurvey}. With the increasing prevalence of social media as well as digital content across the internet, advanced FFDL becomes a critical and imperative demand, which contributes to enhancing the public credibility of digital information and preventing the spread of misinformation.

\begin{figure}[t]
	\centering
	\includegraphics[width=\linewidth]{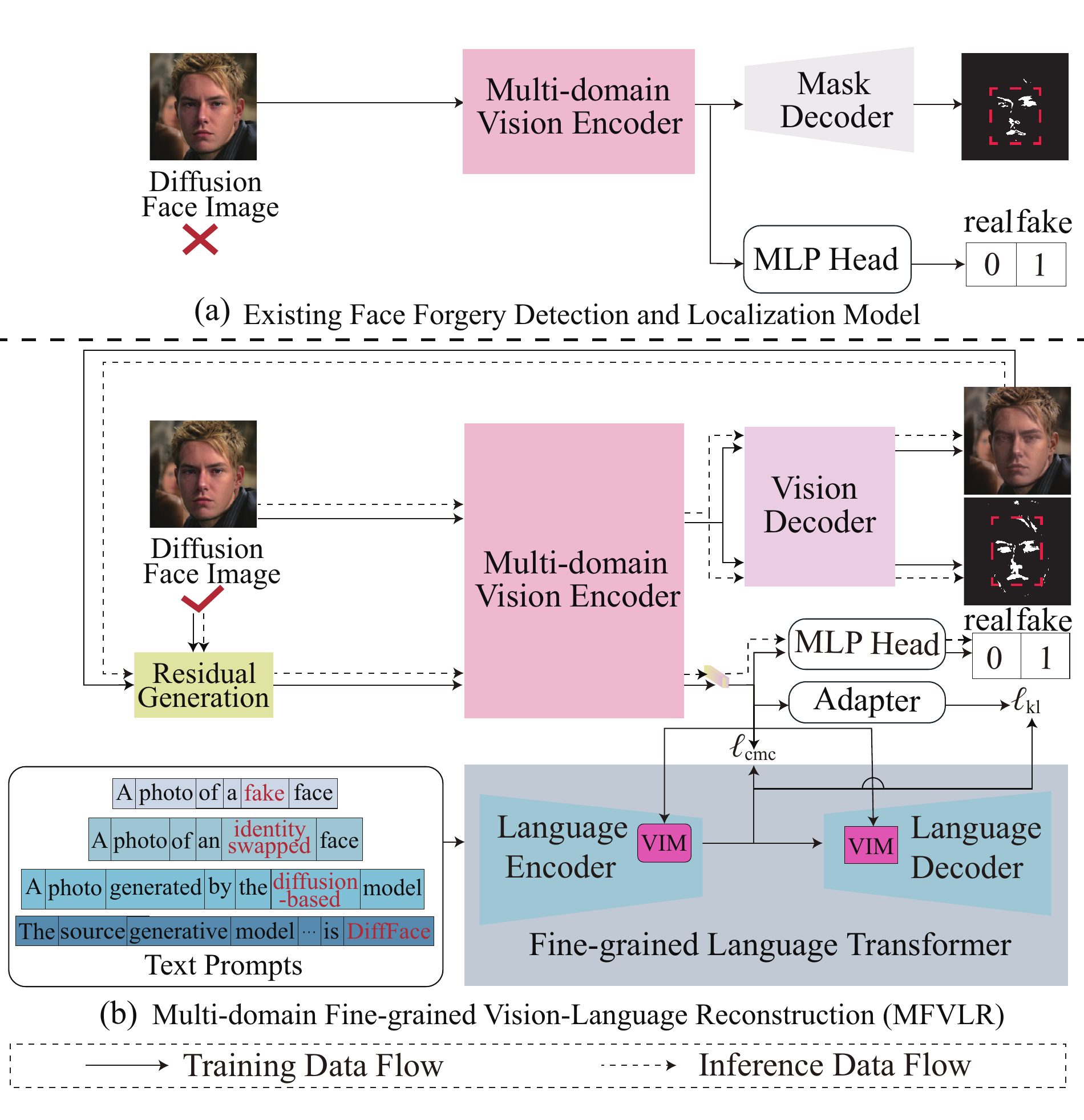} 
	\caption{An overview of the proposed MFVLR. (a) The traditional FFDL models are inclined to capture visual forgery patterns across space and frequency domains, and only study visual modality, which fails to generalize well to face images generated using diffusion models. (b) Our  MFVLR network explores general and fine-grained class-aware language embeddings as well as universal multi-domain visual forgery traces, via fine-grained vision-language reconstruction, to attain robust generalization to DFFDL. }\label{fig1}
	\vspace{-2em}
\end{figure}

Existing FFDL approaches primarily involve two categories, spatial-based methods \cite{DADF,FFD} and multi-domain-based methods \cite{M2TR,MSCCNet,HiFi-Net}. The former aims to explore and localize diverse and comprehensive face forgery patterns in the space domain. The latter intends to mine general face manipulated traces across space and frequency domains. However, there are some limitations for current FFDL models: \textbf{(1) Spatial-based methods \cite{DADF,M2TR} are inclined to extract forgery-insensitive features like identity and background, which results in poor generalization to unseen face forgery images.} For example, DADF \cite{DADF} is designed to explore spatial forgery patterns via the segment anything model (SAM), which demonstrates excellent detection and localization accuracy using within-dataset evaluation, but shows suboptimal performance on cross-dataset protocol. To solve this limitation, considering that previous works \cite{Frank,Durall} have shown that there are evident forgery traces yielded by generative approaches in the high-frequency domain, multi-domain-based methods \cite{MSCCNet,M2TR,HiFi-Net} intend to mine general and comprehensive face forgery patterns across space and frequency domains, to achieve FFDL. However, \textbf{(2) they fail to fully explore the fine-grained text modality, and barely study diffusion-synthesized face forgery detection and localization (DFFDL), which leads to limited generalizability to face images created by diffusions.} For instance, some networks \cite{M2TR,HiFi-Net} are proposed to identify and localize forgery regions in images generated by GANs via combining space and frequency features, but they struggle to consider the text modality, and have hardly evaluated the performance of models on images created by diffusion models. Vision-language-based methods (VLM) \cite{VLFFD,DD-VQA} attempt to capture universal manipulated facial embeddings using vision-language contrastive learning for FFD, but they struggle to achieve the generalizable DFFDL. 
\begin{figure}[t!]
	\centering
	\includegraphics[width=\linewidth]{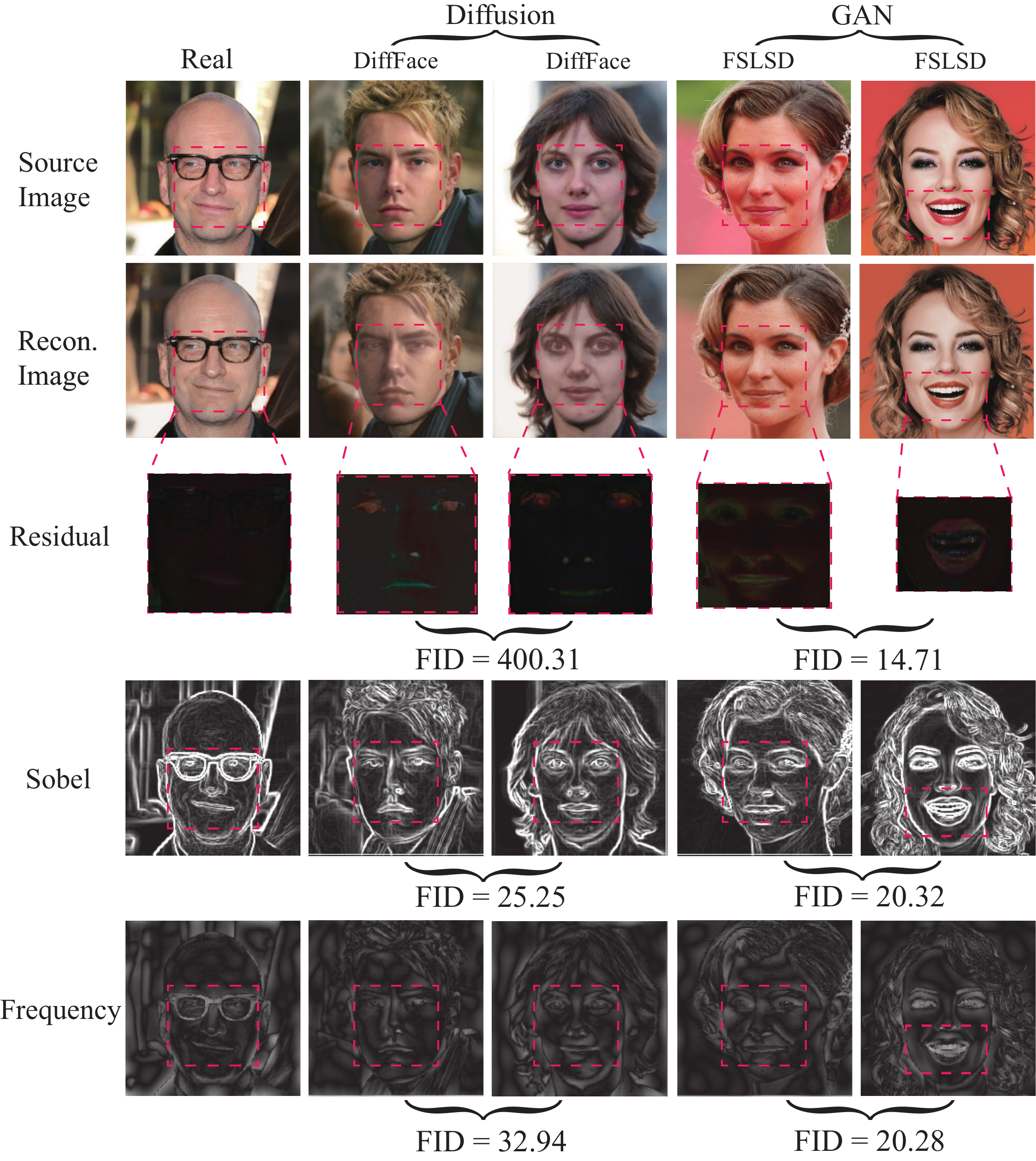} 
	\caption{The visualization of forgery priors from different domains. Each column shows a face yielded by various generators, including diffusion and GAN. The first to third rows represent the source image, reconstructed image, and the residual, respectively. The fourth and fifth rows show the edge image extracted by Sobel and the high-frequency image derived from the fast fourier transform (FFT), respectively. We randomly select 10k real prior images and 10k fake ones to calculate the FID score. The higher the FID score, the greater the difference between the real and fake prior distribution.}\label{fig2} 
\end{figure}

\begin{figure}[t]
	\centering
	\includegraphics[width=\linewidth]{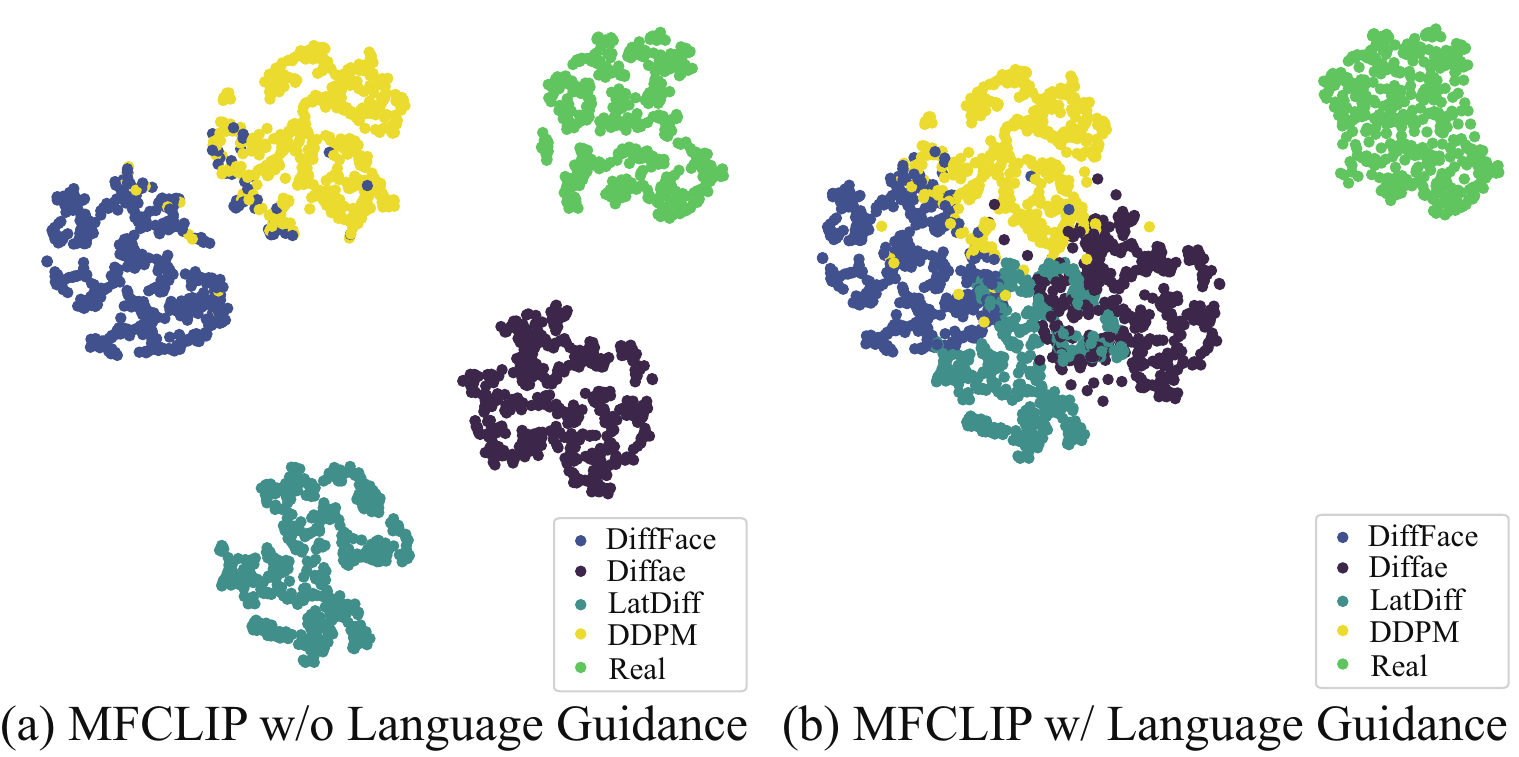} 
	\caption{The t-SNE visualization of various diffusion embeddings generated by MFCLIP \cite{mfclip} with or without language guidance. We randomly select 500 real samples, and for each diffusion model, we pick 500 fake samples.}\label{fig21} 
\end{figure}

Based on the aforementioned discussion, we aim to study fine-grained class-aware general language embeddings through language reconstruction, to enhance the learning of diffusion forgery-sensitive visual features, to realize generalizable DFFDL. In this paper, inspired by image reconstruction, we design a multi-domain fine-grained vision-language reconstruction (MFVLR) model. Unlike prior FFDL work which merely mines multi-domain face manipulated patterns, our MFVLR approach conducts the language-guided diffusion image-residual forgery representation and localization learning using the vision-language matching, to achieve generalization to face images generated by diffusions. In detail, our MFVLR framework mainly differs from existing FFDL models in the following aspects (see Fig.~\ref{fig1}): First, we notice that there are evident discrepancies between authentic and forged facial residual images (see Fig.~\ref{fig2}). Concretely, residuals are visually evident in manipulated images, but not visible in real ones. In addition, the residual in the diffusion-generated facial image is more discriminative than that in the GAN-synthesized one, compared to other priors extracted by edge Sobel or the frequency-based FFT. The FID score of the diffusion-generated residual mode is significantly higher than that of other diffusion priors, and there are evident FID score differences between GAN and diffusion-synthesized residual priors, compared to other edge and frequency distributions. This demonstrates the generalization advantage of reconstruction residuals over other forgery cues (e.g., frequency-domain artifacts). Thus, we propose a multi-domain vision encoder (MVE) with a residual encoder, to capture the comprehensive and characteristic residual traces, and diverse image-residual manipulated patterns. Second, we formulate a vision decoder (VD) to achieve the pixel-level forgery localization, and learn general image representations via the appearance image reconstruction. Finally, we are inspired by the observation that the language guidance mitigates diffusion-specific shifts (see Fig.~\ref{fig21}). In detail, without language prompts, diffusion samples are gathered in respective domains. When the language guidance is involved, various face samples synthesized by diffusion models begin to converge and eventually form a unified cluster. Therefore, we design a fine-grained language transformer (FLT) to extract general and fine-grained language features for preferable vision-language matching. Specifically, inspired by image reconstruction, we intend to boost general language embeddings via language reconstruction, and enhance the communication between vision and language representations. However, we note that VLM tends to ignore the interaction with vision features during language modelling. To solve this problem, we design a plug-and-play vision injection module (VIM) to efficiently conduct vision-guided language representation learning via fine-grained integration between vision and language forgery features at the word level. In summary, the contributions of this work are summarized as follows:

$\bullet$ We propose an innovative MFVLR model, which combines residuals with image forgery patterns, as well as enhances visual forgery features across image-residual domains via fine-grained class-aware language-guided face forgery representation and localization learning, to achieve the generalizable DFFDL. 

$\bullet$ We design a fine-grained language transformer to reconstruct language prompts, to enhance the general and comprehensive language representations through a language decoder, which could be integrated into any VLMs like CLIP to facilitate the vision-language feature alignment.

$\bullet$ We devise a novel vision injection module to enhance the communication between the vision and language features, which could be plugged and played into any VLMs to enhance their representation capability with merely a slight growth of the number of parameters.

$\bullet$ Extensive experiments and forgery localization visualizations demonstrate that the proposed approach exceeds the state of the art on various protocols, such as cross-generator evaluation, cross-forgery evaluation, cross-dataset evaluation, and robustness settings.  


	\vspace{-1em}
\section{Related Work}\label{sec2} 
	\vspace{-0.5em}
\subsection{Face Forgery Detection and Localization}
Prior approaches have achieved significant improvement in the domain of FFD. Most detectors capture forged traces in the space and frequency domains. A convolutional vision transformer (CViT) model \cite{CViT} is proposed to integrate CNN with vision transformer (ViT) \cite{ViT}  for face forgery detection, to model global forgery relations among image patches in the spatial domain. Zhu et al. \cite{3DDCS} decomposed a face image into several components, such as 3D geometry, lighting, common texture, and identity texture, and then employed a composition search strategy to automatically identify the most informative ones, their optimal combinations, and the most effective architecture for extracting forgery-related features. Qiao et al. \cite{Qiao} proposed a pseudo-label generator to label the training samples, which are then transmitted into the enhanced contrastive learner to extract and refine discriminative features. Diffusion reconstruction error (DIRE) \cite{DIRE} is designed to achieve diffusion image detection based on the discrepancy between an input image and the corresponding reconstruction.  In addition to relying on spatial features for face forgery detection, there are several frequency-based models. FreqNet \cite{Freq} is proposed to explore high-frequency manipulated representations across space and channels. To mine various forgery traces across different modalities, VLFFD \cite{VLFFD} is proposed to conduct the coarse-and-fine co-training network learning by combining fine-grained text embeddings with coarse-grained visual forgery features. DD-VQA \cite{DD-VQA} designs a multi-modal transformer model to enhance the learning of forgery patterns using text and image contrastive learning, to facilitate the FFD. However, they barely consider pixel-level face forgery localization. Lai et al. devised the detect any deepfakes (DADF) \cite{DADF} network based on the segment anything model (SAM) \cite{SAM}, to explore local and global forgery patterns, to achieve the FFDL. Guo et al. proposed the HiFi-Net \cite{HiFi-Net} to study both comprehensive representations and underlying hierarchical characteristics of various deepfake attributes. MSCCNet \cite{MSCCNet} is designed to capture multi-frequency facial manipulated representations based on various frequency bands, to facilitate the advancement of pixel-level DFFDL. SIDA \cite{SIDA} not only detects image authenticity but also identifies tampered regions via mask prediction and provides textual explanations for the model’s decisions, via large multi-modal models. DiffForensics \cite{DiffForensics} consists of a self-supervised denoising diffusion pre-training stage and a multi-task fine-tuning phase to achieve image forgery detection and localization. By contrast, we observe that there are considerable inconsistencies between authentic and diffusion-generated face residual images, so we study the global residual forgery traces, and integrate them with global spatial forensics to mine visually general diffusion-synthesized manipulated patterns. Furthermore, we focus on extracting fine-grained and general language embeddings corresponding to images using language reconstruction learning. Besides, we improve visual forgery features across image and residual domains by fine-grained vision-language contrastive learning, to achieve DFFDL.
\vspace{-1em}
\subsection{Vision-Language Models}
Unlike traditional image-based methods that utilize an image feature encoder along with a classifier to predict a specified set of prescribed sample classes, vision-language models (VLM) \cite{vls,zhao2024evaluating,yu2023task} like contrastive language-image pre-training (CLIP) \cite{clip} concurrently train an image encoder and a text encoder, to align image-text pairs from training datasets, and then infer on downstream tasks via a zero-shot classifier incorporated with the category names or descriptions from the target dataset, demonstrating powerful image representation capabilities in various fields such as visual question answering and text-guided image editing. CoOp \cite{ coop} explores the relations among context words of a prompt using learnable weights, to adaptively study the text embeddings. To alleviate the inherent one-to-one limitation and realize the flexible cross-modal alignment, SoftCLIP \cite{SoftCLIP} leverages fine-grained intra-modal self-similarities as softened targets for cross-modal contrastive learning. CFPL \cite{cfpl} is proposed to encode different semantic embeddings from text prompts conditioned on visual representations for generalizable face anti-spoofing. C2P-CLIP \cite{C2P-CLIP} introduces category common prompt CLIP, which integrates the category general prompt into the text encoder to add category-related concepts into the image encoder, leading to improved detection performance. MFCLIP \cite{mfclip} integrates fine-grained noise forgery embeddings with global image forgery patterns, and enhances them via flexible fine-grained vision-language matching, to achieve generalization to face images generated by diffusions. Different from traditional VLM which undergoes pretraining on large-scale general natural images, and then finetuning to enhance inference performance on downstream tasks such as FFD, our MFVLR model is trained from scratch using facial images generated by various methods like diffusion, to promote advancements of DFFDL. Besides, our approach is capable of digging visual face forgery representations across image-residual domains, extracting more general and fine-grained language embeddings using the language reconstruction scheme, and studying diverse and comprehensive vision-language features via the plug-and-play VIM. 
\vspace{-1em}
\subsection{Multi-Modal Fusion for Face Forgery Detection}
Deng et al. \cite{VG} propose an accumulated attention mechanism that iteratively aggregates attention over the image, query, and objects, and progressively filters out irrelevant noise while emphasizing the most useful information. MMDet \cite{mmdet} uses the multi-modal forgery representations from large multi-modal models (LMMs) to identify diffusion-generated videos, which mines features based on spatial artifacts and temporal inconsistencies via in-and-across frame attention modules. VLFFD \cite{VLFFD} introduces the face forgery text generator, which generates accurate text descriptions by leveraging forgery masks for initial region and type identification, followed by a structured prompting strategy to guide MLLMs in reducing hallucinations. M2F2-Det \cite{M2F2} adapts CLIP into the deepfake detection domain and leverages a bridge adapter to enhance LLM integration and generate reliable detection explanations. By contrast, based on the novel finding of the residual prior, our method extracts general diffusion-generated face forgery representations across image-residual domains, refines language embeddings through a language reconstruction scheme, and explores more precise vision-language matching to achieve DFFDL.
\begin{figure*}[t!]
	\centering
	\includegraphics[width=\linewidth]{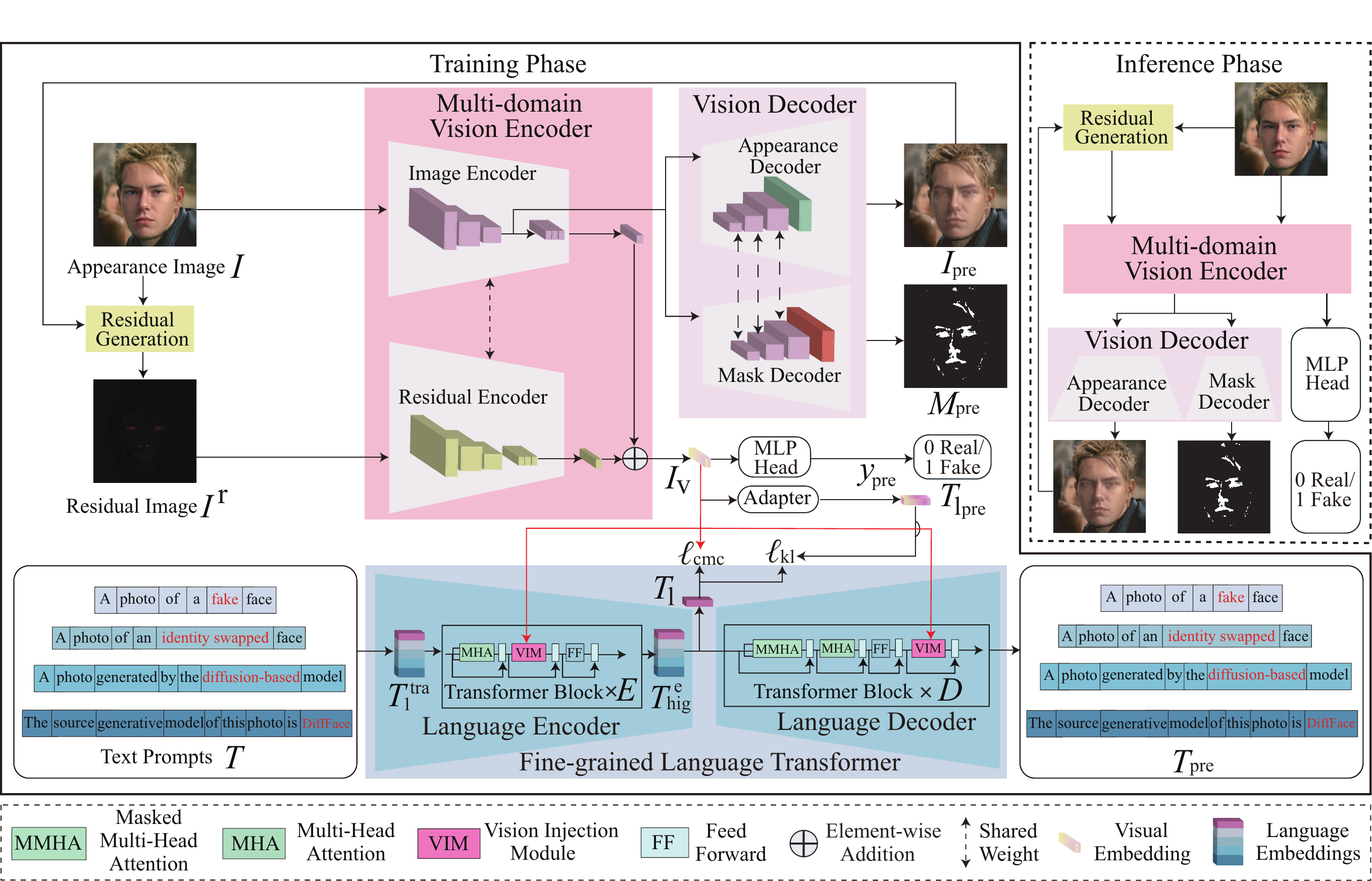} 
	\caption{The workflow of MFVLR. Given an input image-text pair, we first send the appearance image to MVE to study local appearance forgery features, which are then imparted to a VD module to reconstruct the appearance image and achieve pixel-level forgery localization. After that, the reconstructed appearance image and original image are sent to the residual generation module to generate the residual image, which is then fed into MVE to extract visual forgery patterns across the image and residual domains. Meanwhile, we transmit the corresponding text and visual forgery features to FLT to derive global language features and reconstructed text prompts based on the fine-grained vision-language interaction via VIM. Finally, visual forgery features are transferred into the adapter to match with global language ones via the KL loss, and the MLP head to yield the final prediction, respectively. During testing, the trained MVE, VD, and MLP head module are applied to achieve DFFDL.}\label{fig4} \vspace{-1em}
\end{figure*}
\vspace{-1em}
\section{Methodology} \label{sec3}

\subsection{Method Overview}

To achieve fine-grained language-guided visual face forgery representation learning and forgery localization, we devise the multi-domain fine-grained vision-language reconstruction (MFVLR) model for generalizable DFFDL. Unlike vision-language-based models such as MFCLIP, which rely on fine-grained image, noise, and text embeddings, our approach specifically integrates residual-aware features with global image forgery patterns and enhances them via an appearance decoder, and further improves the fine-grained text features via language reconstruction. In addition to boosting global visual forgery traces across image-residual domains, we devise an innovative plug-and-play vision injection module (VIM) to conduct diverse, fine-grained global residual-aware vision-language interaction for precise vision-language matching. As Fig.~\ref{fig4} illustrates, MFVLR mainly includes three components: multi-domain vision encoder (MVE), vision decoder (VD), and fine-grained language transformer (FLT). During the training stage, given an input image-text pair $(I, T)\in \mathcal{D}$ with one-hot ground truth labels $y\in\big\{[0,1]^T,[1,0]^T\}$ and ground truth masks $M\in\mathbb{R}^{224\times224}$ from the GenFace dataset $\mathcal{D}$ \cite{genface}, where we generate the fine-grained text prompts $T$ based on the hierarchical fine-grained labels, and the ground truth masks $M$ based on the fake and corresponding source image pairs, MFVLR first extracts local appearance forgery features $I_\text{loc}$ from the appearace image $I$, which are then transmitted to VD to generate the predicted appearance image $I_\text{pre}$ and the predicted mask $M_\text{pre}$. Thereafter, $I$ and $I_\text{pre}$ are fed into the residual generation module to produce the residual image $I^\text{r}$. MFVLR captures global visual forgery features $I_\text{v}$ across image and residual domains from $I^\text{r}$ and $I$ through MVE. Meanwhile, MFVLR encodes comprehensive and diverse language representations from $T$ via FLT. After that, $I_\text{v}$ is fed into an adapter to match with language embeddings, and the multilayer perceptron (MLP) head with a full connection layer to produce the final predicted detection logits $y_\text{pre}$, respectively. We further employ the cross-modal contrastive loss to conduct vision-language matching.

During testing, to prevent information leakage from texts, and considering that face images created by the new generative model may lack text prompts, we incorporate fine-grained texts to train our network exclusively during training. In detail, given an appearance image $I$, MFVLR yields local appearance forgery features $I_\text{loc}$ using MVE, which are then transmitted to VD to derive $I_\text{pre}$ and mask $M_\text{pre}$. Thereafter, $I_\text{pre}$ and $I$ are transferred to the residual generation module to yield the residual image $I_\text{r}$, which is then transferred into MVE to generate $I_\text{v}$ across the image-residual domains. $I_\text{v}$ is fed into the MLP head to generate  $y_\text{pre}$.

\vspace{-1em}
\subsection{Multi-domain Vision Encoder} \label{secmve}
Different from traditional multi-domain FFD approaches \cite{M2TR,twostream} that tend to coarsely focus on the communication between the local frequency or noise and RGB information, we design the multi-domain vision encoder (MVE) to dig global image manipulated patterns as well as fine-grained residual forgery traces, and fuse them, simply and effectively, to capture comprehensive diffusion image-residual forgery patterns. As shown in Fig.~\ref{fig4}, MVE primarily contains an image encoder (IE) and a residual encoder (RE). 

{\bfseries\setlength\parindent{0em}Image encoder.} To learn global diffusion facial forgery representations, we design IE, which consists of a UNet encoder \cite{unet} and a transformer encoder (TE) with $B$ image transformer blocks $\text{TB}_j^\text{i}$, $j=1,2,…, B$. Specifically, given an input appearance image $I\in\mathbb{R}^{3\times224\times224}$, IE extracts the local appearance forgery features $I_\text{loc} \in\mathbb{R}^{ c\times h\times w}$ via the UNet encoder, where $c$, $h$, $w$ denotes the channel, height, and width of the feature map. Thereafter, $I_\text{loc}$ is transferred into VD (see Sec.~\ref{secvd}) to acquire the predicted appearance image $I_\text{pre}\in\mathbb{R}^{3\times224\times224}$, which is then fed into the residual generation module to obtain the residual image $I^\text{r}\in\mathbb{R}^{3\times224\times224}$, i.e.,  $I^\text{r} = |I_\text{pre} - I|$.
Meanwhile, IE sends the $I_\text{loc}$ to the transformer encoder (TE), to explore global relations among feature patches. Specifically, $I_\text{loc}$ is first flattened and projected to 2D token sequences with the dimension of $d$ along the channel. Thereafter, it is appended with a learnable class token to learn the global image forgery embeddings, to gain $I_\text{tok}=\text{App}(\text{Proj}(\text{Flat}(I_\text{loc})))\in\mathbb{R}^{ (hw+1)\times d}$, and then added with a learnable position embedding $P_\text{i}\in\mathbb{R}^{(hw+1)\times d}$ to study the position information. That is,
\begin{align}
	I_1^\text{tra}=I_\text{tok}+P_\text{i}.
\end{align}
Afterwards, it is successively transmitted into $B$ blocks, i.e.,
\begin{align}
	\text{TE}(I_1^{\text{tra}})
	&=\text{TB}_B^\text{i}\circ\text{TB}_{B-1}^\text{i}\circ \cdots \circ  \text{TB}_2^\text{i} \circ\text{TB}_1^\text{i}(I_1^\text{tra}) \nonumber \\
	&=\text{TB}_B^\text{i}\circ\text{TB}_{B-1}^\text{i}\circ \cdots \circ \text{TB}_2^\text{i}(I_2^\text{tra})
	\nonumber \\
	&= \cdots =\text{TB}_B^\text{i}\circ\text{TB}_{B-1}^\text{i}(I_{B-1}^\text{tra})\nonumber \\
	&= \text{TB}_B^\text{i}(I_B^\text{tra})     \nonumber \\
	&=I_\text{TE},
\end{align}
where $\circ$ denotes the function decomposition. IE extracts the global appearance forgery representations $I_\text{g}\in\mathbb{R}^{1\times d} $ using the class token in $ I_\text{TE}$. 

{\bfseries\setlength\parindent{0em} Residual encoder.} To capture global and fine-grained residual forgery traces, we propose the residual encoder (RE), which shares the same architecture and weights with IE, to reduce the number of parameters. Likewise, RE encodes the global residual forgery features $I^\text{r}_\text{g}\in\mathbb{R}^{1\times d}$ from the residual image $I^\text{r}$, i.e. $I^\text{r}_\text{g}=\text{RE}(I^\text{r})$.
Finally, MVE generates the abundant image-residual visual manipulated patterns $I_\text{v}$ by integrating  $I^\text{r}_\text{g}$ with $I_\text{g}$, i.e., $I_\text{v}=I^\text{r}_\text{g}+I_\text{g}$, which are then transmitted to the adapter and the MLP head, respectively.
\vspace{-1em}
\subsection{Vision Decoder}\label{secvd}
To promote the model to learn universal forgery representations and achieve pixel-level manipulation localization, we devise the vision decoder (VD). As Fig.~\ref{fig4} illustrates, VD is composed of an appearance decoder and a mask decoder.

{\bfseries\setlength\parindent{0em}  Appearance decoder.} To capture general facial forgery traces and extract the residual information, we devise the appearance decoder (AD). AD consists of a UNet decoder along with an appearance reconstruction module with a convolutional layer. Specifically, given the local appearance forgery features $I_\text{loc}$ derived from IE, AD studies the general image features to generate the predicted appearance image $I_\text{pre}$, i.e., $I_\text{pre}= \text{AD}(I_\text{loc})$.

{\bfseries\setlength\parindent{0em} Mask decoder.} 
To achieve face manipulation localization, we devise the mask decoder (MD). MD is composed of a UNet decoder along with a manipulation localization module with a convolutional layer. The UNet decoder adopts the same network and weights as those in AD, to decrease the number of parameters. Specifically, given local appearance forgery features $I_\text{loc}$ (see Sec.~\ref{secmve}), MD extracts the discriminative forgery patterns to yield the predicted mask $M_\text{pre}\in\mathbb{R}^{f\times 224\times224}$, i.e., $M_\text{pre}= \text{MD}(I_\text{loc})$, where $f$ is the number of detection category.
 
\begin{figure}[t!]
	\centering
	\includegraphics[width=\linewidth]{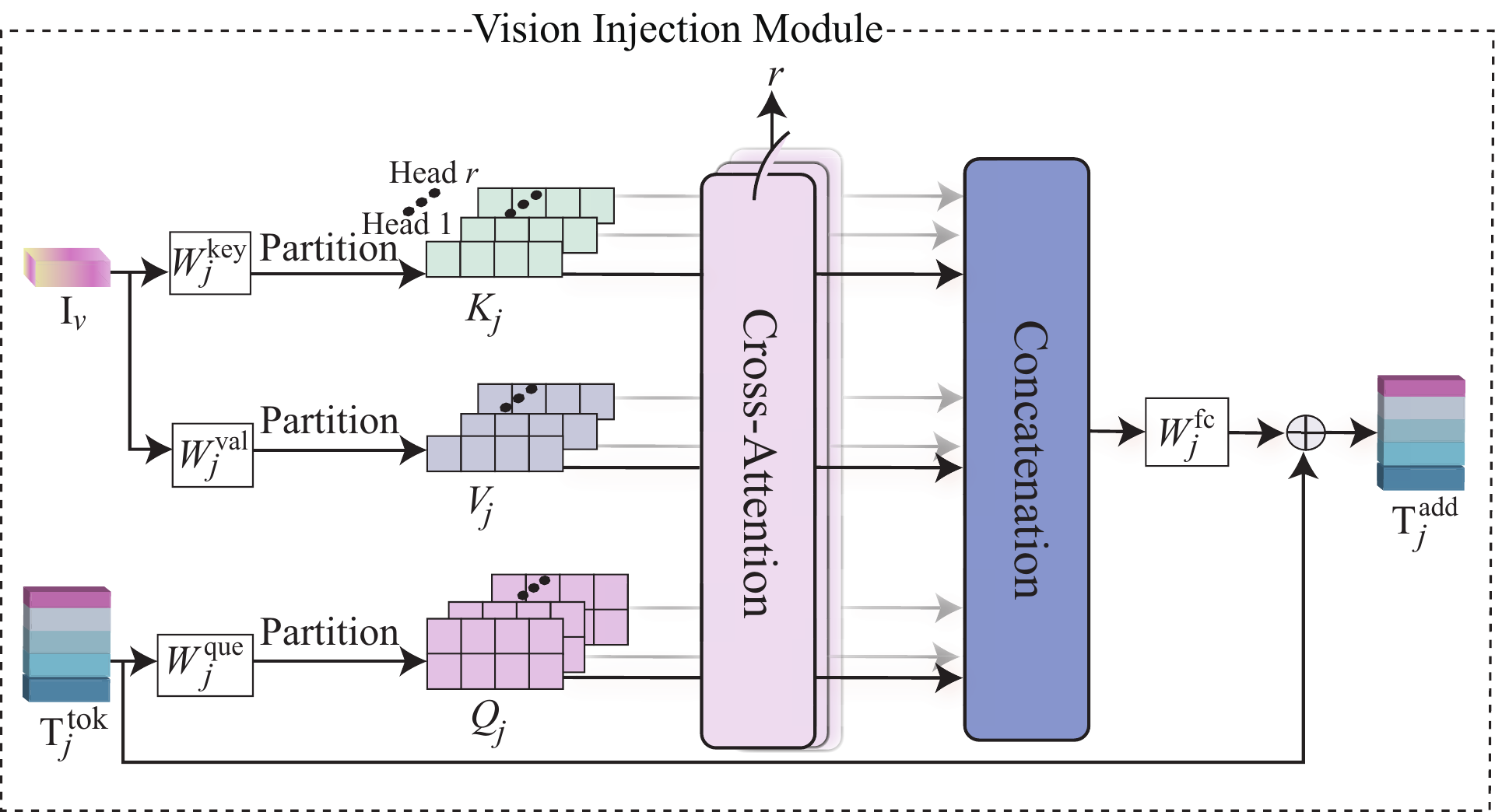} 
	\caption{The pipeline of VIM. Vision forgery embeddings are projected into the key and value metrics, respectively, and the language feature is used as a query to interact with the vision forged feature at the word level, to extract diverse fine-grained vision-language representations.
	}\label{vim}
\end{figure}
\vspace{-1em}
\subsection{Fine-grained Language Transformer}\label{flt}
Unlike traditional language transformer models that merely explore the global relationships among words, we design the fine-grained language transformer (FLT) to capture global language representations and diverse fused vision-language embeddings. As Fig.~\ref{fig4} shows, FLT consists of a language encoder and a language decoder, both of which include the vision injection module, to conduct the vision-guided language representation learning.

 	\begin{figure}[t!]
 	\centering
 	\includegraphics[width=\linewidth]{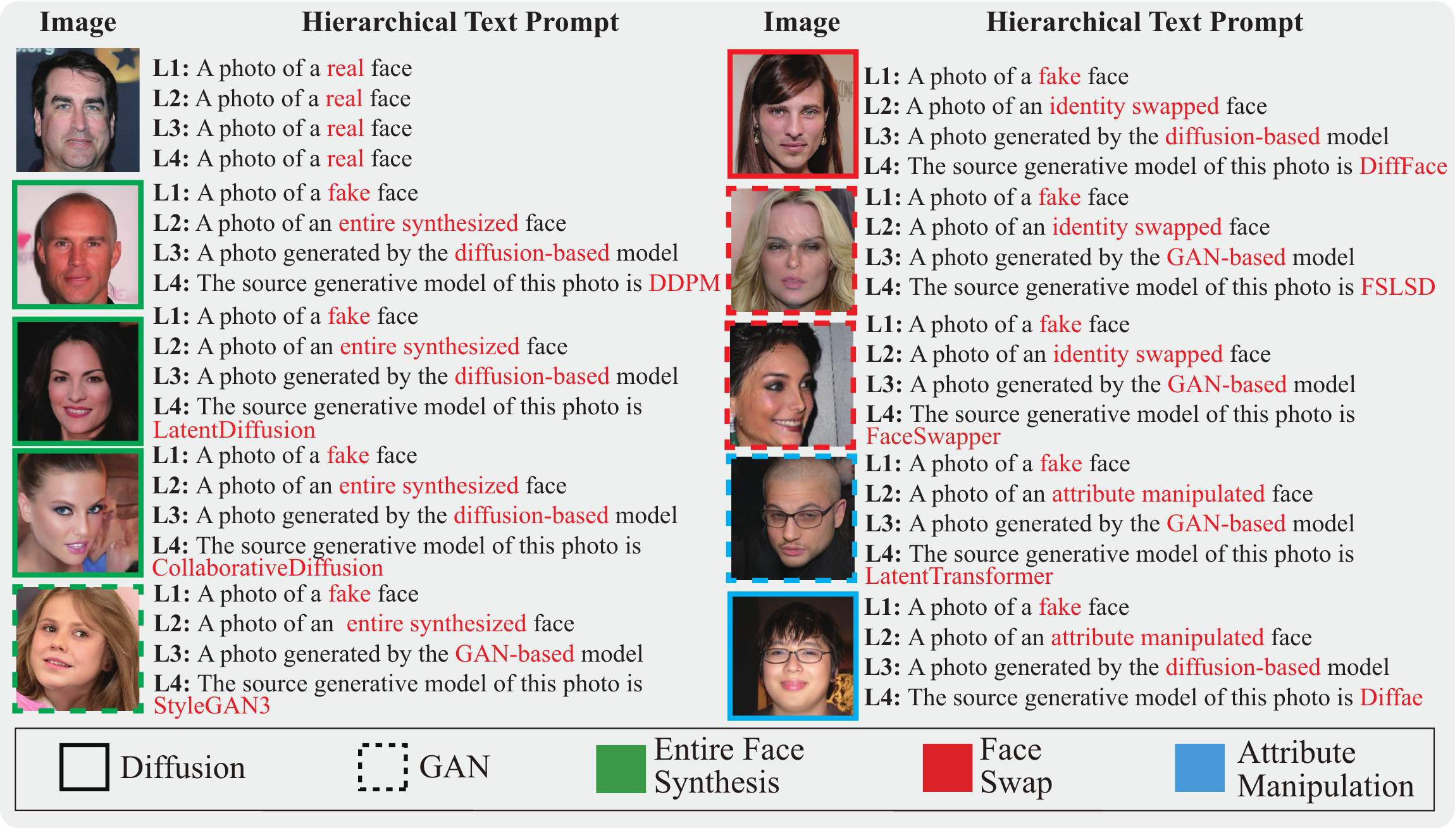} 
 	\caption{ The typical level 1 (L1) to level 4 (L4) examples for different forgery types and generators. }\label{figl14}
 	\vspace{-1em}
 \end{figure}

{\bfseries\setlength\parindent{0em}Language encoder.} To encode the fine-grained language representations, we design the language encoder (LE), which consists of $E$ LE transformer blocks ${\text{TB}_j^\text{e}}$, $j=1,2,…, E$. We introduce the fine-grained text generator (FTG) \cite{mfclip} to generate hierarchical text prompts  $T$ from level 1 to level 4 (see Fig.~\ref{figl14}) for each image based on hierarchical fine-grained labels offered by the GenFace dataset. Specifically, given $T$, we utilize the tokenizer \cite{clip} to acquire a sequence of word tokens $T_\text{tok}^\text{e}\in\mathbb{R}^{n}$, where $n$ denotes the number of word tokens. It is then projected to the low-level language embeddings $T_\text{low}^\text{e}\in\mathbb{R}^{n\times d} $ via a vocabulary $W_\text{voc}\in\mathbb{R}^{s\times d}$, where $s$ is the vocabulary size, and integrated with position embedding $P_\text{e}\in\mathbb{R}^{n\times d}$, i.e., $T_\text{1}^\text{tra}={\ T}_\text{low}^\text{e} +P_\text{e}$. Afterwards, it is sequentially transmitted to $E$ blocks to derive high-level language embeddings $T^\text{e}_\text{hig}$, i.e., 
\begin{align}
	\text{LE}(T_\text{1}^{\text{tra}})
	&=\text{TB}_E^\text{e}\circ\text{TB}_{E-1}^\text{e}\circ \cdots \circ  \text{TB}_2^\text{e} \circ\text{TB}_1^\text{e}(T_1^\text{tra}) \nonumber \\
	&=\text{TB}_E^\text{e}\circ\text{TB}_{E-1}^\text{e}\circ \cdots \circ \text{TB}_2^\text{e}(T_2^\text{tra})
	\nonumber \\
	&= \cdots =\text{TB}_E^\text{e}\circ\text{TB}_{E-1}^\text{e}(T_{E-1}^\text{tra})\nonumber \\
	&= \text{TB}_E^\text{e}(T_E^\text{tra})     \nonumber \\
	&=T^\text{e}_\text{hig}.
\end{align}
LE studies the global language embeddings $T_\text{l}\in\mathbb{R}^{1\times d} $ using the last token in $T^\text{e}_\text{hig}$.

{\bfseries\setlength\parindent{2em} Transformer block.} Unlike vanilla transformer blocks which only model the global relationships among words, and tend to struggle with comprehensive and diverse global information due to the limitation of a single modality, our transformer block introduces the proposed plug-and-play vision injection module (VIM) to conduct vision-guided language representation learning to extract extensive global vision-language representations. Specifically, as Fig.~\ref{fig4} illustrates, in the $j$-th LE transformer block ${\text{TB}_j^\text{e}}$, the enhanced language embedding $T_j^\text{tra}$ is first fed to a layer normalization (LN) to facilitate the model learning, and then transmitted to the multi-head attention (MHA) module to capture the global relations among words, $T_j^\text{tok}=\text{MHA}_j^\text{e}(\text{LN}_j^\text{e}(T_j^\text{tra}))+T_j^\text{tra}\in\mathbb{R}^{n\times d}$. To study the abundant vision-language representations efficiently and provide a flexible framework that can easily be integrated with different architectures, we design the VIM module. Unlike previous interaction mechanisms, our VIM method is designed to achieve comprehensive and fine-grained interaction between vision and language features at the word level, efficiently for DFFDL, which could be plugged and played into transformer-based vision-language models without requiring extensive architectural changes and parameter count growth. As Fig.~\ref{vim} illustrates, VIM first transforms the diverse language representations $T_j^\text{tok}$ to query using the learnable parameter matrices $W^\text{que}_j\in\mathbb{R}^{d\times d}$, and the vision forgery class token embedding $I_\text{v}\in\mathbb{R}^{1\times d}$ to key and value using the learnable weight matrices $W^\text{key}_j$ and $W^\text{val}_j\in\mathbb{R}^{d\times d}$, respectively. Note that only the class token instead of all tokens or patch tokens from the vision forged features, is used as the key and value, which ensures that generating the attention map incurs a linear computational cost instead of a quadratic one.
\begin{align}
q_j &= T_j^\text{tok}W^\text{que}_j,\\
k_j &= I_\text{v}W^\text{key}_j,\\
v_j &= I_\text{v}W^\text{val}_j.
\end{align}
 The query, key, and value matrices are then partitioned into $r$ heads, respectively,
\begin{align}
	\{Q_{j,i}\in\mathbb{R}^{n\times \frac{d}{r}}\}^r_{i=1} &= \text{Pa}(q_j),\\
	\{K_{j,i}\in\mathbb{R}^{1\times \frac{d}{r}}\}^r_{i=1} &= \text{Pa}(k_j),\\
	\{V_{j,i}\in\mathbb{R}^{1\times \frac{d}{r}}\}^r_{i=1} &= \text{Pa}(v_j).
\end{align}
Unlike traditional fusions like summation which may merge information at a coarse level, to achieve the diverse and fine-grained interaction between the vision and language features, we conduct the cross-attention calculation in parallel in each head space to inject vision forged features into each word,
\begin{align}
 T_{j,i}^\text{glo} &=\delta(\frac{Q_{j,i}K_{j,i}^T}{\sqrt{\frac{r}{d}}})V_{j,i},
\end{align}
where $\delta$ is the softmax function. Thereafter, the cross-attention outputs from all heads are concatenated to yield diverse global vision-language representations,
\begin{align}
	T_{j}^\text{glo} &=\text{Cat}(\{ T_{j,i}^\text{glo}\}^r_{i=1})\in\mathbb{R}^{n\times d}.
\end{align}
Thereafter, it is fed into a fully connected layer and then added with $T_j^\text{tok}$ to enhance the vision-language representation, i.e., $T_{j}^\text{add} = T_{j}^\text{glo}W^\text{fc}_j+T_j^\text{tok}$, where $W^\text{fc}_j\in\mathbb{R}^{d\times d}$ is the learnable weight of the fully connected layer.
Finally, it is transmitted into a feed forward (FF) network with a fully connected layer to yield $T_{j+1}^\text{tra}=\text{FF}_j^\text{e}(T_{j}^\text{add})+T_{j}^\text{add}\in\mathbb{R}^{n\times d}$.

{\bfseries\setlength\parindent{0em} Language decoder.} Inspired by image reconstruction methods, we propose to perform text reconstruction to enhance the general language representation capabilities of the model. Text reconstruction refers to the process of recovering or generating the original text prompt from a given text prompt. Therefore, we design the language decoder (LD) with $D$ LD transformer blocks ${\text{TB}_j^\text{d}}$, $j=1,2,…, D$. Unlike conventional language decoders, we introduce visual embeddings to guide the model to reconstruct text prompts using VIM. Specifically, given the low-level language embeddings $T_\text{low}^\text{e}\in\mathbb{R}^{n\times d}$ of the input fine-grained text prompt $T$, the high-level language embeddings $T^\text{e}_\text{hig}$ generated by LE, and the visual forgery features $I_\text{v}$ derived from MVE, $T_\text{low}^\text{e}$ is appended with a begin token and removed the last token to keep the number of tokens unchanged and predict the next word in sentences, and then added with the position embedding $P_\text{d}\in\mathbb{R}^{n\times d}$ to gain the $T_{\text{t}_1}^\text{tra}\in\mathbb{R}^{n\times d}$. We sequentially fed $T_{\text{t}_1}^\text{tra}$, $T^\text{e}_\text{hig}$, and $I_\text{v}$ to $D$ blocks to derive reconstruced text embeddings $T^\text{d}_\text{rec}\in\mathbb{R}^{n\times d}$, i.e.,
\begin{align}
	\text{LD}(T_{\text{t}_1}^{\text{tra}},T^\text{e}_\text{hig},I_\text{v})
	&=\text{TB}_D^\text{d}\circ \cdots \circ  \text{TB}_2^\text{d} \circ\text{TB}_1^\text{d}(T_{\text{t}_1}^\text{tra},T^\text{e}_\text{hig},I_\text{v}) \nonumber \\
	&=\text{TB}_D^\text{d}\circ \cdots \circ \text{TB}_2^\text{d}(T_{\text{t}_2}^\text{tra},T^\text{e}_\text{hig},I_\text{v})
	\nonumber \\
	&= \cdots  = \text{TB}_D^\text{d}(T_{\text{t}_D}^\text{tra},T^\text{e}_\text{hig},I_\text{v})  \nonumber \\
	&=T^\text{d}_\text{rec}.
\end{align}

As Fig.~\ref{fig4} illustrates, each LD transformer block ${\text{TB}_j^\text{d}}$ contains the masked multi-head attention module (MMHA) $\text{MMHA}_j^\text{d}$, multi-head attention (MHA) $\text{MHA}_j^\text{d}$, the FF layer $\text{FF}_j^\text{d}$, and the VIM module $\text{VIM}_j^\text{d}$. In ${\text{TB}_j^\text{d}}$, $T_{\text{t}_j}^\text{tra}$ is fed into the MMHA module to prevent information leakage, to obtain language embeddings in which each word only integrates the information of itself and previous ones, and then performed residual addition as follows:
\begin{align} T_{\text{t}_j}^\text{mmha}=\text{MMHA}_j^\text{d}(T_{\text{t}_j}^\text{tra})+T_{\text{t}_j}^\text{tra}.
\end{align}	
	 After that, $T_{\text{t}_j}^\text{mmha}$ and $T^\text{e}_\text{hig}$ are transmitted to the MHA module to achieve the communication between the low-level and high-level language embeddings, to produce fused language representations. They are then conducted residual addition, i.e., 
	 \begin{align} T_{\text{t}_j}^\text{mha}=\text{MHA}_j^\text{d}(T_{\text{t}_j}^\text{mmha},T^\text{e}_\text{hig})+T_{\text{t}_j}^\text{mmha}.
\end{align}	
Thereafter, $T_{\text{t}_j}^\text{mha}$ is fed to the FF layer and then executed residual summation as below:
	 \begin{align} T_{\text{t}_j}^\text{ff}=\text{FF}_j^\text{d}(T_{\text{t}_j}^\text{mha})+T_{\text{t}_j}^\text{mha}.
\end{align}	
Finally, $T_{\text{t}_j}^\text{ff}$ and visual forgery features $I_\text{v}$ are transferred to the VIM to enhance the interaction between the vision and language representations, and then added with $T_{\text{t}_j}^\text{ff}$ to improve representations. That is,
	 \begin{align} 
	 	T_{\text{t}_{j+1}}^\text{tra
	 	}=\text{VIM}_j^\text{d}(T_{\text{t}_j}^\text{ff},I_\text{v})+T_{\text{t}_j}^\text{ff}.
\end{align}	

\subsection{Loss Function}
In Table~\ref{tabnota}, we include a dedicated table summarizing the key notation system.

{\bfseries\setlength\parindent{0em} Appearance reconstruction loss.}
To enhance general image forgery representations, we introduce the appearance reconstruction (AR) loss to reduce the discrepancy between the predicted appearance image $I_\text{pre}$ (see Sec.~\ref{secmve}) and the input appearance image $I$, as follows:
\begin{align}
	\mathcal{L}_\text{ar}=\frac{1}{b}\sum_{u=1}^{b} {(I^u-I_\text{pre}^u)}^2, 
\end{align}
where $b$ denotes the number of samples in a batch, and $u\in b$ is the index of samples.

{\bfseries\setlength\parindent{0em} Forgery localization loss.} 
To create the ground truth mask, we pair the forgery images such as face-swapped images and attribute-edited ones with their corresponding source image, measure the absolute pixel-wise difference in the RGB channels, convert it into grayscale, and then divide it by 255 to yield a map with the range of [0, 1]. We empirically define the threshold of 0.1 to generate the ground truth mask $M\in\mathbb{R}^{224\times224}$. We set the ground truth mask of real face images to 0, and that of the entire synthesized face images to 1.
To diminish the divergence between the predicted mask $M_\text{pre}$ and the ground truth mask $M$, we design the forgery localization (FL) loss, to supervise the model to achieve pixel-level predictions. That is,
\begin{align}
	\mathcal{L}_\text{fl}=\ \frac{1}{b}\sum_{u=1}^{b}-({M^u})^T\text{log}(M_\text{pre}^u).
\end{align}

{\bfseries\setlength\parindent{0em}  Kullback-leibler divergence loss.} We fed visual forgery embeddings $I_\text{v}$ to an adapter with a full connection layer, to generate the predicted language features $T_{\text{l}_\text{{pre}}}\in\mathbb{R}^{1\times d}$. To bring $T_{\text{l}_\text{{pre}}}$ closer to the global language embeddings $T_\text{l}\in\mathbb{R}^{1\times d}$ derived from LE (see Sec.~\ref{flt}), we introduce the kullback-leibler (KL) divergence loss, to conduct the feature alignment, i.e., 
\begin{align}
	\mathcal{L}_\text{kl}= \frac{1}{b}\sum_{u=1}^{b}{\delta{(T_\text{l}^u)}^T\text{log}\frac{{\delta(T}_\text{l}^u)}{\delta(T_{\text{l}_{\text{pre}}}^u)}},
\end{align}
where $\delta$ is the softmax function with temperature 0.5, to smooth features.

{\bfseries\setlength\parindent{0em}Cross-modal contrastive loss.}
To conduct vision-language contrastive learning, like CLIP, we introduce the cross-modal contrastive (CMC) loss, to improve the visual forgery features. Formally, for the vision-language embedding pairs $\{(I_\text{v}^{u},T_\text{l}^{u})\}_{u=1}^{b}$, we compute the vision-language cosine similarity vector and the language-vision one as follows:
\begin{align}
	S_\text{v2l}^{um}(I_\text{v}, T_\text{l})&=\frac{\text{exp}(\text{sim}{(I}_\text{v}^u{,T}_\text{l}^m)/\tau)}{\sum_{m=1}^{b}{\text{exp}(\text{sim}{(I}_\text{v}^u{,T}_\text{l}^m)/\tau)}}, \\
	S_\text{l2v}^{um}({T_\text{l},I}_\text{v})& = \frac{\text{exp}(\text{sim}{(T}_\text{l}^u{,I}_\text{v}^m)/\tau)}{\sum_{m=1}^{b}{\text{exp}(\text{sim}{(T}_\text{l}^u{,I}_\text{v}^m)/\tau)}},
\end{align}
where $\tau$ is a trainable temperature weight initialized with
0.07, and the function sim(·) conducts a dot product to
calculate the similarity scores. $u, m=1,2,...,b$ are the index of the sample. 
 The one-hot label $y_\text{pa}$ of the $u$-th pair is defined as $y_\text{pa}^u=\{{y_\text{pa}^{um}}\}_{m=1}^b$, $y_\text{pa}^{uu}=1$,
$y_\text{pa}^{um,u\neq m}=0$. To draw positive pairs together, while separating negative ones, the CMC loss is denoted as $\mathcal{L}_\text{cmc}=\ (\mathcal{L}_\text{v2l}+\mathcal{L}_\text{l2v})/2$, where
\begin{align}
	\mathcal{L}_\text{v2l}
	&=\frac{1}{b}\sum_{u=1}^{b}{-({y_\text{pa}^u})^T\text{log}(}S_\text{v2l}^u\left(I_\text{v},T_\text{l}\right)),\\
	\mathcal{L}_\text{l2v}
	&=\frac{1}{b}\sum_{u=1}^{b}{-({y_\text{pa}^u})^T\text{log}(}S_\text{l2v}^u(T_\text{l},I_\text{v})).
\end{align}

{\bfseries\setlength\parindent{0em} Language reconstruction loss.}
To facilitate the general fine-grained class-aware language representations and achieve language reconstruction, we introduce the language reconstruction (LR) loss function, to minimize the distance between reconstructed text prompts and the authentic ones. In detail, given the reconstructed text features $T^\text{d}_\text{rec}\in\mathbb{R}^{n\times d}$ derived from LD (see Sec.~\ref{flt}), we acquire the predicted reconstructed text label $T_\text{pre}$ as follows:
\begin{align}
	T_\text{pre}=T^\text{d}_\text{rec}W^T_\text{voc}\in\mathbb{R}^{n\times s},
\end{align}	
where $W_\text{voc}\in\mathbb{R}^{s\times d}$ denotes the vocabulary and $s$ is the vocabulary size. The LR loss is formally expressed as follows:
\begin{align}
	\mathcal{L}_\text{lr}= \frac{1}{b}\sum_{u=1}^{b}\sum_{x=1}^{n}-({T^{u,x}_\text{gt}})^T\text{log}(T_\text{pre}^{u,x}),
\end{align}
where $x$ is the index of the word, and $T_\text{gt}\in\mathbb{R}^{n\times s}$ denotes the text ground truth one-hot label.

{\bfseries\setlength\parindent{0em}  Forgery detection loss.} To mine discriminative forgery patterns, we introduce the forgery detection (FD) loss function. Specifically, we fed visual forgery features $I_\text{v}$ derived from MVE (see Sec.~\ref{secmve}) to the MLP head consisting of a fully connected layer, to yield the predicted detection logits $y_\text{pre}$. The FD loss is denoted as follows:
\begin{align}
	\mathcal{L}_\text{fd}=\ \frac{1}{b}\sum_{u=1}^{b}{-({y^u})^T\text{log}(}y_\text{pre}^u).
\end{align}

The total loss function is denoted as follow:
\begin{align}
	\mathcal{L}=		\mathcal{L}_\text{fd}+\mathcal{L}_\text{lr}+\mathcal{L}_\text{cmc}+	\mathcal{L}_\text{fl}+\mathcal{L}_\text{ar}+\mathcal{L}_\text{kl}.
\end{align}

\begin{table}[t]
	\caption{Summary of key notation system used in this paper.
\vspace{-1.5em}
		\label{tabnota}}
	\setlength{\tabcolsep}{0.3mm}{ 
		\begin{tabular}{rcccc}\toprule\multicolumn{1}{c}{Symbol} & \multicolumn{4}{c}{Description} \\\midrule    \multicolumn{1}{l}{$I$, $I^\text{r}$, $I_\text{pre}$}& \multicolumn{4}{l}{Appearance, residual, and predicted appearance image } \\    \multicolumn{1}{l}{ $I_\text{loc}$, $I_\text{g}$}& \multicolumn{4}{l}{Local, and global appearance forgery features } \\     \multicolumn{1}{l}{$I^\text{r}_\text{g}$, $I_\text{v}$}& \multicolumn{4}{l}{Global residual, and visual forgery features } \\
			\multicolumn{1}{l}{	$\text{TB}_j^\text{i}$,$\text{TB}_j^\text{e}$,$\text{TB}_j^\text{d}$}& \multicolumn{4}{l}{Image, LE, and LD transformer block of the $j$-th. } \\
			\multicolumn{1}{l}{ $M$,	$M_\text{pre}$  }& \multicolumn{4}{l}{  Ground truth mask, and predicted mask  } \\ 
			\multicolumn{1}{l}{ $y$,	$y_\text{pre}$ }& \multicolumn{4}{l}{  Image ground truth one-hot labels, predicted detection logits } \\
			\multicolumn{1}{l}{$T$, $T_\text{pre}$ }& \multicolumn{4}{l}{Fine-grained text prompts, predicted reconstructed text labels} \\
			\multicolumn{1}{l}{ $T_\text{l}$, $T_\text{low}^\text{e}$, $T^\text{e}_\text{hig}$ }& \multicolumn{4}{l}{Global, low-level, and high-level language embeddings} \\
			\multicolumn{1}{l}{$T^\text{d}_\text{rec}$, $T_{\text{l}_\text{{pre}}}$ }& \multicolumn{4}{l}{Reconstructed text features, predicted language features} \\  \multicolumn{1}{l}{$T_\text{gt}$, $W_\text{voc}$ }& \multicolumn{4}{l}{Text ground truth one-hot labels, and vocabulary} \\

			\bottomrule\end{tabular}%
	}

\end{table}
\vspace{-1em}

\section{Experiments}\label{sec4}

\begin{table}[t]
	\caption{Cross-forgery generalization on GenFace. ACC and AUC scores (\%) on remaining manipulations, after training using one manipulation. EFS is entire face synthesis. AM is attribute manipulation. FS is face swapping. \label{tab1} \vspace{-0.9em}}
	\setlength{\tabcolsep}{0.85mm}{
		\begin{tabular}{cccccccc}     \\\midrule\multirow{3}[6]{*}{Training Set} & \multirow{3}[6]{*}{Model} & \multicolumn{6}{c}{Testing Set } \\\cmidrule{3-8}      &       & \multicolumn{2}{c}{EFS} & \multicolumn{2}{c}{AM} & \multicolumn{2}{c}{FS} \\\cmidrule{3-8}      &       & ACC   & \cellcolor[gray]{0.9}AUC   & ACC   &  \cellcolor[gray]{0.9}AUC   & ACC   &  \cellcolor[gray]{0.9}AUC \\\midrule\multirow{7}[2]{*}{EFS} & Xception \cite{Xception} & -      &     \cellcolor[gray]{0.9}-  & 50.00  & \cellcolor[gray]{0.9}63.14 & 68.06  & \cellcolor[gray]{0.9}79.52 \\      & ViT \cite{ViT}  &      - &   \cellcolor[gray]{0.9} -   & 54.69 & \cellcolor[gray]{0.9}65.86 & 53.13 & \cellcolor[gray]{0.9}61.43  \\      & CViT \cite{CViT}  &      - &  \cellcolor[gray]{0.9}  -   & 50.02 & \cellcolor[gray]{0.9}63.53 & 72.79  & \cellcolor[gray]{0.9}73.82  \\    & DIRE \cite{DIRE}  &  -     &  \cellcolor[gray]{0.9}\cellcolor[gray]{0.9} -    & 50.03  & \cellcolor[gray]{0.9}76.14 & 74.03  & \cellcolor[gray]{0.9}77.89  \\      & FreqNet \cite{Freq}&     -  &     \cellcolor[gray]{0.9}-  & 50.00  & \cellcolor[gray]{0.9}75.41 & 76.48  & \cellcolor[gray]{0.9}69.62  \\      & CLIP \cite{clip} &    -   &  \cellcolor[gray]{0.9}  -   & 56.05  & \cellcolor[gray]{0.9}64.32  & 53.06  & \cellcolor[gray]{0.9}61.40 
			\\       & MFCLIP \cite{mfclip} &    -   &  \cellcolor[gray]{0.9}  -   & 58.05  & \cellcolor[gray]{0.9}78.76  & 76.88  & \cellcolor[gray]{0.9}81.99
			\\    & FatFormer \cite{FatFormer} &    -   &  \cellcolor[gray]{0.9}  -   & 55.89 & \cellcolor[gray]{0.9}66.90  & 56.21  & \cellcolor[gray]{0.9}64.89 
			\\      
			& VLFFD \cite{VLFFD} &    -   &  \cellcolor[gray]{0.9}  -   & 55.31 & \cellcolor[gray]{0.9}73.89  & 69.54  & \cellcolor[gray]{0.9}70.31
			\\      & DD-VQA \cite{DD-VQA} &    -   &  \cellcolor[gray]{0.9}  -   & 55.96 & \cellcolor[gray]{0.9}74.02 & 70.23  & \cellcolor[gray]{0.9}80.02
			\\     & M2TR \cite{M2TR} & \multicolumn{1}{c}{-}      &  \multicolumn{1}{c}{\cellcolor[gray]{0.9}-}     &50.43  & \cellcolor[gray]{0.9}64.89& 70.98  & \cellcolor[gray]{0.9}80.45  \\  
			&MSCCNet \cite{MSCCNet} & \multicolumn{1}{c}{-}      &  \multicolumn{1}{c}{\cellcolor[gray]{0.9}-}     & 57.30 & \cellcolor[gray]{0.9}75.98 & 75.21 & \cellcolor[gray]{0.9}81.73 \\    & \textbf{MFVLR (Ours)} &     -  &   \cellcolor[gray]{0.9}  -  & \textbf{60.13}  & \cellcolor[gray]{0.9}\textbf{80.72} & \textbf{79.26} & \cellcolor[gray]{0.9}\textbf{84.03} \\\midrule\multirow{7}[2]{*}{AM} & Xception \cite{Xception}& 50.20  & \cellcolor[gray]{0.9}51.45  &      - &   \cellcolor[gray]{0.9}-    & 50.11  & \cellcolor[gray]{0.9}54.57  \\      & ViT \cite{ViT}   & 50.29  & \cellcolor[gray]{0.9}60.37  &   -    &  \cellcolor[gray]{0.9}-    & 50.19  & \cellcolor[gray]{0.9}55.04  \\      & CViT \cite{CViT}  & 50.15  & \cellcolor[gray]{0.9}70.32&    -   &   \cellcolor[gray]{0.9}-    & 50.02  & \cellcolor[gray]{0.9}60.74  \\     & DIRE \cite{DIRE} & 51.14  & \cellcolor[gray]{0.9}72.41  &   -    &    \cellcolor[gray]{0.9}-   & 51.24 & \cellcolor[gray]{0.9}70.45  \\      & FreqNet \cite{Freq}& 50.88  & \cellcolor[gray]{0.9}74.68  &   -    &   \cellcolor[gray]{0.9}\cellcolor[gray]{0.9}-    & 50.33  & \cellcolor[gray]{0.9}76.34  \\     & CLIP \cite{clip}  & 52.48  & \cellcolor[gray]{0.9}60.96 &    -   &   \cellcolor[gray]{0.9}-    & 51.01  & \cellcolor[gray]{0.9}55.21  	\\   & MFCLIP \cite{mfclip}  & 53.29  & \cellcolor[gray]{0.9}87.76 &    -   &   \cellcolor[gray]{0.9}-    & 52.61 & \cellcolor[gray]{0.9}80.31  	\\    & FatFormer \cite{FatFormer}  & 51.67 & \cellcolor[gray]{0.9}63.90 &    -   &   \cellcolor[gray]{0.9}-    & 51.82  & \cellcolor[gray]{0.9}59.03	\\   
			& VLFFD \cite{VLFFD} &   52.06   &  \cellcolor[gray]{0.9}68.98& - & \cellcolor[gray]{0.9}-  & 51.52  & \cellcolor[gray]{0.9}74.40
			\\      & DD-VQA \cite{DD-VQA} &    52.15  &  \cellcolor[gray]{0.9}73.10  & - & \cellcolor[gray]{0.9}- & 52.07  & \cellcolor[gray]{0.9}77.50
			\\   & M2TR \cite{M2TR} & \multicolumn{1}{c}{51.09}      &  \multicolumn{1}{c}{\cellcolor[gray]{0.9}84.29}     &-  & \cellcolor[gray]{0.9}-& 51.17  & \cellcolor[gray]{0.9}78.10  \\  
			&MSCCNet \cite{MSCCNet} & 52.69  & \cellcolor[gray]{0.9}86.95 &   -    &    \cellcolor[gray]{0.9}-   & 51.33 & \cellcolor[gray]{0.9}79.03 \\     & \textbf{MFVLR (Ours)} & \textbf{56.73}&\cellcolor[gray]{0.9}\textbf{89.03}&    -   &   \cellcolor[gray]{0.9}-   & \textbf{54.62}  & \cellcolor[gray]{0.9}\textbf{83.40}  \\\midrule\multirow{7}[2]{*}{FS} & Xception \cite{Xception}& 50.42  & \cellcolor[gray]{0.9}76.48  & 53.75  & \cellcolor[gray]{0.9}75.62 &    -   & \cellcolor[gray]{0.9}- \\      & ViT \cite{ViT}     & 51.09  & \cellcolor[gray]{0.9}69.16  & 52.37  & \cellcolor[gray]{0.9}78.11 &   -    & \cellcolor[gray]{0.9}- \\      & CViT \cite{CViT} & 50.22  & \cellcolor[gray]{0.9}73.88  & 49.98 & \cellcolor[gray]{0.9}73.75&   -    & \cellcolor[gray]{0.9}- \\      & DIRE \cite{DIRE}  & 54.06  & \cellcolor[gray]{0.9}79.65  & 52.13  & \cellcolor[gray]{0.9}78.32  &    -   & \cellcolor[gray]{0.9}- \\      & FreqNet \cite{Freq}& 53.46  & \cellcolor[gray]{0.9}73.68  & 51.97  & \cellcolor[gray]{0.9}74.18  &    -   & \cellcolor[gray]{0.9}- \\      & CLIP \cite{clip} & 52.10  & \cellcolor[gray]{0.9}71.47  & 50.16 & \cellcolor[gray]{0.9}62.27 &   -    & \cellcolor[gray]{0.9}- \\     & MFCLIP \cite{mfclip} & 60.08 & \cellcolor[gray]{0.9}84.93  & 62.58 & \cellcolor[gray]{0.9}80.38 &   -    & \cellcolor[gray]{0.9}- \\   	 & FatFormer \cite{FatFormer} & 54.06  & \cellcolor[gray]{0.9}74.78  & 53.81 & \cellcolor[gray]{0.9}65.89&   -    & \cellcolor[gray]{0.9}- \\     & VLFFD \cite{VLFFD} &   55.38   &  \cellcolor[gray]{0.9}  80.74& 54.60& \cellcolor[gray]{0.9}77.14 & -  & \cellcolor[gray]{0.9}-
			\\      & DD-VQA \cite{DD-VQA} &    56.02  &  \cellcolor[gray]{0.9} 81.10  & 54.79 & \cellcolor[gray]{0.9}76.05& - & \cellcolor[gray]{0.9}-
			\\   & M2TR \cite{M2TR} & \multicolumn{1}{c}{52.08}      &  \multicolumn{1}{c}{\cellcolor[gray]{0.9}78.55}     &55.12  & \cellcolor[gray]{0.9}79.52& -  & \cellcolor[gray]{0.9}-  \\  
			&MSCCNet \cite{MSCCNet} & \multicolumn{1}{c}{58.29}  & \cellcolor[gray]{0.9}82.75  & 56.98 & \cellcolor[gray]{0.9}80.56&   -    & \cellcolor[gray]{0.9}- \\   &\textbf{MFVLR (Ours)} & \textbf{61.13} & \cellcolor[gray]{0.9}\textbf{85.98} & \textbf{64.92} & \cellcolor[gray]{0.9}\textbf{83.73} &     -  & \cellcolor[gray]{0.9}- \\\bottomrule\end{tabular}%
	}
		\vspace{-2em}
\end{table}

\begin{table*}[t!]
	\caption{Cross-diffusion evaluation on GenFace. ACC, AUC, and mIoU scores (\%) on remaining diffusion-based generators, after training using one diffusion-based generator.\label{tabdiff1}}
	\setlength{\tabcolsep}{1.2mm}{
		\begin{tabular}{cccccccccccccccrr}		
	\toprule
	\multirow{3}[5]{*}{Training Set} & \multirow{3}[5]{*}{Model} &
	\multicolumn{15}{c}{\raisebox{0.6ex}{Testing Set }} \\
	\cmidrule(lr){3-17}
	& & \multicolumn{3}{c}{DDPM} & \multicolumn{3}{c}{LatDiff} &
	\multicolumn{3}{c}{CollDiff} & \multicolumn{3}{c}{DiffFace} &
	\multicolumn{3}{c}{Diffae} \\
	\cmidrule(lr){3-5}\cmidrule(lr){6-8}\cmidrule(lr){9-11}\cmidrule(lr){12-14}\cmidrule(lr){15-17}
	& & ACC & AUC & \cellcolor[gray]{0.9}mIoU$\uparrow$
	& ACC & AUC & \cellcolor[gray]{0.9}mIoU$\uparrow$
	& ACC & AUC & \cellcolor[gray]{0.9}mIoU$\uparrow$
	& ACC & AUC & \cellcolor[gray]{0.9}mIoU$\uparrow$
	& ACC & AUC & \cellcolor[gray]{0.9}mIoU$\uparrow$ \\
	\midrule
			\multirow{7}[1]{*}{DDPM} & Xception \cite{Xception}&     -  &   - & \cellcolor[gray]{0.9} -  & 50.00  & 63.57 &\cellcolor[gray]{0.9} - & 50.48  & 75.94 &\cellcolor[gray]{0.9} - & 53.07 & 96.80 &\cellcolor[gray]{0.9} - & \multicolumn{1}{c}{50.07} & \multicolumn{1}{c}{87.49} &\multicolumn{1}{c}{\cellcolor[gray]{0.9}-}\\      & ViT \cite{ViT}  &    -   &   -  & \cellcolor[gray]{0.9}-   & 50.67  & 54.28& \cellcolor[gray]{0.9}- & 52.94  & 74.53 & \cellcolor[gray]{0.9}- & 61.31 & 88.58& \cellcolor[gray]{0.9}- & \multicolumn{1}{c}{50.21} & \multicolumn{1}{c}{49.17}& \multicolumn{1}{c}{\cellcolor[gray]{0.9}-}\\      & CViT \cite{CViT} &   -    &     -&\multicolumn{1}{c}{\cellcolor[gray]{0.9}-}  & 50.00  & 74.51&\multicolumn{1}{c}{\cellcolor[gray]{0.9}-} & 50.07  & 73.24 &\multicolumn{1}{c}{\cellcolor[gray]{0.9}-} & 51.51 & 96.21 &\multicolumn{1}{c}{\cellcolor[gray]{0.9}-} & \multicolumn{1}{c}{50.28} & \multicolumn{1}{c}{92.62 }&\multicolumn{1}{c}{\cellcolor[gray]{0.9}-} \\  
			& DIRE \cite{DIRE} &  -     &  -  &  \cellcolor[gray]{0.9}  -  & 50.18  & 62.50 & \cellcolor[gray]{0.9} - & 50.21  & 69.95 &  \cellcolor[gray]{0.9}- & 56.54 & 94.33  &  \cellcolor[gray]{0.9}-& \multicolumn{1}{c}{51.12} & \multicolumn{1}{c}{88.55 }& \multicolumn{1}{c}{\cellcolor[gray]{0.9}-} \\      & FreqNet \cite{Freq}&   -    &    -  & \multicolumn{1}{c}{\cellcolor[gray]{0.9}-} & 50.00  & 49.09 & \multicolumn{1}{c}{\cellcolor[gray]{0.9}-} & 50.21  & 71.52 & \multicolumn{1}{c}{\cellcolor[gray]{0.9}-} & 50.26 &91.94& \multicolumn{1}{c}{ \cellcolor[gray]{0.9}-} & \multicolumn{1}{c}{53.77} & \multicolumn{1}{c}{91.76}& \multicolumn{1}{c}{\cellcolor[gray]{0.9}-} \\      & CLIP \cite{clip} &    -  &    - & \multicolumn{1}{c}{\cellcolor[gray]{0.9}-}  & 51.79  &57.63& \multicolumn{1}{c}{ \cellcolor[gray]{0.9}-} & 52.28  & 60.95 & \multicolumn{1}{c}{\cellcolor[gray]{0.9}-} & 76.71 &93.23 & \multicolumn{1}{c}{ \cellcolor[gray]{0.9}-}& \multicolumn{1}{c}{49.86} & \multicolumn{1}{c}{45.66}& \multicolumn{1}{c}{\cellcolor[gray]{0.9}-} \\   & MFCLIP \cite{mfclip} &    -  &    - & \multicolumn{1}{c}{\cellcolor[gray]{0.9}-}  & 55.69  &79.80& \multicolumn{1}{c}{ \cellcolor[gray]{0.9}-} & 55.78  & 77.09 & \multicolumn{1}{c}{\cellcolor[gray]{0.9}-} & \textbf{88.89} &\textbf{99.99} & \multicolumn{1}{c}{ \cellcolor[gray]{0.9}-}& \multicolumn{1}{c}{55.94} & \multicolumn{1}{c}{98.76}& \multicolumn{1}{c}{\cellcolor[gray]{0.9}-} \\  
			& FatFormer \cite{FatFormer} &    -  &    - & \multicolumn{1}{c}{\cellcolor[gray]{0.9}-}  & 53.70  & 59.43 &\multicolumn{1}{c}{\cellcolor[gray]{0.9}-} &   54.66& 63.21 & \multicolumn{1}{c}{\cellcolor[gray]{0.9}-} & 78.02 & 94.80 & \multicolumn{1}{c}{\cellcolor[gray]{0.9}-} & \multicolumn{1}{c}{51.65} & \multicolumn{1}{c}{48.27} & \multicolumn{1}{c}{\cellcolor[gray]{0.9}-} \\  
			& VLFFD \cite{VLFFD}& -& -& \multicolumn{1}{c}{\cellcolor[gray]{0.9}-}  &   51.73    &   72.96 & \multicolumn{1}{c}{\cellcolor[gray]{0.9}-}  & 53.70& 74.91& \multicolumn{1}{c}{\cellcolor[gray]{0.9}-} & 77.46& 95.11 & \multicolumn{1}{c}{\cellcolor[gray]{0.9}-} & \multicolumn{1}{c}{52.83} & \multicolumn{1}{c}{88.87} & \multicolumn{1}{c}{\cellcolor[gray]{0.9}-} \\      & DD-VQA\cite{DD-VQA} & -  & - & \multicolumn{1}{c}{\cellcolor[gray]{0.9}-} &    51.45   &    73.99  & \multicolumn{1}{c}{\cellcolor[gray]{0.9}-}  & 53.84 & 76.08 & \multicolumn{1}{c}{\cellcolor[gray]{0.9}-} & 77.93& 96.05& \multicolumn{1}{c}{\cellcolor[gray]{0.9}-} & \multicolumn{1}{c}{54.01} & \multicolumn{1}{c}{89.36}& \multicolumn{1}{c}{\cellcolor[gray]{0.9}-}\\    & C2P-CLIP\cite{C2P-CLIP} & -  & - & \multicolumn{1}{c}{\cellcolor[gray]{0.9}-} &    51.67  &    65.79  & \multicolumn{1}{c}{\cellcolor[gray]{0.9}-}  & 54.38 & 75.09 & \multicolumn{1}{c}{\cellcolor[gray]{0.9}-} & 80.62& 97.59& \multicolumn{1}{c}{\cellcolor[gray]{0.9}-} & \multicolumn{1}{c}{53.62} & \multicolumn{1}{c}{86.97}& \multicolumn{1}{c}{\cellcolor[gray]{0.9}-}\\ 	  & SIDA \cite{SIDA} & -  & - & \multicolumn{1}{c}{\cellcolor[gray]{0.9}-} &    57.64   &    94.37  & \multicolumn{1}{c}{\cellcolor[gray]{0.9}49.26}  & 57.89 & 78.25 & \multicolumn{1}{c}{\cellcolor[gray]{0.9}54.78} & 87.96& \textbf{99.99}& \multicolumn{1}{c}{\cellcolor[gray]{0.9}55.43} & \multicolumn{1}{c}{56.28} & \multicolumn{1}{c}{99.09}& \multicolumn{1}{c}{\cellcolor[gray]{0.9}69.56}\\ 	  & DiffForensics \cite{DiffForensics} & -  & - & \multicolumn{1}{c}{\cellcolor[gray]{0.9}-} &    57.28   &    72.96  & \multicolumn{1}{c}{\cellcolor[gray]{0.9}56.99}  & 53.84 & 80.65 & \multicolumn{1}{c}{\cellcolor[gray]{0.9}55.26} & 76.31& 94.38& \multicolumn{1}{c}{\cellcolor[gray]{0.9}56.89} & \multicolumn{1}{c}{58.97} & \multicolumn{1}{c}{98.95}& \multicolumn{1}{c}{\cellcolor[gray]{0.9}70.42}\\ 		& M2TR \cite{M2TR} & \multicolumn{1}{c}{-}      &  \multicolumn{1}{c}{-}  &  \multicolumn{1}{c}{\cellcolor[gray]{0.9}-}    & 50.02  & 40.18 &  \multicolumn{1}{c}{\cellcolor[gray]{0.9}39.97}  & 50.04  & 79.17 &  \multicolumn{1}{c}{\cellcolor[gray]{0.9}49.91} & 50.24  & 44.60 &  \multicolumn{1}{c}{\cellcolor[gray]{0.9}42.16} & 50.39  &\multicolumn{1}{c}{60.44} & \multicolumn{1}{c}{\cellcolor[gray]{0.9}59.17}
			\\  
			& HiFi-Net \cite{HiFi-Net} & \multicolumn{1}{c}{-}      &  \multicolumn{1}{c}{-}   &  \multicolumn{1}{c}{\cellcolor[gray]{0.9}-}    & 50.11  & 59.43  &  \multicolumn{1}{c}{\cellcolor[gray]{0.9}42.31} & 52.54  & 79.43&  \multicolumn{1}{c}{\cellcolor[gray]{0.9}53.49} & 68.90  & 82.45 &  \multicolumn{1}{c}{\cellcolor[gray]{0.9}53.28}& 54.52  & \multicolumn{1}{c}{83.90} &  \multicolumn{1}{c}{\cellcolor[gray]{0.9}70.99}\\
			&MSCCNet \cite{MSCCNet} & \multicolumn{1}{c}{-}      &  \multicolumn{1}{c}{-}  & \multicolumn{1}{c}{\cellcolor[gray]{0.9}-}    & 52.87  & 65.76& \multicolumn{1}{c}{\cellcolor[gray]{0.9}47.52}  & 53.67 & \textbf{82.56} & \cellcolor[gray]{0.9}56.09 & 59.78 & 87.46& \cellcolor[gray]{0.9}55.77 & 53.98  & \multicolumn{1}{c}{85.43} &\multicolumn{1}{c}{ \cellcolor[gray]{0.9}71.95} \\
			& \textbf{MFVLR (Ours)} &  -     &    - & \multicolumn{1}{c}{\cellcolor[gray]{0.9}-} &  \textbf{60.32} & \textbf{98.41}  & \multicolumn{1}{c}{\cellcolor[gray]{0.9}\textbf{59.34}} &\textbf{55.09} & 78.04& \multicolumn{1}{c}{\cellcolor[gray]{0.9}\textbf{58.69}}& 82.65 & 99.98& \multicolumn{1}{c}{\cellcolor[gray]{0.9}\textbf{58.49}} & \multicolumn{1}{c}{\textbf{69.88}} & \multicolumn{1}{c}{\textbf{99.90}}& \multicolumn{1}{c}{\cellcolor[gray]{0.9}\textbf{73.87}}
			\\\midrule	
			\multirow{7}[1]{*}{LatDiff} & Xception \cite{Xception}& 50.13  & 76.45& \multicolumn{1}{c}{\cellcolor[gray]{0.9}-}  &   -    &   -  & \multicolumn{1}{c}{\cellcolor[gray]{0.9}-}  & 50.00  & 37.05& \multicolumn{1}{c}{\cellcolor[gray]{0.9}-}  & 51.51 & 96.21& \multicolumn{1}{c}{\cellcolor[gray]{0.9}-} & \multicolumn{1}{c}{50.28 } & \multicolumn{1}{c}{92.62}& \multicolumn{1}{c}{\cellcolor[gray]{0.9}-} \\      & ViT \cite{ViT}  & 66.60  &80.74 & \multicolumn{1}{c}{ \cellcolor[gray]{0.9}-} &   -    &  -  & \multicolumn{1}{c}{ \cellcolor[gray]{0.9}-}  & 53.20  & 54.09 & \multicolumn{1}{c}{\cellcolor[gray]{0.9}-} & 52.93 &57.59 & \multicolumn{1}{c}{ \cellcolor[gray]{0.9}-}& \multicolumn{1}{c}{57.60 } & \multicolumn{1}{c}{65.33}& \multicolumn{1}{c}{\cellcolor[gray]{0.9}-} \\      & CViT \cite{CViT}  & 51.73  & 90.76 & \multicolumn{1}{c}{\cellcolor[gray]{0.9}-} &  -     &   -  & \multicolumn{1}{c}{\cellcolor[gray]{0.9}-}  & 50.00  &52.36& \multicolumn{1}{c}{ \cellcolor[gray]{0.9}-}  & 50.01 & 44.50 & \multicolumn{1}{c}{\cellcolor[gray]{0.9}-} & \multicolumn{1}{c}{47.88} & \multicolumn{1}{c}{44.04} & \multicolumn{1}{c}{\cellcolor[gray]{0.9}-}\\   
			& DIRE \cite{DIRE}& 50.01  & 46.29 & \multicolumn{1}{c}{\cellcolor[gray]{0.9}-} &      - &    -  & \multicolumn{1}{c}{\cellcolor[gray]{0.9}-} & 50.02  & 43.13 & \multicolumn{1}{c}{\cellcolor[gray]{0.9}-} & 50.01  & 48.51 & \multicolumn{1}{c}{\cellcolor[gray]{0.9}-}& \multicolumn{1}{c}{50.05} & \multicolumn{1}{c}{45.98} & \multicolumn{1}{c}{\cellcolor[gray]{0.9}-}\\      & FreqNet \cite{Freq}& 78.50  & 83.32 & \multicolumn{1}{c}{\cellcolor[gray]{0.9}-} &   -    &  - & \multicolumn{1}{c}{ \cellcolor[gray]{0.9}-}   & \textbf{79.04}  & 89.58 & \multicolumn{1}{c}{\cellcolor[gray]{0.9}-} & 50.18 &89.93& \multicolumn{1}{c}{ \cellcolor[gray]{0.9}-} & \multicolumn{1}{c}{38.16} & \multicolumn{1}{c}{37.87} & \multicolumn{1}{c}{\cellcolor[gray]{0.9}-}\\      & CLIP \cite{clip} & 62.33  & 74.53 & \multicolumn{1}{c}{\cellcolor[gray]{0.9}-} &    -   &    - & \multicolumn{1}{c}{\cellcolor[gray]{0.9}-}  & 52.99  & 53.23 & \multicolumn{1}{c}{\cellcolor[gray]{0.9}-} & 55.71 & 61.05& \multicolumn{1}{c}{\cellcolor[gray]{0.9}-} & \multicolumn{1}{c}{56.64} & \multicolumn{1}{c}{64.38}& \multicolumn{1}{c}{\cellcolor[gray]{0.9}-} \\       & MFCLIP \cite{mfclip} & 99.99  & 99.99 & \multicolumn{1}{c}{\cellcolor[gray]{0.9}-} &    -   &    - & \multicolumn{1}{c}{\cellcolor[gray]{0.9}-}  & 65.08  &77.07 & \multicolumn{1}{c}{\cellcolor[gray]{0.9}-} & 99.98& 99.98& \multicolumn{1}{c}{\cellcolor[gray]{0.9}-} & \multicolumn{1}{c}{97.92} & \multicolumn{1}{c}{99.99}& \multicolumn{1}{c}{\cellcolor[gray]{0.9}-} \\  
			& FatFormer \cite{FatFormer} & 63.99 & 76.26 & \multicolumn{1}{c}{\cellcolor[gray]{0.9}-} &    -   &   - & \multicolumn{1}{c}{ \cellcolor[gray]{0.9}-}  & 54.90  & 56.14& \multicolumn{1}{c}{\cellcolor[gray]{0.9}-} & 57.89 & 63.04& \multicolumn{1}{c}{\cellcolor[gray]{0.9}-} & \multicolumn{1}{c}{58.92} & \multicolumn{1}{c}{66.03}& \multicolumn{1}{c}{\cellcolor[gray]{0.9}-} \\    
			& VLFFD \cite{VLFFD}& 87.35  & 92.64 & \multicolumn{1}{c}{\cellcolor[gray]{0.9}-} &   -    &   - & \multicolumn{1}{c}{\cellcolor[gray]{0.9}-}   & 56.32 & 67.81& \multicolumn{1}{c}{\cellcolor[gray]{0.9}-} & 87.24 & 89.25& \multicolumn{1}{c}{\cellcolor[gray]{0.9}-} & \multicolumn{1}{c}{68.53} & \multicolumn{1}{c}{93.02} & \multicolumn{1}{c}{\cellcolor[gray]{0.9}-}\\      & DD-VQA \cite{DD-VQA} & 88.46  & 94.00 & \multicolumn{1}{c}{\cellcolor[gray]{0.9}-} &    -   &    - & \multicolumn{1}{c}{\cellcolor[gray]{0.9}-}  & 57.84  & 73.02 & \multicolumn{1}{c}{\cellcolor[gray]{0.9}-} & 88.61 & 97.05& \multicolumn{1}{c}{\cellcolor[gray]{0.9}-} & \multicolumn{1}{c}{70.56} & \multicolumn{1}{c}{93.70}& \multicolumn{1}{c}{\cellcolor[gray]{0.9}-}\\      & C2P-CLIP  \cite{C2P-CLIP} & 94.37  & 99.86 & \multicolumn{1}{c}{\cellcolor[gray]{0.9}-} &    \multicolumn{1}{c}{-}   &   \multicolumn{1}{c}{-} & \multicolumn{1}{c}{\cellcolor[gray]{0.9}-}  &66.89 & 79.84 & \multicolumn{1}{c}{\cellcolor[gray]{0.9}-} & 89.47& 98.16& \multicolumn{1}{c}{\cellcolor[gray]{0.9}-} & \multicolumn{1}{c}{73.95} & \multicolumn{1}{c}{96.87}& \multicolumn{1}{c}{\cellcolor[gray]{0.9}-}\\ 	  & SIDA \cite{SIDA} & 92.43  & 96.87 & \multicolumn{1}{c}{\cellcolor[gray]{0.9}69.66} &    \multicolumn{1}{c}{-}  &    \multicolumn{1}{c}{-}  & \multicolumn{1}{c}{\cellcolor[gray]{0.9}-}  & 55.80 & 80.74 & \multicolumn{1}{c}{\cellcolor[gray]{0.9}52.37} & 90.84& 97.19& \multicolumn{1}{c}{\cellcolor[gray]{0.9}69.45} & \multicolumn{1}{c}{76.50} & \multicolumn{1}{c}{95.83}& \multicolumn{1}{c}{\cellcolor[gray]{0.9}71.79}\\ 	  & DiffForensics \cite{DiffForensics} & 91.56 & 97.39& \multicolumn{1}{c}{\cellcolor[gray]{0.9}70.83} &   \multicolumn{1}{c}{-} &   \multicolumn{1}{c}{-} & \multicolumn{1}{c}{\cellcolor[gray]{0.9}-}  & 56.71 & 78.49 & \multicolumn{1}{c}{\cellcolor[gray]{0.9}53.08} & 87.56& 97.59& \multicolumn{1}{c}{\cellcolor[gray]{0.9}70.63} & \multicolumn{1}{c}{75.98} & \multicolumn{1}{c}{94.15}& \multicolumn{1}{c}{\cellcolor[gray]{0.9}72.53}\\	& M2TR \cite{M2TR} & 86.04    & 99.88 & \multicolumn{1}{c}{\cellcolor[gray]{0.9}67.92} & \multicolumn{1}{c}{-}  &\multicolumn{1}{c}{-} & \multicolumn{1}{c}{\cellcolor[gray]{0.9}-} & 50.02  &56.42& \multicolumn{1}{c}{\cellcolor[gray]{0.9}49.81}& 50.03  & 66.25& \multicolumn{1}{c}{\cellcolor[gray]{0.9}62.14} & 53.56 &\multicolumn{1}{c}{89.89}& \multicolumn{1}{c}{\cellcolor[gray]{0.9}69.20} \\  
			& HiFi-Net \cite{HiFi-Net} & \multicolumn{1}{c}{89.08}      &  99.89  & \multicolumn{1}{c}{\cellcolor[gray]{0.9}68.13}    & -  & - & \multicolumn{1}{c}{\cellcolor[gray]{0.9}-} & 51.49  & 50.76& \multicolumn{1}{c}{\cellcolor[gray]{0.9}47.10}  & 52.08  & 68.79 & \multicolumn{1}{c}{\cellcolor[gray]{0.9}64.59} & 57.65  &91.05 & \multicolumn{1}{c}{ \cellcolor[gray]{0.9}71.26}\\
			&MSCCNet \cite{MSCCNet} & \multicolumn{1}{c}{90.23}      &   \multicolumn{1}{c}{ 99.52}  & \multicolumn{1}{c}{\cellcolor[gray]{0.9}65.48}     & -  & \multicolumn{1}{c}{-}& \multicolumn{1}{c}{\cellcolor[gray]{0.9}-}   & 53.34  & 63.60 & \multicolumn{1}{c}{\cellcolor[gray]{0.9}51.26}  & 70.22  & 82.54& \multicolumn{1}{c}{\cellcolor[gray]{0.9}68.17}   & 58.90  & 92.43 & \multicolumn{1}{c}{\cellcolor[gray]{0.9}68.17} \\
			& \textbf{MFVLR (Ours)} & \textbf{100} & \textbf{100} &   \cellcolor[gray]{0.9}\textbf{74.89} &      - &  -    &   \cellcolor[gray]{0.9}-   & 56.34  & \textbf{99.50} & \multicolumn{1}{c}{\cellcolor[gray]{0.9}\textbf{55.71}} & \textbf{100} & \textbf{100}& \multicolumn{1}{c}{\cellcolor[gray]{0.9}\textbf{73.20}} & \multicolumn{1}{c}{\textbf{99.91}} & \multicolumn{1}{c}{\textbf{100}} & \multicolumn{1}{c}{\cellcolor[gray]{0.9}\textbf{75.63} }
\\\bottomrule\end{tabular}%
}
\vspace{-2em}
\end{table*}

\begin{table*}[t!]
	\caption{Cross-diffusion evaluation on GenFace. ACC, AUC, and mIoU scores (\%) on remaining diffusion-based generators, after training using one diffusion-based generator.\label{tabdiff2}}
	\setlength{\tabcolsep}{1.2mm}{
		\begin{tabular}{cccccccccccccccrr}		
\toprule
\multirow{3}[5]{*}{Training Set} & \multirow{3}[5]{*}{Model} &
\multicolumn{15}{c}{\raisebox{0.6ex}{Testing Set }} \\
\cmidrule(lr){3-17}
& & \multicolumn{3}{c}{DDPM} & \multicolumn{3}{c}{LatDiff} &
\multicolumn{3}{c}{CollDiff} & \multicolumn{3}{c}{DiffFace} &
\multicolumn{3}{c}{Diffae} \\
\cmidrule(lr){3-5}\cmidrule(lr){6-8}\cmidrule(lr){9-11}\cmidrule(lr){12-14}\cmidrule(lr){15-17}
& & ACC & AUC & \cellcolor[gray]{0.9}mIoU$\uparrow$
& ACC & AUC & \cellcolor[gray]{0.9}mIoU$\uparrow$
& ACC & AUC & \cellcolor[gray]{0.9}mIoU$\uparrow$
& ACC & AUC & \cellcolor[gray]{0.9}mIoU$\uparrow$
& ACC & AUC & \cellcolor[gray]{0.9}mIoU$\uparrow$ \\
\midrule
			
			\multirow{7}[2]{*}{CollDiff} & Xception \cite{Xception}& 55.69  & 96.31 & \multicolumn{1}{c}{\cellcolor[gray]{0.9}-} & 49.98  & 70.45& \multicolumn{1}{c}{\cellcolor[gray]{0.9}-} &     -  &   -  & \multicolumn{1}{c}{\cellcolor[gray]{0.9} -} & 50.17 & 71.98& \multicolumn{1}{c}{\cellcolor[gray]{0.9}-} & \multicolumn{1}{c}{50.19} & \multicolumn{1}{c}{59.50 }& \multicolumn{1}{c}{\cellcolor[gray]{0.9}-} \\      & ViT \cite{ViT}   & 55.06  & 61.83& \multicolumn{1}{c}{\cellcolor[gray]{0.9}-}  & 51.03  &48.64& \multicolumn{1}{c}{ \cellcolor[gray]{0.9}-} &   -    &    -  & \multicolumn{1}{c}{\cellcolor[gray]{0.9}-} & 50.26 & 50.32 & \multicolumn{1}{c}{\cellcolor[gray]{0.9}-}& \multicolumn{1}{c}{50.51} & \multicolumn{1}{c}{49.97}& \multicolumn{1}{c}{\cellcolor[gray]{0.9}-} \\      & CViT \cite{CViT}  & 99.74  &99.97& \multicolumn{1}{c}{ \cellcolor[gray]{0.9}-}  & 49.98  & 47.59 & \multicolumn{1}{c}{\cellcolor[gray]{0.9}-}&   -    &   -   & \multicolumn{1}{c}{\cellcolor[gray]{0.9}-} & 81.97 & 98.80 & \multicolumn{1}{c}{\cellcolor[gray]{0.9}-} & \multicolumn{1}{c}{90.56} & \multicolumn{1}{c}{99.83}& \multicolumn{1}{c}{\cellcolor[gray]{0.9}-} \\ 
			& DIRE \cite{DIRE} & 91.52  & 81.05 & \multicolumn{1}{c}{\cellcolor[gray]{0.9}-} & 50.02  & 65.99 & \multicolumn{1}{c}{\cellcolor[gray]{0.9}-} &    -   &     -  & \multicolumn{1}{c}{\cellcolor[gray]{0.9}-}& 60.79 & 93.68 & \multicolumn{1}{c}{\cellcolor[gray]{0.9}-}& \multicolumn{1}{c}{56.85} & \multicolumn{1}{c}{96.86} & \multicolumn{1}{c}{\cellcolor[gray]{0.9}-}\\      & FreqNet \cite{Freq}& 93.48  & 85.42& \multicolumn{1}{c}{\cellcolor[gray]{0.9}-}  & 49.91  & 55.59 & \multicolumn{1}{c}{\cellcolor[gray]{0.9}-} &   -    &    -  & \multicolumn{1}{c}{\cellcolor[gray]{0.9}-} & 50.04 & 61.18& \multicolumn{1}{c}{\cellcolor[gray]{0.9}-} & \multicolumn{1}{c}{99.95} & \multicolumn{1}{c}{99.98}& \multicolumn{1}{c}{\cellcolor[gray]{0.9}-} \\      & CLIP \cite{clip} & 51.58  & 53.92 & \multicolumn{1}{c}{\cellcolor[gray]{0.9}-} & 50.21  & 46.17& \multicolumn{1}{c}{\cellcolor[gray]{0.9}-} &     -  &    - & \multicolumn{1}{c}{\cellcolor[gray]{0.9}-}  & 50.26 & 48.64& \multicolumn{1}{c}{\cellcolor[gray]{0.9}-} & \multicolumn{1}{c}{49.82} & \multicolumn{1}{c}{48.36}& \multicolumn{1}{c}{\cellcolor[gray]{0.9}-} \\         & MFCLIP \cite{mfclip} & \textbf{100}  & \textbf{100} & \multicolumn{1}{c}{\cellcolor[gray]{0.9}-} & \textbf{99.19} & 99.97& \multicolumn{1}{c}{\cellcolor[gray]{0.9}-} &     -  &    - & \multicolumn{1}{c}{\cellcolor[gray]{0.9}-}  &\textbf{99.63} & 99.96& \multicolumn{1}{c}{\cellcolor[gray]{0.9}-} & \multicolumn{1}{c}{93.94} & \multicolumn{1}{c}{\textbf{99.99}}& \multicolumn{1}{c}{\cellcolor[gray]{0.9}-} \\      
			& FatFormer \cite{FatFormer} & 53.81  & 55.52& \multicolumn{1}{c}{\cellcolor[gray]{0.9}-}  & 53.90 & 49.03 & \multicolumn{1}{c}{\cellcolor[gray]{0.9}-}&     -  &    - & \multicolumn{1}{c}{\cellcolor[gray]{0.9}-}  & 54.80 & 52.71& \multicolumn{1}{c}{\cellcolor[gray]{0.9}-} & \multicolumn{1}{c}{52.90} & \multicolumn{1}{c}{50.65}& \multicolumn{1}{c}{\cellcolor[gray]{0.9}-} \\      
			& VLFFD \cite{VLFFD}& 94.25  & 96.01 & \multicolumn{1}{c}{\cellcolor[gray]{0.9}-} &  56.89   &   64.06  & \multicolumn{1}{c}{\cellcolor[gray]{0.9}-} & -  & -& \multicolumn{1}{c}{\cellcolor[gray]{0.9}-} & 82.29 &94.67& \multicolumn{1}{c}{ \cellcolor[gray]{0.9}-} & \multicolumn{1}{c}{72.43} & \multicolumn{1}{c}{92.99}& \multicolumn{1}{c}{\cellcolor[gray]{0.9}-} \\      & DD-VQA \cite{DD-VQA} & 93.20  & 97.44 & \multicolumn{1}{c}{\cellcolor[gray]{0.9}-} &   55.60   &    70.76  & \multicolumn{1}{c}{\cellcolor[gray]{0.9}-} & -  & - & \multicolumn{1}{c}{\cellcolor[gray]{0.9}-} & 83.77 & 91.59& \multicolumn{1}{c}{\cellcolor[gray]{0.9}-}& \multicolumn{1}{c}{73.08} & \multicolumn{1}{c}{92.94}& \multicolumn{1}{c}{\cellcolor[gray]{0.9}-}\\     & C2P-CLIP \cite{C2P-CLIP} & 94.86 &97.05 & \multicolumn{1}{c}{\cellcolor[gray]{0.9}-} &    57.09   &    79.86  & \multicolumn{1}{c}{\cellcolor[gray]{0.9}-}  & \multicolumn{1}{c}{-}  & \multicolumn{1}{c}{-}  & \multicolumn{1}{c}{\cellcolor[gray]{0.9}-} & 86.94& 95.83& \multicolumn{1}{c}{\cellcolor[gray]{0.9}-} & \multicolumn{1}{c}{76.97} & \multicolumn{1}{c}{93.85}& \multicolumn{1}{c}{\cellcolor[gray]{0.9}-}\\ 	  & SIDA \cite{SIDA} & 95.80  & 98.53 & \multicolumn{1}{c}{\cellcolor[gray]{0.9}96.76} &    59.31  &   86.78  & \multicolumn{1}{c}{\cellcolor[gray]{0.9}90.59}  &  \multicolumn{1}{c}{-}   &  \multicolumn{1}{c}{-}   & \multicolumn{1}{c}{\cellcolor[gray]{0.9}-} & 89.57& 97.86& \multicolumn{1}{c}{\cellcolor[gray]{0.9}69.50} & \multicolumn{1}{c}{79.59} & \multicolumn{1}{c}{94.67}& \multicolumn{1}{c}{\cellcolor[gray]{0.9}69.86}\\ 	  & DiffForensics \cite{DiffForensics} & 96.78  & 99.06 & \multicolumn{1}{c}{\cellcolor[gray]{0.9}97.82} &    95.46   &    96.59  & \multicolumn{1}{c}{\cellcolor[gray]{0.9}89.59}  &  \multicolumn{1}{c}{-} &  \multicolumn{1}{c}{-} & \multicolumn{1}{c}{\cellcolor[gray]{0.9}-} & 90.56& 98.97& \multicolumn{1}{c}{\cellcolor[gray]{0.9}70.43} & \multicolumn{1}{c}{80.67} & \multicolumn{1}{c}{95.69}& \multicolumn{1}{c}{\cellcolor[gray]{0.9}70.53}\\	& M2TR \cite{M2TR} & \multicolumn{1}{c}{99.84}      &  \multicolumn{1}{c}{99.99} &  \multicolumn{1}{c}{\cellcolor[gray]{0.9}97.23}    & 92.37  & 99.93 &  \multicolumn{1}{c}{\cellcolor[gray]{0.9}88.98}  &  \multicolumn{1}{c}{-}  & \multicolumn{1}{c}{-} & \multicolumn{1}{c}{\cellcolor[gray]{0.9}-} & 54.03  & 97.10 & \cellcolor[gray]{0.9}65.58  & 90.40 &99.94&  \multicolumn{1}{c}{\cellcolor[gray]{0.9}54.75} \\  
			& HiFi-Net \cite{HiFi-Net} & \multicolumn{1}{c}{95.32}      &  \multicolumn{1}{c}{99.45} & \multicolumn{1}{c}{\cellcolor[gray]{0.9}94.59}     & 92.44  & 98.90& \multicolumn{1}{c}{\cellcolor[gray]{0.9}87.54} & -  & -& \multicolumn{1}{c}{\cellcolor[gray]{0.9}-}& 56.86  & 98.31& \multicolumn{1}{c}{\cellcolor[gray]{0.9}67.50} &\multicolumn{1}{c}{93.11} & 99.95 & \multicolumn{1}{c}{\cellcolor[gray]{0.9}60.98}\\
			&MSCCNet \cite{MSCCNet} & \multicolumn{1}{c}{99.90}      &  \multicolumn{1}{c}{99.99}    & \multicolumn{1}{c}{\cellcolor[gray]{0.9}97.96} & 93.45  & 99.02 & \multicolumn{1}{c}{\cellcolor[gray]{0.9}89.57}& -  & - & \cellcolor[gray]{0.9}-& 60.32  & 99.02 & \multicolumn{1}{c}{\cellcolor[gray]{0.9}68.56} & 94.52  &99.30 & \multicolumn{1}{c}{ \cellcolor[gray]{0.9}68.07} \\ 
			& \textbf{MFVLR (Ours)} & \textbf{100}& \textbf{\textbf{100}} & \multicolumn{1}{c}{\cellcolor[gray]{0.9}\textbf{98.56}} & 96.70 & \textbf{99.99}&\multicolumn{1}{c}{\cellcolor[gray]{0.9}\textbf{91.37}} &     -  &    -  & \multicolumn{1}{c}{\cellcolor[gray]{0.9}-}  & 92.64 & \textbf{99.98}& \multicolumn{1}{c}{\cellcolor[gray]{0.9}\textbf{72.94}}  & \multicolumn{1}{c}{\textbf{99.96}}& \multicolumn{1}{c}{\textbf{99.99}}& \multicolumn{1}{c}{\cellcolor[gray]{0.9}\textbf{71.32} } \\\midrule\multirow{7}[2]{*}{DiffFace} & Xception \cite{Xception}& 99.98  & 99.98 & \multicolumn{1}{c}{\cellcolor[gray]{0.9}-} & 50.00  & 48.31& \multicolumn{1}{c}{\cellcolor[gray]{0.9}-} & 50.00  & 52.97 & \multicolumn{1}{c}{\cellcolor[gray]{0.9}-} &   -    &    -  & \multicolumn{1}{c}{\cellcolor[gray]{0.9}-} & \multicolumn{1}{c}{50.07} & \multicolumn{1}{c}{87.49} & \multicolumn{1}{c}{\cellcolor[gray]{0.9}-}\\      & ViT \cite{ViT}    & 87.00  & 97.01& \multicolumn{1}{c}{\cellcolor[gray]{0.9}-}  & 49.11  & 42.97& \multicolumn{1}{c}{\cellcolor[gray]{0.9}-} & 49.82  & 49.00 & \multicolumn{1}{c}{\cellcolor[gray]{0.9}-} &    -   &    -  & \multicolumn{1}{c}{\cellcolor[gray]{0.9}-} & \multicolumn{1}{c}{48.74} & \multicolumn{1}{c}{40.90 }& \multicolumn{1}{c}{\cellcolor[gray]{0.9}-} \\      & CViT \cite{CViT}  & 50.08  & 77.27& \multicolumn{1}{c}{\cellcolor[gray]{0.9}-}  & 49.98  & 54.22 & \multicolumn{1}{c}{\cellcolor[gray]{0.9}-}& 50.04  & 59.65 & \multicolumn{1}{c}{\cellcolor[gray]{0.9}-} &     -  &  -    & \multicolumn{1}{c}{\cellcolor[gray]{0.9}-} & \multicolumn{1}{c}{47.52} & \multicolumn{1}{c}{40.44} & \multicolumn{1}{c}{\cellcolor[gray]{0.9}-}\\     & DIRE \cite{DIRE} & 50.04  & 75.32 & \multicolumn{1}{c}{\cellcolor[gray]{0.9}-} & 50.00  & 47.21 & \multicolumn{1}{c}{\cellcolor[gray]{0.9}-} & 59.79  & 96.91& \multicolumn{1}{c}{\cellcolor[gray]{0.9}-} &    -   &    -   & \multicolumn{1}{c}{\cellcolor[gray]{0.9}-}& \multicolumn{1}{c}{\textbf{73.55}} & \multicolumn{1}{c}{96.53}& \multicolumn{1}{c}{\cellcolor[gray]{0.9}-} \\      & FreqNet \cite{Freq}& 49.85  & 82.21 & \multicolumn{1}{c}{\cellcolor[gray]{0.9}-} & 51.77  & 73.37 & \multicolumn{1}{c}{\cellcolor[gray]{0.9}-} & 49.70  & 75.62& \multicolumn{1}{c}{\cellcolor[gray]{0.9}-}  &      - &   -  & \multicolumn{1}{c}{\cellcolor[gray]{0.9} -} & \multicolumn{1}{c}{43.43} & \multicolumn{1}{c}{58.69}& \multicolumn{1}{c}{\cellcolor[gray]{0.9}-} \\      & CLIP \cite{clip} & 76.34  & 91.74 & \multicolumn{1}{c}{\cellcolor[gray]{0.9}-} & 49.98  & 50.59& \multicolumn{1}{c}{\cellcolor[gray]{0.9}-} & 51.56  & 57.34 & \multicolumn{1}{c}{\cellcolor[gray]{0.9}-} &    -   &   -  & \multicolumn{1}{c}{\cellcolor[gray]{0.9} -} & \multicolumn{1}{c}{55.90 } & \multicolumn{1}{c}{51.90 }& \multicolumn{1}{c}{\cellcolor[gray]{0.9}-} \\        & MFCLIP \cite{mfclip} & \textbf{99.99} & \textbf{99.99} & \multicolumn{1}{c}{\cellcolor[gray]{0.9}-} & 85.32  & \textbf{99.94}& \multicolumn{1}{c}{\cellcolor[gray]{0.9}-} & 50.57  & 75.40 & \multicolumn{1}{c}{\cellcolor[gray]{0.9}-} &    -   &   -  & \multicolumn{1}{c}{\cellcolor[gray]{0.9} -} & \multicolumn{1}{c}{52.12 } & \multicolumn{1}{c}{\textbf{99.92 }}& \multicolumn{1}{c}{\cellcolor[gray]{0.9}-} \\     
			& FatFormer \cite{FatFormer} & 79.54  & 94.20 & \multicolumn{1}{c}{\cellcolor[gray]{0.9}-} & 52.89  & 53.94 & \multicolumn{1}{c}{\cellcolor[gray]{0.9}-}& 53.76  & 59.02& \multicolumn{1}{c}{\cellcolor[gray]{0.9}-}  &    -   &   -  & \multicolumn{1}{c}{\cellcolor[gray]{0.9} -} & \multicolumn{1}{c}{57.89 } & \multicolumn{1}{c}{54.64} & \multicolumn{1}{c}{\cellcolor[gray]{0.9}-}\\     
			& VLFFD \cite{VLFFD}& 77.84 &94.76& \multicolumn{1}{c}{ \cellcolor[gray]{0.9}-}  &   62.33    &   84.97 & \multicolumn{1}{c}{\cellcolor[gray]{0.9}-}  & 53.66  & 60.23& \multicolumn{1}{c}{\cellcolor[gray]{0.9}-}  & - & - & \multicolumn{1}{c}{\cellcolor[gray]{0.9}-}& \multicolumn{1}{c}{58.08} & \multicolumn{1}{c}{87.96} & \multicolumn{1}{c}{\cellcolor[gray]{0.9}-}\\      & DD-VQA \cite{DD-VQA} & 90.22  & 99.98& \multicolumn{1}{c}{\cellcolor[gray]{0.9}-} &    63.75  &    83.88& \multicolumn{1}{c}{\cellcolor[gray]{0.9}-}  & 54.07 & 59.01& \multicolumn{1}{c}{\cellcolor[gray]{0.9}-} & - & - & \multicolumn{1}{c}{\cellcolor[gray]{0.9}-}& \multicolumn{1}{c}{59.89} & \multicolumn{1}{c}{88.40}& \multicolumn{1}{c}{\cellcolor[gray]{0.9}-}\\      & C2P-CLIP \cite{C2P-CLIP} &77.86  & 95.89 & \multicolumn{1}{c}{\cellcolor[gray]{0.9}-} &   83.45  &  85.74  & \multicolumn{1}{c}{\cellcolor[gray]{0.9}-}  & 55.67 & 72.49 & \multicolumn{1}{c}{\cellcolor[gray]{0.9}-} & \multicolumn{1}{c}{- }&  \multicolumn{1}{c}{- }& \multicolumn{1}{c}{\cellcolor[gray]{0.9}-} & \multicolumn{1}{c}{58.97} & \multicolumn{1}{c}{92.66}& \multicolumn{1}{c}{\cellcolor[gray]{0.9}-}\\ 	  & SIDA \cite{SIDA} & 91.48 & 98.56 & \multicolumn{1}{c}{\cellcolor[gray]{0.9}59.77} &    83.69   &    86.04  & \multicolumn{1}{c}{\cellcolor[gray]{0.9}58.94}  & 56.71 & 75.62 & \multicolumn{1}{c}{\cellcolor[gray]{0.9}56.15} & \multicolumn{1}{c}{-}&\multicolumn{1}{c}{-}& \multicolumn{1}{c}{\cellcolor[gray]{0.9}-} & \multicolumn{1}{c}{57.65} & \multicolumn{1}{c}{90.25}& \multicolumn{1}{c}{\cellcolor[gray]{0.9}73.04}\\ 	  & DiffForensics \cite{DiffForensics} & 90.87 & 97.53& \multicolumn{1}{c}{\cellcolor[gray]{0.9}60.38} &    80.72   &    87.49  & \multicolumn{1}{c}{\cellcolor[gray]{0.9}52.81}  & 58.96 & 77.59 & \multicolumn{1}{c}{\cellcolor[gray]{0.9}55.64} &\multicolumn{1}{c}{-}& \multicolumn{1}{c}{-}& \multicolumn{1}{c}{\cellcolor[gray]{0.9}-} & \multicolumn{1}{c}{56.89} & \multicolumn{1}{c}{97.96}& \multicolumn{1}{c}{\cellcolor[gray]{0.9}70.36}\\	& M2TR \cite{M2TR} & \multicolumn{1}{c}{50.52}      &  \multicolumn{1}{c}{90.64}  & \cellcolor[gray]{0.9}49.99    & 50.00  & 89.83  & \cellcolor[gray]{0.9}49.98 & 50.00  & 69.36& \cellcolor[gray]{0.9}49.98 & \multicolumn{1}{c}{-}  & \multicolumn{1}{c}{-}& \multicolumn{1}{c}{\cellcolor[gray]{0.9}-} & 50.04  &89.90 & \multicolumn{1}{c}{\cellcolor[gray]{0.9}69.18} \\  
			& HiFi-Net \cite{HiFi-Net} & \multicolumn{1}{c}{53.87}      &  \multicolumn{1}{c}{92.42} & \multicolumn{1}{c}{\cellcolor[gray]{0.9}51.67}    & 53.76  & 88.55& \multicolumn{1}{c}{\cellcolor[gray]{0.9}51.93} & 53.08 & 74.87& \multicolumn{1}{c}{\cellcolor[gray]{0.9}52.76} & \multicolumn{1}{c}{-}  & -  & \cellcolor[gray]{0.9}- & 53.61  & 92.04 & \multicolumn{1}{c}{\cellcolor[gray]{0.9}71.89} \\
			&MSCCNet \cite{MSCCNet} & \multicolumn{1}{c}{54.88}      &  \multicolumn{1}{c}{94.07}   & \multicolumn{1}{c}{\cellcolor[gray]{0.9}52.08}   & 52.98  &88.43& \multicolumn{1}{c}{ \cellcolor[gray]{0.9}50.62} & 55.76  & 78.45& \multicolumn{1}{c}{\cellcolor[gray]{0.9}54.07} & \multicolumn{1}{c}{-}   &- & \cellcolor[gray]{0.9}- & 54.76  &94.65 & \multicolumn{1}{c}{ \cellcolor[gray]{0.9}72.94}  \\ 
			& \textbf{MFVLR (Ours)} & \textbf{99.99} & \textbf{99.99} &\multicolumn{1}{c}{\cellcolor[gray]{0.9}\textbf{65.60}} & \textbf{86.64} & 93.60&\multicolumn{1}{c}{\cellcolor[gray]{0.9}\textbf{71.68}} &\textbf{61.97} & \textbf{98.43} &\multicolumn{1}{c}{\cellcolor[gray]{0.9}\textbf{57.28}}  &    -   &   - & \multicolumn{1}{c}{\cellcolor[gray]{0.9} -}   & \multicolumn{1}{c}{54.63} & \multicolumn{1}{c}{98.97}& \multicolumn{1}{c}{\cellcolor[gray]{0.9}\textbf{75.47}}  \\\midrule\multirow{7}[2]{*}{Diffae} & Xception \cite{Xception}& 53.96  & 94.01& \multicolumn{1}{c}{\cellcolor[gray]{0.9}-}  & 49.98  & 61.27& \multicolumn{1}{c}{\cellcolor[gray]{0.9}-} & 50.12  & 68.04 & \multicolumn{1}{c}{\cellcolor[gray]{0.9}-} & 49.99 & 52.69& \multicolumn{1}{c}{\cellcolor[gray]{0.9}-} &\multicolumn{1}{c}{-}   &\multicolumn{1}{c}{-} & \multicolumn{1}{c}{\cellcolor[gray]{0.9}-}\\      & ViT  \cite{ViT}   & 50.52  & 50.04& \multicolumn{1}{c}{\cellcolor[gray]{0.9}-}  & 49.45  & 45.06& \multicolumn{1}{c}{\cellcolor[gray]{0.9}-} & 49.70  & 47.99 & \multicolumn{1}{c}{\cellcolor[gray]{0.9}-} & 49.51 & 46.50 & \multicolumn{1}{c}{\cellcolor[gray]{0.9}-} &\multicolumn{1}{c}{-}     &\multicolumn{1}{c}{-} & \multicolumn{1}{c}{\cellcolor[gray]{0.9}-} \\      & CViT \cite{CViT} & 57.78  & 98.04 & \multicolumn{1}{c}{\cellcolor[gray]{0.9}-} & 50.23  & 83.37& \multicolumn{1}{c}{\cellcolor[gray]{0.9}-} & 50.00  & 54.42 & \multicolumn{1}{c}{\cellcolor[gray]{0.9}-} & 50.13 & 80.12& \multicolumn{1}{c}{\cellcolor[gray]{0.9}-} &  \multicolumn{1}{c}{-}      & \multicolumn{1}{c}{-}& \multicolumn{1}{c}{\cellcolor[gray]{0.9}-} \\   & DIRE \cite{DIRE}  & 57.45  & 94.21& \multicolumn{1}{c}{\cellcolor[gray]{0.9}-}  & 50.07  & 62.30& \multicolumn{1}{c}{\cellcolor[gray]{0.9}-}  & 50.12  & 74.84 & \multicolumn{1}{c}{\cellcolor[gray]{0.9}-} & 64.14 & 99.02& \multicolumn{1}{c}{\cellcolor[gray]{0.9}-} &  \multicolumn{1}{c}{-}   &  \multicolumn{1}{c}{-}& \multicolumn{1}{c}{\cellcolor[gray]{0.9}-}\\      & FreqNet \cite{Freq}& 53.48  &94.58 & \multicolumn{1}{c}{\cellcolor[gray]{0.9}-} & 49.91  & 44.41 & \multicolumn{1}{c}{\cellcolor[gray]{0.9}-} & 59.86  & 53.10 & \multicolumn{1}{c}{\cellcolor[gray]{0.9}-} & 50.04 & 48.82 & \multicolumn{1}{c}{\cellcolor[gray]{0.9}-}&\multicolumn{1}{c}{-}&   \multicolumn{1}{c}{-} & \multicolumn{1}{c}{\cellcolor[gray]{0.9}-}\\      & CLIP \cite{clip}  & 50.52  & 51.58 & \multicolumn{1}{c}{\cellcolor[gray]{0.9}-} & 49.91  &  50.11 & \multicolumn{1}{c}{\cellcolor[gray]{0.9}-}& 50.80  & 52.07 & \multicolumn{1}{c}{\cellcolor[gray]{0.9}-} & 49.97 &50.32 & \multicolumn{1}{c}{ \cellcolor[gray]{0.9}-}& \multicolumn{1}{c}{-}& \multicolumn{1}{c}{-}& \multicolumn{1}{c}{\cellcolor[gray]{0.9}-} \\     & MFCLIP \cite{mfclip}  & \textbf{98.99}  & \textbf{99.99} & \multicolumn{1}{c}{\cellcolor[gray]{0.9}-} & \textbf{99.82}  &  99.98 & \multicolumn{1}{c}{\cellcolor[gray]{0.9}-}& 60.07  & 75.80 & \multicolumn{1}{c}{\cellcolor[gray]{0.9}-} & \textbf{99.99} &\textbf{99.99} & \multicolumn{1}{c}{ \cellcolor[gray]{0.9}-}& \multicolumn{1}{c}{-}& \multicolumn{1}{c}{-}& \multicolumn{1}{c}{\cellcolor[gray]{0.9}-} \\    
			& FatFormer \cite{FatFormer}  & 53.56 & 54.85 & \multicolumn{1}{c}{\cellcolor[gray]{0.9}-} & 52.01 &  53.89& \multicolumn{1}{c}{\cellcolor[gray]{0.9}-} & 52.58  & 54.05 & \multicolumn{1}{c}{\cellcolor[gray]{0.9}-} & 52.57 & 53.65& \multicolumn{1}{c}{\cellcolor[gray]{0.9}-} & \multicolumn{1}{c}{-}& \multicolumn{1}{c}{-} & \multicolumn{1}{c}{\cellcolor[gray]{0.9}-}\\    
			& VLFFD \cite{VLFFD}& 62.79  & 93.43 & \multicolumn{1}{c}{\cellcolor[gray]{0.9}-} &   60.77   &   82.54 & \multicolumn{1}{c}{\cellcolor[gray]{0.9}-} & 52.78  & 65.00 & \multicolumn{1}{c}{\cellcolor[gray]{0.9}-} & 63.29 & 98.36& \multicolumn{1}{c}{\cellcolor[gray]{0.9}-}& \multicolumn{1}{c}{-} & \multicolumn{1}{c}{-}& \multicolumn{1}{c}{\cellcolor[gray]{0.9}-} \\      & DD-VQA \cite{DD-VQA} & 64.45 & 95.06 & \multicolumn{1}{c}{\cellcolor[gray]{0.9}-} &    61.78  &   83.11 & \multicolumn{1}{c}{ \cellcolor[gray]{0.9}-}& 52.87  & 69.57 & \multicolumn{1}{c}{\cellcolor[gray]{0.9}-} & 60.99 & 96.47& \multicolumn{1}{c}{\cellcolor[gray]{0.9}-} & \multicolumn{1}{c}{-} & \multicolumn{1}{c}{-}& \multicolumn{1}{c}{\cellcolor[gray]{0.9}-}\\     & C2P-CLIP \cite{C2P-CLIP} & 67.55 & 97.86 & \multicolumn{1}{c}{\cellcolor[gray]{0.9}-} &   63.89  &   95.37 & \multicolumn{1}{c}{\cellcolor[gray]{0.9}-}  & 55.92 & 76.39 & \multicolumn{1}{c}{\cellcolor[gray]{0.9}-} & 65.87& 97.54& \multicolumn{1}{c}{\cellcolor[gray]{0.9}-} & \multicolumn{1}{c}{-} & \multicolumn{1}{c}{-}& \multicolumn{1}{c}{\cellcolor[gray]{0.9}-}\\ 	  & SIDA \cite{SIDA} &69.86 & 98.67 & \multicolumn{1}{c}{\cellcolor[gray]{0.9}70.41} &    65.89  &    96.20 & \multicolumn{1}{c}{\cellcolor[gray]{0.9}57.64}  & 56.99 & 79.47 & \multicolumn{1}{c}{\cellcolor[gray]{0.9}56.49} & 68.96& 98.05& \multicolumn{1}{c}{\cellcolor[gray]{0.9}76.59} & \multicolumn{1}{c}{-} & \multicolumn{1}{c}{-}& \multicolumn{1}{c}{\cellcolor[gray]{0.9}-}\\ 	  & DiffForensics \cite{DiffForensics} & 70.68  & 99.04 & \multicolumn{1}{c}{\cellcolor[gray]{0.9}71.28} &   66.17  &    97.34  & \multicolumn{1}{c}{\cellcolor[gray]{0.9}56.39}  & 57.66 & 74.21 & \multicolumn{1}{c}{\cellcolor[gray]{0.9}55.04} & 70.27& 98.93& \multicolumn{1}{c}{\cellcolor[gray]{0.9}77.68} & \multicolumn{1}{c}{-} & \multicolumn{1}{c}{-}& \multicolumn{1}{c}{\cellcolor[gray]{0.9}-}\\ & M2TR \cite{M2TR} & \multicolumn{1}{c}{50.01}      &  \multicolumn{1}{c}{87.78}  & \multicolumn{1}{c}{\cellcolor[gray]{0.9}67.92}    & 50.00  & 74.32& \multicolumn{1}{c}{\cellcolor[gray]{0.9}49.66} & 50.00  & 65.35& \multicolumn{1}{c}{\cellcolor[gray]{0.9}49.81} & 50.01  &78.61 & \multicolumn{1}{c}{ \cellcolor[gray]{0.9}72.14}& \multicolumn{1}{c}{-}  & \multicolumn{1}{c}{-} & \multicolumn{1}{c}{\cellcolor[gray]{0.9}-} \\  
			& HiFi-Net \cite{HiFi-Net} & \multicolumn{1}{c}{54.94}      &  \multicolumn{1}{c}{89.00}   & \multicolumn{1}{c}{\cellcolor[gray]{0.9}68.97}     & 53.48  & 76.32  & \multicolumn{1}{c}{\cellcolor[gray]{0.9}52.45}   & 52.77 & 69.25 & \multicolumn{1}{c}{\cellcolor[gray]{0.9}50.97}& 53.06  & 82.37 & \multicolumn{1}{c}{\cellcolor[gray]{0.9}74.85} & \multicolumn{1}{c}{-} & \multicolumn{1}{c}{-} & \multicolumn{1}{c}{\cellcolor[gray]{0.9}-}  \\
			&MSCCNet \cite{MSCCNet} & \multicolumn{1}{c}{55.80}      &  \multicolumn{1}{c}{92.04}  & \multicolumn{1}{c}{\cellcolor[gray]{0.9}69.17}   & 56.98  & 79.01& \multicolumn{1}{c}{\cellcolor[gray]{0.9}55.69} & 54.55 & 72.78 & \multicolumn{1}{c}{\cellcolor[gray]{0.9}51.34} & 54.83  & 85.07 &\multicolumn{1}{c}{\cellcolor[gray]{0.9}75.96}&  \multicolumn{1}{c}{-}  &  \multicolumn{1}{c}{-}&  \multicolumn{1}{c}{\cellcolor[gray]{0.9}-}   \\ 
			& \textbf{MFVLR (Ours)} & 94.65 & 99.76&  \multicolumn{1}{c}{\cellcolor[gray]{0.9}\textbf{73.84}}   & 68.97&  \textbf{99.99} &  \multicolumn{1}{c}{\cellcolor[gray]{0.9}\textbf{58.09}}   &  \textbf{61.79}   &  \textbf{79.73}&  \multicolumn{1}{c}{\cellcolor[gray]{0.9}\textbf{58.09}}     & 84.60 &  99.83&  \multicolumn{1}{c}{\cellcolor[gray]{0.9}\textbf{79.15}}  &\multicolumn{1}{c}{-}& \multicolumn{1}{c}{-} & \multicolumn{1}{c}{\cellcolor[gray]{0.9}-} 
			\\\bottomrule\end{tabular}%
	}
		\vspace{-1em}
\end{table*}

\subsection{Experiment Setup}

{\bfseries\setlength\parindent{0em} Implementation details.} We deployed the model via PyTorch on the Tesla V100 GPU. We set the number of transformer blocks $B$, $E$, and $D$ in MFVLR to 4, 12, and 7, respectively. The vocabulary size $s$, the category $f$, the feature dimension $d$, the number of word tokens $n$, and the batch size $b$ are set to 49408, 2, 512, 308, and 8, respectively. The channel $c$, height $h$, and width $w$ are set to 1024, 14, and 14, respectively. Our model is trained using the Adam optimizer with a learning rate of 1e-4 and weight decay of 1e-3. The scheduler is used to reduce the learning rate by ten times every 15 epochs.

{\bfseries\setlength\parindent{0em} Metrics.} We utilized accuracy (ACC) and area under the receiver operating characteristic curve (AUC) for image-level detection metrics. We employed the mean of class-wise intersection over union (mIoU) for pixel-level localization metrics. The higher the value, the better the performance.

{\bfseries\setlength\parindent{0em}  Datasets.} We leveraged the hierarchical face forgery dataset GenFace \cite{genface} to achieve DFFDL. GenFace includes a large number of fake images generated by various advanced generators, which involve three categories: entire face synthesis (EFS) (e.g., DDPM \cite{ddpm}, LatDiff \cite{lad}, CollDiff \cite{coll}), attribute manipulation (AM) (e.g., Diffae \cite{diffae}, LatTrans \cite{Latent}, IAFaces \cite{IA-FaceS}), and face swapping (FS) (e.g., FSLSD \cite{FSLSD}, FaceSwapper \cite{FaceSwapper}). We performed the cross-forgery protocol and cross-generator evaluation on GenFace. Four deepfake datasets are employed to conduct the cross-dataset evaluation, to demonstrate the generalization and robustness of frameworks: FF++ \cite{FF++}, DFDC \cite{DFDC}, Celeb-DF \cite{Celeb-DF}, and DF-1.0 \cite{DF1.0}.

\vspace{-1em}
\subsection{Comparison with the State of the Art}

We studied the performance of state-of-the-art methods on face images created via diffusion models using the GenFace dataset. We chose some models including space-based Xception \cite{Xception}, frequency-based FreqNet \cite{Freq}, transformer-based ViT \cite{ViT}, CViT \cite{CViT} and FatFormer \cite{FatFormer}, vision-language-based CLIP \cite{clip}, VLFFD \cite{VLFFD}, MFCLIP \cite{mfclip}, C2P-CLIP \cite{C2P-CLIP}, as well as DD-VQA \cite{DD-VQA}, DIRE \cite{DIRE} employed to detect diffusion images, various FFDL methods like M2TR \cite{M2TR}, DADF \cite{DADF}, SIDA \cite{SIDA} as well as HiFi-Net \cite{HiFi-Net}, and the DFFDL model MSCCNet \cite{MSCCNet} and DiffForensics \cite{DiffForensics}.

	\begin{figure}[t!]
	\centering
	\includegraphics[width=\linewidth]{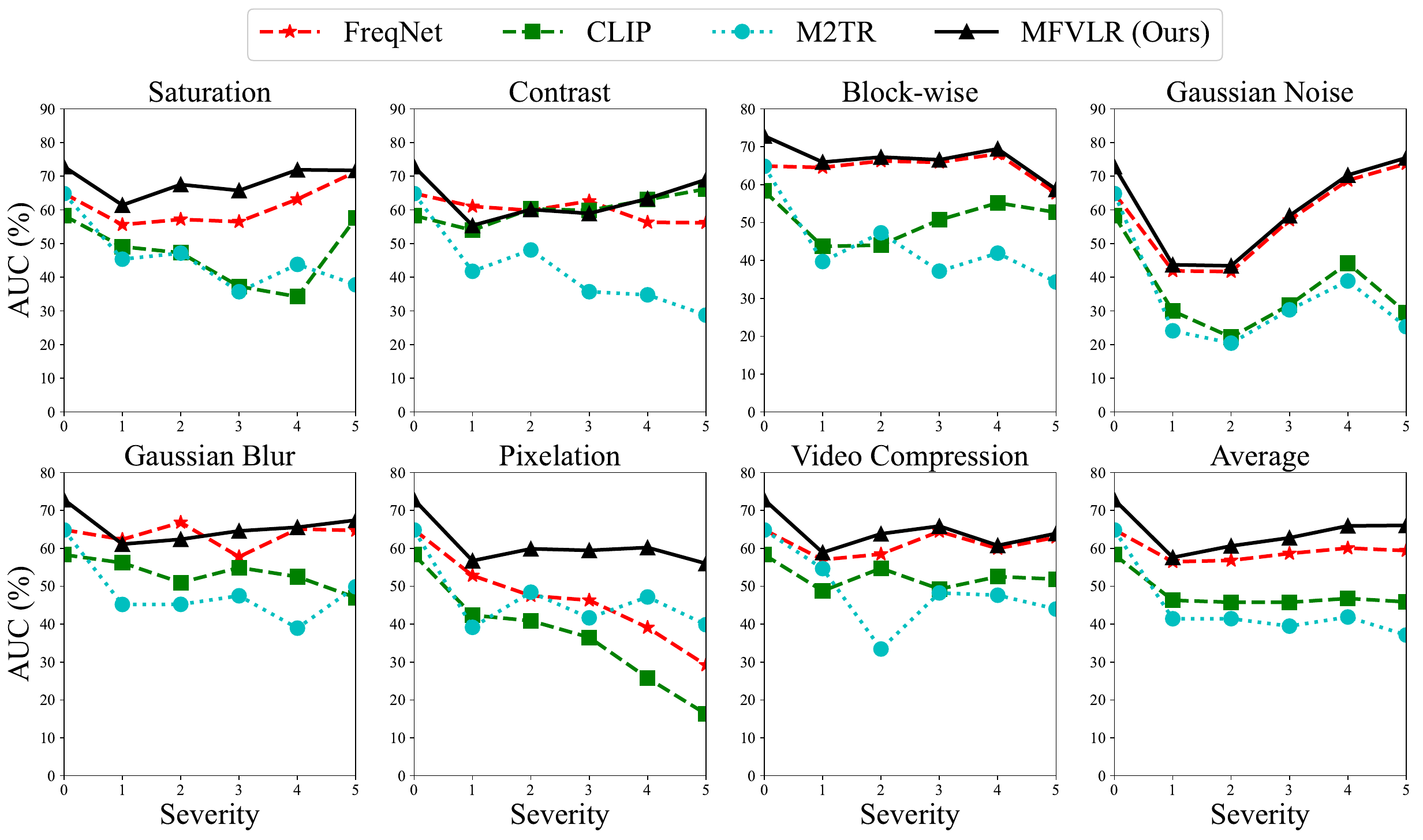} 
	\caption{ The robustness of models to unseen various image perturbations. }\label{figrob}
	\vspace{-2em}
\end{figure}

\begin{table}[t]
	\caption{Cross-generator evaluation on AM. ACC and AUC scores (\%) on remaining generators, after training using one generator. 
		\label{tabam}}
	\setlength{\tabcolsep}{1.0mm}{
		\begin{tabular}{rrrrrrrr}\toprule
			\multicolumn{1}{c}{\multirow{3}[6]{*}{Training Set}} &
			\multicolumn{1}{c}{\multirow{3}[6]{*}{Model}} &
			\multicolumn{6}{c}{\rule{0pt}{2.4ex}Testing Set } \\
			\cmidrule{3-8}
			& & \multicolumn{2}{c}{Diffae} & \multicolumn{2}{c}{LatTrans} & \multicolumn{2}{c}{IAFaces} \\
			\cmidrule{3-8}
			& & \multicolumn{1}{c}{ACC} & \multicolumn{1}{c}{\cellcolor[gray]{0.9}AUC}
			& \multicolumn{1}{c}{ACC} & \multicolumn{1}{c}{\cellcolor[gray]{0.9}AUC}
			& \multicolumn{1}{c}{ACC} & \multicolumn{1}{c}{\cellcolor[gray]{0.9}AUC} \\
			\midrule \multicolumn{1}{c}{\multirow{7}[2]{*}{Diffae}} & \multicolumn{1}{c}{Xception \cite{Xception}} &    \multicolumn{1}{c}{-}   &    \multicolumn{1}{c}{\cellcolor[gray]{0.9}-}   & \multicolumn{1}{c}{50.00 } & \multicolumn{1}{c}{\cellcolor[gray]{0.9}68.76} & \multicolumn{1}{c}{52.05 } & \multicolumn{1}{c}{\cellcolor[gray]{0.9}55.51} \\      & \multicolumn{1}{c}{ViT \cite{ViT}} &     \multicolumn{1}{c}{-}  &     \multicolumn{1}{c}{\cellcolor[gray]{0.9}-}  & \multicolumn{1}{c}{55.35} & \multicolumn{1}{c}{\cellcolor[gray]{0.9}76.10} & \multicolumn{1}{c}{50.45 } & \multicolumn{1}{c}{\cellcolor[gray]{0.9}54.20 } \\      & \multicolumn{1}{c}{CViT \cite{CViT}} &   \multicolumn{1}{c}{-}    &    \multicolumn{1}{c}{\cellcolor[gray]{0.9}-}  & \multicolumn{1}{c}{50.00 } & \multicolumn{1}{c}{\cellcolor[gray]{0.9}52.03} & \multicolumn{1}{c}{49.95 } & \multicolumn{1}{c}{\cellcolor[gray]{0.9}66.64 } \\    
		 	 & \multicolumn{1}{c}{M2TR \cite{M2TR}} &   \multicolumn{1}{c}{-}    &    \multicolumn{1}{c}{\cellcolor[gray]{0.9}-}  & \multicolumn{1}{c}{50.00 } & \multicolumn{1}{c}{\cellcolor[gray]{0.9}71.28} & \multicolumn{1}{c}{50.00 } & \multicolumn{1}{c}{\cellcolor[gray]{0.9}64.93 } \\ 
		 		 	 & \multicolumn{1}{c}{DADF \cite{DADF}} &   \multicolumn{1}{c}{-}    &    \multicolumn{1}{c}{\cellcolor[gray]{0.9}-}  & \multicolumn{1}{c}{50.37 } & \multicolumn{1}{c}{\cellcolor[gray]{0.9}74.24} & \multicolumn{1}{c}{53.95 } & \multicolumn{1}{c}{\cellcolor[gray]{0.9}65.39 } \\ 
		 	 		 	 & \multicolumn{1}{c}{HiFi-Net \cite{HiFi-Net}} &   \multicolumn{1}{c}{-}    &    \multicolumn{1}{c}{\cellcolor[gray]{0.9}-}  & \multicolumn{1}{c}{51.42 } & \multicolumn{1}{c}{\cellcolor[gray]{0.9}75.07} & \multicolumn{1}{c}{52.12 } & \multicolumn{1}{c}{\cellcolor[gray]{0.9}68.03 } \\ 
		 	 		 	  	 & \multicolumn{1}{c}{MSCCNet \cite{MSCCNet}} &   \multicolumn{1}{c}{-}    &    \multicolumn{1}{c}{\cellcolor[gray]{0.9}-}  & \multicolumn{1}{c}{50.56 } & \multicolumn{1}{c}{\cellcolor[gray]{0.9}76.88} & \multicolumn{1}{c}{54.67} & \multicolumn{1}{c}{\cellcolor[gray]{0.9}70.39 } \\ 
		
			  & \multicolumn{1}{c}{DIRE \cite{DIRE}} &    \multicolumn{1}{c}{-}   &    \multicolumn{1}{c}{\cellcolor[gray]{0.9}-}   & \multicolumn{1}{c}{50.00 } & \multicolumn{1}{c}{\cellcolor[gray]{0.9}41.69} & \multicolumn{1}{c}{50.30 } & \multicolumn{1}{c}{\cellcolor[gray]{0.9}63.01 } \\      & \multicolumn{1}{c}{FreqNet \cite{Freq}} &   \multicolumn{1}{c}{-}    &  \multicolumn{1}{c}{\cellcolor[gray]{0.9}-}     & \multicolumn{1}{c}{50.21 } & \multicolumn{1}{c}{\cellcolor[gray]{0.9}51.06} & \multicolumn{1}{c}{50.00 } & \multicolumn{1}{c}{\cellcolor[gray]{0.9}63.91 } \\      & \multicolumn{1}{c}{CLIP \cite{clip}} &    \multicolumn{1}{c}{-}   &  \multicolumn{1}{c}{\cellcolor[gray]{0.9}-}     & \multicolumn{1}{c}{50.00 } & \multicolumn{1}{c}{\cellcolor[gray]{0.9}58.83} & \multicolumn{1}{c}{49.80 } & \multicolumn{1}{c}{\cellcolor[gray]{0.9}49.65 } \\     & \multicolumn{1}{c}{MFCLIP \cite{mfclip}} &    \multicolumn{1}{c}{-}   &  \multicolumn{1}{c}{\cellcolor[gray]{0.9}-}     & \multicolumn{1}{c}{59.96 } & \multicolumn{1}{c}{\cellcolor[gray]{0.9}79.76} & \multicolumn{1}{c}{54.55 } & \multicolumn{1}{c}{\cellcolor[gray]{0.9}90.27 } \\  
			& \multicolumn{1}{c}{FatFormer \cite{FatFormer}} &    \multicolumn{1}{c}{-}   &  \multicolumn{1}{c}{\cellcolor[gray]{0.9}-}     & \multicolumn{1}{c}{50.34 } & \multicolumn{1}{c}{\cellcolor[gray]{0.9}59.67} & \multicolumn{1}{c}{50.93 } & \multicolumn{1}{c}{\cellcolor[gray]{0.9}52.45 } \\  
			& \multicolumn{1}{c}{VLFFD \cite{VLFFD}} &    \multicolumn{1}{c}{-}   &  \multicolumn{1}{c}{\cellcolor[gray]{0.9}-}     & \multicolumn{1}{c}{51.02 } & \multicolumn{1}{c}{\cellcolor[gray]{0.9}60.74} & \multicolumn{1}{c}{51.02 } & \multicolumn{1}{c}{\cellcolor[gray]{0.9}72.81 } \\    
			& \multicolumn{1}{c}{DD-VQA \cite{DD-VQA}} &    \multicolumn{1}{c}{-}   &  \multicolumn{1}{c}{\cellcolor[gray]{0.9}-}     & \multicolumn{1}{c}{51.36 } & \multicolumn{1}{c}{\cellcolor[gray]{0.9}69.04} & \multicolumn{1}{c}{53.84 } & \multicolumn{1}{c}{\cellcolor[gray]{0.9}74.90 } \\    
			& \multicolumn{1}{c}{\textbf{MFVLR (Ours)}} &     \multicolumn{1}{c}{-}  &   \multicolumn{1}{c}{\cellcolor[gray]{0.9}-}    & \multicolumn{1}{c}{\textbf{59.98} } & \multicolumn{1}{c}{\textbf{\cellcolor[gray]{0.9}82.31}} & \multicolumn{1}{c}{\textbf{69.31 }} & \multicolumn{1}{c}{\textbf{\cellcolor[gray]{0.9}91.86}} \\\midrule\multicolumn{1}{c}{\multirow{7}[2]{*}{LatTrans}} & \multicolumn{1}{c}{Xception \cite{Xception}} & \multicolumn{1}{c}{51.31 } & \multicolumn{1}{c}{\cellcolor[gray]{0.9}63.14 } &  \multicolumn{1}{c}{-}     &  \multicolumn{1}{c}{\cellcolor[gray]{0.9}-}     & \multicolumn{1}{c}{50.05 } & \multicolumn{1}{c}{\cellcolor[gray]{0.9}50.07 } \\      & \multicolumn{1}{c}{ViT \cite{ViT}} & \multicolumn{1}{c}{49.73 } & \multicolumn{1}{c}{\cellcolor[gray]{0.9}49.03 } &     \multicolumn{1}{c}{-}  &      \multicolumn{1}{c}{\cellcolor[gray]{0.9}-} & \multicolumn{1}{c}{50.05 } & \multicolumn{1}{c}{\cellcolor[gray]{0.9}50.97 } \\      & \multicolumn{1}{c}{CViT \cite{CViT}} & \multicolumn{1}{c}{50.76 } & \multicolumn{1}{c}{\cellcolor[gray]{0.9}62.08 } &    \multicolumn{1}{c}{-}   &    \multicolumn{1}{c}{\cellcolor[gray]{0.9}-}   & \multicolumn{1}{c}{50.25 } & \multicolumn{1}{c}{\cellcolor[gray]{0.9}51.71 } \\     
		 & \multicolumn{1}{c}{M2TR \cite{M2TR}} &   \multicolumn{1}{c}{50.42}    &    \multicolumn{1}{c}{\cellcolor[gray]{0.9}76.96}  & \multicolumn{1}{c}{- } & \multicolumn{1}{c}{\cellcolor[gray]{0.9}-} & \multicolumn{1}{c}{50.00 } & \multicolumn{1}{c}{\cellcolor[gray]{0.9}63.06 } \\ 
	& \multicolumn{1}{c}{DADF \cite{DADF}} &   \multicolumn{1}{c}{52.90}    &    \multicolumn{1}{c}{\cellcolor[gray]{0.9}77.53}  & \multicolumn{1}{c}{- } & \multicolumn{1}{c}{\cellcolor[gray]{0.9}-} & \multicolumn{1}{c}{51.90 } & \multicolumn{1}{c}{\cellcolor[gray]{0.9}65.78 } \\ 
	& \multicolumn{1}{c}{HiFi-Net \cite{HiFi-Net}} &   \multicolumn{1}{c}{53.80}    &    \multicolumn{1}{c}{\cellcolor[gray]{0.9}78.54}  & \multicolumn{1}{c}{-} & \multicolumn{1}{c}{\cellcolor[gray]{0.9}-} & \multicolumn{1}{c}{52.05 } & \multicolumn{1}{c}{\cellcolor[gray]{0.9}66.22 } \\ 
	& \multicolumn{1}{c}{MSCCNet \cite{MSCCNet}} &   \multicolumn{1}{c}{54.70}    &    \multicolumn{1}{c}{\cellcolor[gray]{0.9}79.03}  & \multicolumn{1}{c}{- } & \multicolumn{1}{c}{\cellcolor[gray]{0.9}-} & \multicolumn{1}{c}{52.18 } & \multicolumn{1}{c}{\cellcolor[gray]{0.9}67.41 } \\ 
			 & \multicolumn{1}{c}{DIRE \cite{DIRE}} & \multicolumn{1}{c}{50.02 } & \multicolumn{1}{c}{\cellcolor[gray]{0.9}52.35 } &   \multicolumn{1}{c}{-}    & \multicolumn{1}{c}{\cellcolor[gray]{0.9}-}      & \multicolumn{1}{c}{50.00 } & \multicolumn{1}{c}{\cellcolor[gray]{0.9}56.72 } \\      & \multicolumn{1}{c}{FreqNet \cite{Freq}} & \multicolumn{1}{c}{49.88 } & \multicolumn{1}{c}{\cellcolor[gray]{0.9}49.16 } &    \multicolumn{1}{c}{-}   &   \multicolumn{1}{c}{\cellcolor[gray]{0.9}-}    & \multicolumn{1}{c}{50.00 } & \multicolumn{1}{c}{\cellcolor[gray]{0.9}60.38 } \\      & \multicolumn{1}{c}{CLIP \cite{clip}} & \multicolumn{1}{c}{50.02 } & \multicolumn{1}{c}{\cellcolor[gray]{0.9}47.69 } & \multicolumn{1}{c}{-}      &    \multicolumn{1}{c}{\cellcolor[gray]{0.9}-}   & \multicolumn{1}{c}{50.00 } & \multicolumn{1}{c}{\cellcolor[gray]{0.9}53.05 } \\      & \multicolumn{1}{c}{MFCLIP \cite{mfclip}} & \multicolumn{1}{c}{53.00 } & \multicolumn{1}{c}{\cellcolor[gray]{0.9}75.00 } & \multicolumn{1}{c}{-}      &    \multicolumn{1}{c}{\cellcolor[gray]{0.9}-}   & \multicolumn{1}{c}{55.26 } & \multicolumn{1}{c}{\cellcolor[gray]{0.9}75.41 } \\  
			& \multicolumn{1}{c}{FatFormer \cite{FatFormer}} &    \multicolumn{1}{c}{50.05 }   &  \multicolumn{1}{c}{\cellcolor[gray]{0.9}50.21}     & \multicolumn{1}{c}{-} & \multicolumn{1}{c}{\cellcolor[gray]{0.9}-} & \multicolumn{1}{c}{50.09 } & \multicolumn{1}{c}{\cellcolor[gray]{0.9}54.87 } \\    
			& \multicolumn{1}{c}{VLFFD \cite{VLFFD}} &    \multicolumn{1}{c}{50.07}   &  \multicolumn{1}{c}{\cellcolor[gray]{0.9}63.76}     & \multicolumn{1}{c}{- } & \multicolumn{1}{c}{\cellcolor[gray]{0.9}-} & \multicolumn{1}{c}{50.23 } & \multicolumn{1}{c}{\cellcolor[gray]{0.9}65.96 } \\    
			& \multicolumn{1}{c}{DD-VQA \cite{DD-VQA}} &    \multicolumn{1}{c}{52.65}   &  \multicolumn{1}{c}{\cellcolor[gray]{0.9}65.00}     & \multicolumn{1}{c}{- } & \multicolumn{1}{c}{\cellcolor[gray]{0.9}-} & \multicolumn{1}{c}{50.12 } & \multicolumn{1}{c}{\cellcolor[gray]{0.9}66.45 } \\    
			& \multicolumn{1}{c}{\textbf{MFVLR (Ours)}}  &   \multicolumn{1}{c}{\textbf{67.04}}    &    \multicolumn{1}{c}{\cellcolor[gray]{0.9}\textbf{84.69}}  & \multicolumn{1}{c}{- } & \multicolumn{1}{c}{\cellcolor[gray]{0.9}-} & \multicolumn{1}{c}{\textbf{56.38 }} & \multicolumn{1}{c}{\cellcolor[gray]{0.9}\textbf{76.29} } \\\midrule    \\\end{tabular}%
	}
	\vspace{-4em}
\end{table}

{\bfseries\setlength\parindent{0em} Cross-forgery evaluation.} To evaluate the generalization capability of different methods, we conducted a cross-forgery protocol. Specifically, networks are trained using facial images of one manipulation, and tested on those of the remaining forgeries. As shown in Table~\ref{tab1}, the detection results of our method outperform most approaches, highlighting the exceptional generalization ability of MFVLR. Notably, for FFDL models, the AUC of our model surpasses M2TR and MSCCNet by around 4.7\% and 2.1\%, respectively, on EFS after training with AM. We argued that our method owns more powerful capacity of capturing the diverse and general relations between vision and language modalities than other FFDL approaches. 

\begin{table}[t]
	\caption{The ablation results of MVLRT. We tested detectors on FSLSD and LatDiff, after training using DiffFace. \label{tababla}}
	\vspace{-1em}
	\setlength{\tabcolsep}{1.0mm}{
		\begin{tabular}{cccccccccccc}
			\toprule
			\multicolumn{6}{c}{\multirow{2}[7]{*}{Model}} &
			\multicolumn{6}{c}{Testing Set} \\
			\cmidrule(lr){7-12}
			& \multicolumn{5}{c}{} &
			\multicolumn{3}{c}{FSLSD} &
			\multicolumn{3}{c}{LatDiff} \\
			\midrule
			IE & AD & RE & MD & LE & LD&
			ACC & AUC & 	\cellcolor[gray]{0.9}mIoU &
			ACC & AUC & 	\cellcolor[gray]{0.9}mIoU \\
			\midrule
			\checkmark & \multicolumn{1}{c}{-} & \multicolumn{1}{c}{-} & \multicolumn{1}{c}{-} & \multicolumn{1}{c}{-} &
			- & 50.00 &	
			47.72 & \multicolumn{1}{c}{\cellcolor[gray]{0.9}-} & 58.40&70.46&\multicolumn{1}{c}{\cellcolor[gray]{0.9}-}\\			
			\checkmark  & \checkmark & \multicolumn{1}{c}{-} & \multicolumn{1}{c}{-} & \multicolumn{1}{c}{-} &
			\multicolumn{1}{c}{-}&50.00 &
			50.86 & \cellcolor[gray]{0.9}- &59.96 &73.06& \cellcolor[gray]{0.9}-\\
			\checkmark & \multicolumn{1}{c}{-} & \multicolumn{1}{c}{-} & \checkmark & \multicolumn{1}{c}{-} &
			\multicolumn{1}{c}{-}& 50.00 &
			48.62& 	\cellcolor[gray]{0.9}47.96& 60.32&72.58&\cellcolor[gray]{0.9}52.94 \\
			\checkmark & \checkmark & \checkmark  & \multicolumn{1}{c}{-} & \multicolumn{1}{c}{-} &
			\multicolumn{1}{c}{-} & 52.76 &
			59.27 &\multicolumn{1}{c}{\cellcolor[gray]{0.9}-} & 63.97&80.23&\multicolumn{1}{c}{\cellcolor[gray]{0.9}-}\\
			\checkmark  & \checkmark & \checkmark & \checkmark & \multicolumn{1}{c}{-} & \multicolumn{1}{c}{-}&53.68 &
			60.73&  \cellcolor[gray]{0.9}58.97& 65.75&83.40&\cellcolor[gray]{0.9}63.67 \\
			
			\checkmark & \checkmark & \checkmark & \checkmark & \checkmark &\multicolumn{1}{c}{-}&
			54.03& 62.15 & 	\cellcolor[gray]{0.9}61.54 &
			70.36 & 86.08  & 	\cellcolor[gray]{0.9}65.14 \\
			\checkmark & \checkmark & \checkmark & \checkmark & \checkmark &\checkmark&
			\textbf{58.97}   & \textbf{69.02} & 		\cellcolor[gray]{0.9}\textbf{65.04}  & \textbf{86.64} & \textbf{93.60} & \textbf{\cellcolor[gray]{0.9}71.68} \\
			\bottomrule
		\end{tabular}
	}
		\vspace{-1em}
\end{table}

\begin{table}[t!]
	\caption{Cross-generator evaluation on FS. ACC and AUC scores (\%) on remaining generators, after training using one generator. \label{tabfs}}
	\setlength{\tabcolsep}{0.9mm}{
		\begin{tabular}{cccccccc}\toprule\multirow{3}[6]{*}{Training Set} & \multirow{3}[6]{*}{Model} & \multicolumn{6}{c}{Testing Set } \\\cmidrule{3-8}      &       & \multicolumn{2}{c}{DiffFace} & \multicolumn{2}{c}{FSLSD} & \multicolumn{2}{c}{FaceSwapper} \\\cmidrule{3-8}      &       & ACC   & \cellcolor[gray]{0.9}AUC   & ACC   & \cellcolor[gray]{0.9}AUC   & ACC   & \cellcolor[gray]{0.9}AUC \\\midrule\multirow{7}[2]{*}{DiffFace} & Xception \cite{Xception}&     \multicolumn{1}{c}{-}  &   \multicolumn{1}{c}{\cellcolor[gray]{0.9}-}  & 50.00  & \cellcolor[gray]{0.9}48.45 & 50.00  & \cellcolor[gray]{0.9}83.60 \\      & ViT \cite{ViT}  &  \multicolumn{1}{c}{-}     & \multicolumn{1}{c}{\cellcolor[gray]{0.9}-}      & 52.27  & \cellcolor[gray]{0.9}54.47 & 65.58  & \cellcolor[gray]{0.9}86.02  \\      & CViT \cite{CViT} & \multicolumn{1}{c}{-}      &  \multicolumn{1}{c}{\cellcolor[gray]{0.9}-}     & 50.00  & \cellcolor[gray]{0.9}49.28 & 50.04  & \cellcolor[gray]{0.9}79.17  \\    
			& M2TR \cite{M2TR} & \multicolumn{1}{c}{-}      &  \multicolumn{1}{c}{\cellcolor[gray]{0.9}-}     &50.82  & \cellcolor[gray]{0.9}56.90& 51.49  & \cellcolor[gray]{0.9}95.48  \\  
			& DADF \cite{DADF} & \multicolumn{1}{c}{-}   &  \multicolumn{1}{c}{\cellcolor[gray]{0.9}-}     & 51.70  & \cellcolor[gray]{0.9}58.39 & 52.49 & \cellcolor[gray]{0.9}96.03  \\  
			& HiFi-Net \cite{HiFi-Net} & \multicolumn{1}{c}{-}      &  \multicolumn{1}{c}{\cellcolor[gray]{0.9}-}     & 53.25  & \cellcolor[gray]{0.9}59.36 &  54.79 & \cellcolor[gray]{0.9}97.16 \\
			&MSCCNet \cite{MSCCNet} & \multicolumn{1}{c}{-}      &  \multicolumn{1}{c}{\cellcolor[gray]{0.9}-}     & 54.57  & \cellcolor[gray]{0.9}60.93 & 58.10 & \cellcolor[gray]{0.9}98.46 \\
			& DIRE \cite{DIRE} &  \multicolumn{1}{c}{-}     & \multicolumn{1}{c}{\cellcolor[gray]{0.9}-}      & 50.00  & \cellcolor[gray]{0.9}55.49 & 50.00  & \cellcolor[gray]{0.9}88.01  \\      & FreqNet \cite{Freq}&  \multicolumn{1}{c}{-}     &  \multicolumn{1}{c}{\cellcolor[gray]{0.9}-}     & 49.75  & \cellcolor[gray]{0.9}44.42 & 49.65  & \cellcolor[gray]{0.9}69.93  \\      & CLIP \cite{clip} &    \multicolumn{1}{c}{-}   & \multicolumn{1}{c}{\cellcolor[gray]{0.9}-}      & 51.06  & \cellcolor[gray]{0.9}55.37 & 70.54  & \cellcolor[gray]{0.9}92.62  \\       & MFCLIP \cite{mfclip} &    \multicolumn{1}{c}{-}   & \multicolumn{1}{c}{\cellcolor[gray]{0.9}-}      & 55.96 & \cellcolor[gray]{0.9}65.76 & 76.52 & \cellcolor[gray]{0.9}99.93  \\     
			& FatFormer \cite{FatFormer} &    \multicolumn{1}{c}{-}   & \multicolumn{1}{c}{\cellcolor[gray]{0.9}-}      & 51.95  & \cellcolor[gray]{0.9}57.35 & 71.87  & \cellcolor[gray]{0.9}93.76  \\     
			& VLFFD\cite{VLFFD} &    \multicolumn{1}{c}{-}   & \multicolumn{1}{c}{\cellcolor[gray]{0.9}-}      & 52.44  & \cellcolor[gray]{0.9}56.03 & 72.97  & \cellcolor[gray]{0.9}94.03 \\  
			& DD-VQA \cite{DD-VQA} &    \multicolumn{1}{c}{-}   & \multicolumn{1}{c}{\cellcolor[gray]{0.9}-}      & 52.51  & \cellcolor[gray]{0.9}57.89 & 73.74& \cellcolor[gray]{0.9}95.99  \\     
			& \textbf{MFVLR (Ours)} &     \multicolumn{1}{c}{-}  &  \multicolumn{1}{c}{\cellcolor[gray]{0.9}-}     & \textbf{58.97}  & \textbf{\cellcolor[gray]{0.9}69.02}  & \textbf{ 77.31 } & \textbf{\cellcolor[gray]{0.9}99.95} \\\midrule\multirow{7}[2]{*}{FSLSD} & Xception \cite{Xception}& 50.26  & \cellcolor[gray]{0.9}54.22  &      \multicolumn{1}{c}{-} &   \multicolumn{1}{c}{\cellcolor[gray]{0.9}-}    & 51.49  &\cellcolor[gray]{0.9}72.14  \\      & ViT \cite{ViT}  & 50.01  & \cellcolor[gray]{0.9}49.08  &     \multicolumn{1}{c}{-}  & \multicolumn{1}{c}{\cellcolor[gray]{0.9}-}      & 50.21  & \cellcolor[gray]{0.9}64.98  \\      & CViT \cite{CViT}  & 50.03  & \cellcolor[gray]{0.9}47.60  &     \multicolumn{1}{c}{-}  &   \multicolumn{1}{c}{\cellcolor[gray]{0.9}-}    & 50.46  & \cellcolor[gray]{0.9}84.44  \\    
			& M2TR \cite{M2TR} & \multicolumn{1}{c}{50.42}      &  \multicolumn{1}{c}{\cellcolor[gray]{0.9}54.15}     & \multicolumn{1}{c}{-} & \multicolumn{1}{c}{\cellcolor[gray]{0.9}-} & 51.49  & \cellcolor[gray]{0.9}95.48  \\  
			& DADF \cite{DADF} & \multicolumn{1}{c}{50.96}      &  \multicolumn{1}{c}{\cellcolor[gray]{0.9}54.91}     &\multicolumn{1}{c}{-}    & \multicolumn{1}{c}{\cellcolor[gray]{0.9}-} & 52.06  & \cellcolor[gray]{0.9}95.31  \\  
			& HiFi-Net \cite{HiFi-Net} & \multicolumn{1}{c}{51.82}      &  \multicolumn{1}{c}{\cellcolor[gray]{0.9}55.03}     & \multicolumn{1}{c}{-}   & \multicolumn{1}{c}{\cellcolor[gray]{0.9}-}   & 51.77  & \cellcolor[gray]{0.9}96.70  \\
			&MSCCNet \cite{MSCCNet} & \multicolumn{1}{c}{51.06}      &  \multicolumn{1}{c}{\cellcolor[gray]{0.9}54.89}     & \multicolumn{1}{c}{-}   & \multicolumn{1}{c}{\cellcolor[gray]{0.9}-} & 52.79  & \cellcolor[gray]{0.9}\textbf{96.82} \\
			& DIRE \cite{DIRE} & 50.00  & \cellcolor[gray]{0.9}51.13  &  \multicolumn{1}{c}{-}     &     \multicolumn{1}{c}{\cellcolor[gray]{0.9}-}  & 50.14  & \cellcolor[gray]{0.9}57.44  \\      & FreqNet \cite{Freq}& 53.44  & \cellcolor[gray]{0.9}55.08  &   \multicolumn{1}{c}{-}    &  \multicolumn{1}{c}{\cellcolor[gray]{0.9}-}     & 49.22  & \cellcolor[gray]{0.9}72.31  \\      & CLIP \cite{clip} & 49.67  & \cellcolor[gray]{0.9}45.77  &     \multicolumn{1}{c}{-}  &     \multicolumn{1}{c}{\cellcolor[gray]{0.9}-}  & 52.37  & \cellcolor[gray]{0.9}72.90  \\    & MFCLIP \cite{mfclip} &    \multicolumn{1}{c}{53.65}   & \multicolumn{1}{c}{\cellcolor[gray]{0.9}55.59}      & \multicolumn{1}{c}{-}  &  \multicolumn{1}{c}{\cellcolor[gray]{0.9}-} & 55.52 & \cellcolor[gray]{0.9}92.15  \\    
			& FatFormer \cite{FatFormer} & 52.02 & \cellcolor[gray]{0.9}47.33  &     \multicolumn{1}{c}{-}  &     \multicolumn{1}{c}{\cellcolor[gray]{0.9}-}  & 54.71  & \cellcolor[gray]{0.9}73.09  \\      
			& VLFFD\cite{VLFFD} &    \multicolumn{1}{c}{51.78}   & \multicolumn{1}{c}{\cellcolor[gray]{0.9}53.56}      & -  & \cellcolor[gray]{0.9}- & 53.82  & \cellcolor[gray]{0.9}84.66  \\  
			& DD-VQA \cite{DD-VQA} &    \multicolumn{1}{c}{51.39}   & \multicolumn{1}{c}{\cellcolor[gray]{0.9}54.76}      & -  & \cellcolor[gray]{0.9}-& 54.60  & \cellcolor[gray]{0.9}86.51  \\   
			& \textbf{MFVLR (Ours)} & \textbf{58.76} & \cellcolor[gray]{0.9}\textbf{60.77}  &   \multicolumn{1}{c}{-}    &     \multicolumn{1}{c}{\cellcolor[gray]{0.9}-}  &\textbf{60.73} &\cellcolor[gray]{0.9}96.20 \\\bottomrule\end{tabular}%
	}
		\vspace{-1em}
\end{table}

{\bfseries\setlength\parindent{0em}Cross-generator evaluation.} To demonstrate the generalization of various detectors on unseen face images synthesized by diffusion methods, we executed the cross-diffusion evaluation using GenFace. As DFFDL can be regarded as a novel and complicated task, to the best of our knowledge, there have been no thorough experiments to verify the performance of FFDL approaches on cross-diffusion generators. Therefore, our research is the first to evaluate the generalization of networks to unseen facial images synthesized by diffusion models, systematically and extensively. In detail, we trained networks using the images created by one diffusion-based generator and tested them on different ones. As Table~\ref{tabdiff1} displays, the performance of our model outperforms that of most FFDL methods on the cross-diffusion protocol. Specifically, the AUC of the proposed MFVLR approach is about 10\% higher than M2TR on Diffae after training using LatDiff. Besides, the mIoU of MFVLR is around 5.03\%, 2.6\%, and 3.75\% higher than MSCCNet, SIDA, and DiffForensics on DiffFace. In Table~\ref{tabdiff2}, the AUC of our model is nearly 25.9\% and 23\% higher than that of C2P-CLIP and MFCLIP, respectively, on CollDiff after training using DiffFace. We argue that our model not only captures residual-aware general forgery patterns but also realizes precise vision-language matching via language reconstruction, thus alleviating diffusion-specific shifts. In Table~\ref{tabdiff2}, the AUC of MFVLR is around 99\% on various diffusion models after training using CollDiff, realizing the state-of-the-art performance. By contrast, in Table~\ref{tabam}, MFVLR achieves the best 82.31\% ACC on GAN-generated face images created by LatTrans after training using Diffae among other methods, which shows that our method achieves superior generalization to face images generated by diffusion models.
\begin{table}[t]
	\caption{Cross-dataset generalization. ACC and AUC scores on FF++, Celeb-DF, DFDC, and DF-1.0 after training using FF++.
		\label{tabcd}}
	\setlength{\tabcolsep}{1.1mm}{ 
		\begin{tabular}{ccccccccc}\toprule
			\multirow{2}[3]{*}{Method} & \multicolumn{2}{c}{FF++} & \multicolumn{2}{c}{Celeb-DF} & \multicolumn{2}{c}{DFDC} & \multicolumn{2}{c}{DF-1.0} \\
			\cmidrule{2-9}      & ACC   & \cellcolor[gray]{0.9}AUC   & ACC   & \cellcolor[gray]{0.9}AUC   & ACC   & \cellcolor[gray]{0.9}AUC   & ACC   & \cellcolor[gray]{0.9}AUC \\\midrule ViT \cite{ViT}   & 62.44 & \cellcolor[gray]{0.9}67.07 & 62.28 & \cellcolor[gray]{0.9}59.75 & 56.18 & \cellcolor[gray]{0.9}58.31 & 58.05 & \cellcolor[gray]{0.9}61.27 \\CViT \cite{CViT}  & 90.47 & \cellcolor[gray]{0.9}96.69 & 50.75  & \cellcolor[gray]{0.9}64.70  & 60.95 & \cellcolor[gray]{0.9}65.96 & 56.15 & \cellcolor[gray]{0.9}62.42\\M2TR \cite{M2TR}  & 97.93& \cellcolor[gray]{0.9}99.51 & -  & \cellcolor[gray]{0.9}68.20  & - & \cellcolor[gray]{0.9}- & -& \cellcolor[gray]{0.9}- \\RECCE \cite{cao2022end} & 97.06 & \cellcolor[gray]{0.9}99.32 &  \multicolumn{1}{c}{-}     & \cellcolor[gray]{0.9}68.71 &   \multicolumn{1}{c}{-}    & \cellcolor[gray]{0.9}69.06 &   \multicolumn{1}{c}{-}    & \multicolumn{1}{c}{\cellcolor[gray]{0.9}-} \\CEViT \cite{coccomini2022combining} & 93.67 & \cellcolor[gray]{0.9}98.36 & 44.24  & \cellcolor[gray]{0.9}65.29 & 66.14 & \cellcolor[gray]{0.9}75.55 & 62.16 & \cellcolor[gray]{0.9}67.51 \\FoCus \cite{FoCus} & 96.43 & \cellcolor[gray]{0.9}99.15 &    \multicolumn{1}{c}{-}   & \cellcolor[gray]{0.9}76.13 &   \multicolumn{1}{c}{-}    & \cellcolor[gray]{0.9}68.42 &    \multicolumn{1}{c}{-}  &\multicolumn{1}{c}{\cellcolor[gray]{0.9}-}  \\UIA-ViT \cite{UIA-ViT}  &    \multicolumn{1}{c}{-}   & \cellcolor[gray]{0.9}99.33 &   \multicolumn{1}{c}{-}   & \cellcolor[gray]{0.9}82.41 &  \multicolumn{1}{c}{-}     & \cellcolor[gray]{0.9}75.80  &   \multicolumn{1}{c}{-}    & \multicolumn{1}{c}{\cellcolor[gray]{0.9}-} 
			\\CC-Net \cite{3DDCS}  &    \multicolumn{1}{c}{-}   & \cellcolor[gray]{0.9}98.70 &   \multicolumn{1}{c}{-}   & \cellcolor[gray]{0.9}72.04 &  \multicolumn{1}{c}{-}     & \cellcolor[gray]{0.9}72.35  &   \multicolumn{1}{c}{-}    & \multicolumn{1}{c}{\cellcolor[gray]{0.9}-} 
				\\Qiao et al. \cite{Qiao}  &    \multicolumn{1}{c}{-}   & \cellcolor[gray]{0.9}\textbf{100} &   \multicolumn{1}{c}{-}   & \cellcolor[gray]{0.9}70.00 &  \multicolumn{1}{c}{-}     & \cellcolor[gray]{0.9}- &   \multicolumn{1}{c}{-}    & \multicolumn{1}{c}{\cellcolor[gray]{0.9}-} \\Yu et al. \cite{yu}&   \multicolumn{1}{c}{-}    & \cellcolor[gray]{0.9}99.55  &  \multicolumn{1}{c}{-}    & \cellcolor[gray]{0.9}72.86 &   \multicolumn{1}{c}{-}    & \cellcolor[gray]{0.9}69.23 &     \multicolumn{1}{c}{-}  & \multicolumn{1}{c}{\cellcolor[gray]{0.9}-}
			\\Guan et al. \cite{Guan}& - &\cellcolor[gray]{0.9}99.17&  -&\textbf{\cellcolor[gray]{0.9}95.14}& - &\cellcolor[gray]{0.9}74.65& - & \cellcolor[gray]{0.9}- 
			\\CLIP \cite{clip}  &    67.79   &     \cellcolor[gray]{0.9}69.57  &     64.18  &    \cellcolor[gray]{0.9}65.42   &  58.42     &    \cellcolor[gray]{0.9}57.65   &  57.63     &\cellcolor[gray]{0.9}56.01  \\VLFFD \cite{VLFFD} &   \multicolumn{1}{c}{-}    & \cellcolor[gray]{0.9}99.23 &      \multicolumn{1}{c}{-} & \cellcolor[gray]{0.9}84.80  &    \multicolumn{1}{c}{-}   & \cellcolor[gray]{0.9}\textbf{84.74}
			&   \multicolumn{1}{c}{-}    &\multicolumn{1}{c}{\cellcolor[gray]{0.9}-}  \\MSCCNet \cite{MSCCNet}  & 97.21 & \cellcolor[gray]{0.9}98.94 & 65.98  & \cellcolor[gray]{0.9}73.02 &70.56 & \cellcolor[gray]{0.9}73.44 & 65.21 & \cellcolor[gray]{0.9}70.69\\
			MFVLR (Ours) &  96.37    &   \cellcolor[gray]{0.9}99.69 & \textbf{66.73}   &  \cellcolor[gray]{0.9}73.08    &   \textbf{72.89}    &    \cellcolor[gray]{0.9}75.06  &   \textbf{67.68}    & \textbf{\cellcolor[gray]{0.9}73.97} \\\bottomrule
		\end{tabular}%
	}
	\vspace{-2.5em}
\end{table}

{\bfseries\setlength\parindent{0em}Cross-dataset evaluation.} 
		To further assess the generalization capability of MFVLR, we performed cross-dataset evaluations by training models on FF++ and testing them on FF++, CelebDF, DFDC, and DF-1.0. Due to the limited class labels available in FF++, only the first two levels of text prompts in FF++ are leveraged to train our MFVLR model. As shown in Table~\ref{tabcd}, the AUC of the proposed MFVLR approach is approximately 3.28\% higher than the recent model MSCCNet on the DF-1.0 dataset, demonstrating its stronger generalization performance, which is attributed to the powerful global space and residual forgery feature extraction capabilities of our model. Besides, we believed that our model could learn comprehensive and rich language representations, which enables the vision module to capture more diverse and discriminative visual forged traces via vision-language alignment.
		  
\begin{table*}[t]
	\caption{The impact of the VIM module. We tested models on DDPM, LatDiff, DiffFace, and Diffae, after training on CollDiff. * denotes the text or language encoder with VIM, † represents the text or language decoder with VIM, and ‡ means both the text encoder and decoder with VIM or the language encoder and decoder with VIM. \label{vimm}}
	\setlength{\tabcolsep}{2.3mm}{
		\begin{tabular}{ccccccccccccccc}
			\toprule
			\multirow{2}[3]{*}{Method} & \multicolumn{3}{c}{DDPM} & \multicolumn{3}{c}{LatDiff} & \multicolumn{3}{c}{DiffFace} & \multicolumn{3}{c}{Diffae}& \multirow{2}[4]{*}{\shortstack{Params (M)} }   \\\cmidrule{2-13}      & ACC   & AUC & \cellcolor[gray]{0.9}mIoU$\uparrow$   & ACC   & AUC & \cellcolor[gray]{0.9}mIoU$\uparrow$  & ACC   & AUC  & \cellcolor[gray]{0.9}mIoU$\uparrow$ & ACC   & AUC  & \cellcolor[gray]{0.9}mIoU$\uparrow$ \\\midrule CLIP w/o VIM    &     51.58  &    53.92  & \multicolumn{1}{c}{\cellcolor[gray]{0.9}-}  & 50.21     &    46.17 & \multicolumn{1}{c}{\cellcolor[gray]{0.9}-}   &  50.26     &   48.64    & \multicolumn{1}{c}{\cellcolor[gray]{0.9}-} &   49.82  &48.36 &\multicolumn{1}{c}{ \cellcolor[gray]{0.9}-} &  84.23     \\CLIP* w/ VIM    & 55.07 & 59.70  &\multicolumn{1}{c}{\cellcolor[gray]{0.9}-} & 51.29  & 50.97  & \multicolumn{1}{c}{\cellcolor[gray]{0.9}-}& 50.39 & 51.33  & \multicolumn{1}{c}{\cellcolor[gray]{0.9}-}& 51.36 & 52.57  &\multicolumn{1}{c}{\cellcolor[gray]{0.9}-} & 96.82    \\CLIP† w/ VIM    & 53.28 & 59.22 &\multicolumn{1}{c}{\cellcolor[gray]{0.9}-} & 50.04 & 49.62  &\multicolumn{1}{c}{\cellcolor[gray]{0.9}-} & 51.34 & 53.52 & \multicolumn{1}{c}{\cellcolor[gray]{0.9}-}& 50.39 & 54.46& \multicolumn{1}{c}{\cellcolor[gray]{0.9}-}& 115.72     \\CLIP‡ w/ VIM     & 56.30  & 60.21 & \multicolumn{1}{c}{\cellcolor[gray]{0.9}-} & 52.06    & 51.49 & \multicolumn{1}{c}{\cellcolor[gray]{0.9}-} & 52.67  & 54.49  & \multicolumn{1}{c}{\cellcolor[gray]{0.9}-} & 51.57 & 59.90 & \multicolumn{1}{c}{\cellcolor[gray]{0.9}-} & 128.31     \\
			\midrule
			Ours w/o VIM    &95.68  & 97.89  &\multicolumn{1}{c}{\cellcolor[gray]{0.9}94.96} & 92.27  & 93.64 &\cellcolor[gray]{0.9}86.93 &86.75 & 92.69 &\multicolumn{1}{c}{\cellcolor[gray]{0.9}66.98} & 93.25& 93.90 & \cellcolor[gray]{0.9}64.16&121.07  \\ Ours* w/ VIM  & 98.49  & 99.76  &\multicolumn{1}{c}{\cellcolor[gray]{0.9}97.85} & 93.90 & 95.86 &\multicolumn{1}{c}{\cellcolor[gray]{0.9}88.34}  & 86.90 & 94.93 &\multicolumn{1}{c}{\cellcolor[gray]{0.9}68.76} & 94.89 & 95.92&\multicolumn{1}{c}{\cellcolor[gray]{0.9}67.64}& 133.66 \\
			Ours† w/ VIM  & 97.93  & 98.60 &\multicolumn{1}{c}{\cellcolor[gray]{0.9}96.77}  & 93.45 & 94.99&\multicolumn{1}{c}{\cellcolor[gray]{0.9}87.76} & 88.35 & 96.37 &\multicolumn{1}{c}{\cellcolor[gray]{0.9}69.80} & 96.04 &  96.93 &\multicolumn{1}{c}{\cellcolor[gray]{0.9}68.46} &128.41 \\
			Ours‡ w/ VIM  & \textbf{100}   & \textbf{100 } & \multicolumn{1}{c}{\cellcolor[gray]{0.9}\textbf{98.56}}  & \textbf{96.70 }& \textbf{99.98}&\multicolumn{1}{c}{\cellcolor[gray]{0.9}\textbf{91.37}}  & \textbf{92.64} & \textbf{99.98} &\multicolumn{1}{c}{\cellcolor[gray]{0.9}\textbf{72.94}} & \textbf{99.96} & \textbf{99.99}&\multicolumn{1}{c}{\cellcolor[gray]{0.9}\textbf{71.32}} & 141.00  \\
			\bottomrule
		\end{tabular}%
	}
\end{table*}

{\bfseries\setlength\parindent{0em}Robustness to general image corruptions.}
We evaluated the robustness of detectors against various unseen image distortions by training models on GenFace and testing them on distorted images from DF-1.0 \cite{DF1.0}. Seven types of perturbations are included, with each applied at five different intensity levels. As displayed in Fig.~\ref{figrob}, detectors are tested using diverse image perturbations like saturation adjustments, contrast changes, block distortions, white Gaussian noise, blurring, pixelation, and video compression. We computed the average AUC scores of detectors across seven types of corrupted images, with an intensity level of 0 representing pristine images without degradation. The AUC variations for all detectors under different levels of distortion severity are presented in Fig.~\ref{figrob}. The results demonstrate that our model consistently outperforms most networks across seven types of image distortions.

\begin{table}[t!]
	\caption{Effects of loss functions. We tested models on FSLSD and FaceSwapper, after training on DiffFace. \label{loss}}
	\setlength{\tabcolsep}{2.9 mm}{
		\begin{tabular}{ccccc}
			\toprule
			\multicolumn{1}{c}{\multirow{2}[3]{*}{\shortstack{Loss \\ Function}}} & \multicolumn{2}{c}{FSLSD} & \multicolumn{2}{c}{FaceSwapper} \\\cmidrule{2-5}      & ACC   & \cellcolor[gray]{0.9}AUC & ACC   & \cellcolor[gray]{0.9}AUC    \\
			\midrule
			\multicolumn{1}{l}{	$\mathcal{L}_\text{fd}$} &50.00 & \cellcolor[gray]{0.9}47.72 & 50.00  & \cellcolor[gray]{0.9}85.47  \\	\multicolumn{1}{l}{$\mathcal{L}_\text{fd}$+$\mathcal{L}_\text{cmc}$}  &   52.95    &    \cellcolor[gray]{0.9}50.48   &    54.76   &   \cellcolor[gray]{0.9}88.13   \\\multicolumn{1}{l}{$\mathcal{L}_\text{fd}$+$\mathcal{L}_\text{ar}$+$\mathcal{L}_\text{fl}$ } &   53.68     &    \cellcolor[gray]{0.9}60.73&    59.66&  \cellcolor[gray]{0.9}90.48 \\\multicolumn{1}{l}{$\mathcal{L}_\text{fd}$+$\mathcal{L}_\text{ar}$+$\mathcal{L}_\text{fl}$+$\mathcal{L}_\text{cmc}$}  &     53.87  &     \cellcolor[gray]{0.9}61.20  &     64.74  & \cellcolor[gray]{0.9}92.86  \\\multicolumn{1}{l}{$\mathcal{L}_\text{fd}$+$\mathcal{L}_\text{ar}$+$\mathcal{L}_\text{fl}$+$\mathcal{L}_\text{cmc}$+$\mathcal{L}_\text{kl}$}  &     54.03  &     \cellcolor[gray]{0.9}62.15 &     69.70  & \cellcolor[gray]{0.9}96.91    \\
			\multicolumn{1}{c}{	$\mathcal{L}_\text{fd}$+$\mathcal{L}_\text{ar}$+$\mathcal{L}_\text{fl}$+$\mathcal{L}_\text{cmc}$+$\mathcal{L}_\text{kl}$+$\mathcal{L}_\text{lr}$ }& \textbf{58.97}   & \textbf{\cellcolor[gray]{0.9}69.02}   & \textbf{77.31} & \textbf{\cellcolor[gray]{0.9}99.95} \\
			\bottomrule
		\end{tabular}
	}
	\vspace{-1em}
\end{table}

\begin{figure*}[t!]
	\centering
	\includegraphics[width=\linewidth]{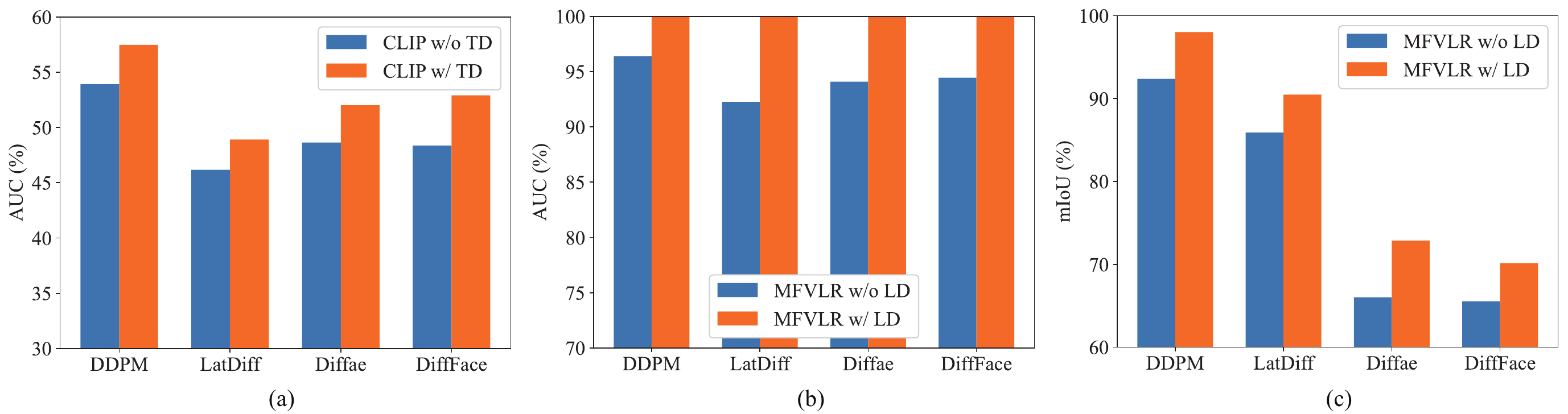} 
	\caption{(a) and (b) display the ablation results of CLIP (w/ TD or w/o TD) and MFVLR (w/ LD or w/o LD), respectively. (c) shows the mIoU scores of MFVLR (w/ LD or w/o LD) on images created by various diffusions. We trained models using CollDiff, and tested them on DDPM, LatDiff, Diffae, and DiffFace. TD is the text decoder, and LD is the language decoder.}\label{ldablation}
	\vspace{-1em}
\end{figure*}

\section{Ablation Study}\label{sec5}

\subsection{Effects of various modules}

In order to assess the impact of each component on learning capability, we evaluated the performance of models on FSLSD and LatDiff after training with DiffFace. Table~\ref{tababla} displays the ablation results of the proposed network. AD enhances the performance by 3.40\% AUC on LatDiff, showing that reconstructing appearance images could enhance the learning of general visual forgery embeddings, to facilitate the advancement of face forgery detection. Meanwhile, MD improves performance by 2.12\% AUC on LatDiff, demonstrating that DFFDL could benefit from the fine-grained pixel-level facial forgery representation learning. The addition of the RE module (+7.17\% AUC and +7.65\% mIoU) is evident, verifying that residuals provide useful knowledge for DFFDL. LE boosts AUC and mIoU by 2.68\% and 1.47\% on LatDiff, respectively, which shows that fine-grained class-aware linguistic representations can promote the model to study complementary and comprehensive visual forgery features, to benefit DFFDL. The addition of LD further boosts the performance by 7.52\% AUC and 6.54\% mIoU, respectively, indicating that reconstructing text prompts could guide the model to learn more general and fine-grained language embeddings, to explore more comprehensive and general visual face manipulated representations via vision-language matching.

 		\vspace{-1em}
 
 \subsection{Influences of transformer blocks in MVE}
 We investigate the impact of the number of transformer blocks in MVE. As Table~\ref{tabtransb} displays, the performance of MFVLR progressively increases with transformer blocks going deeper. In detail, an about 1.48\% increase of AUC could be reached by introducing a transformer block, which verifies that capturing diverse global relations among local feature patches is vital. In addition, performance enhancements could be realized as the depth of the transformer block increases. The AUC reaches a maximum as four transformer blocks are engaged. We argued that too many transformer blocks would yield redundancy and interfere with the model learning.
 
 		\vspace{-1em}
 \subsection{Influence of various loss functions}
 \vspace{-0.2em}
 To study the effectiveness of different loss functions, we performed the cross-generator evaluation by training models using DiffFace and testing them on FSLSD and FaceSwapper. As Table~\ref{loss} displays, when the model is trained with the FD loss function, the AUC is 85.47\% on FaceSwapper. Nevertheless, an around 2.66\% increase of AUC could be realized by adding the CMC loss, demonstrating the significance of cross-modal contrastive learning. Furthermore, the gains from introducing the LR loss (+3.04\%) are obvious. We believed that the LR loss could supervise the model to capture more abundant and fine-grained class-aware language representations, which could benefit the extraction of comprehensive and diverse visual forgery patterns. 
 
 		\vspace{-1em}
\subsection{Impact of introducing VIM into vision-language models}
We examined the effect of the plug-and-play VIM module. To demonstrate that the VIM module could be seamlessly integrated into various vision-language models for performance enhancements, we conducted the cross-generator evaluation by training models using CollDiff and testing them on DDPM, LatDiff, DiffFace and Diffae. As shown in Table~\ref{vimm}, it is evident that VIM enhances the generalizability of VLM to unseen face forgeries created via diffusion networks. For example, due to the introduction of VIM to the CLIP text encoder, the AUC of CLIP is improved by 4.8\% on LatDiff after training using CollDiff, with a growth of 12.59M parameters. Likewise, since VIM is introduced into the MFVLR language encoder, the AUC of MFVLR is increased by 2.22\%. When VIM is added to the CLIP text decoder or the MFVLR language decoder, the performance of CLIP and MFVLR increases by 6.10\% AUC and 3.13\% AUC on Diffae, accordingly. We further investigated the impact of the VIM module on localization performance. In Table~\ref{vimm}, we noticed that the mIoU of our MFVLR model is enhanced by 5.96\% on DiffFace with the addition of VIM. We believed that our model could capture more comprehensive and fine-grained language representations using vision-guided language reconstruction, and thus extract more comprehensive and fine-grained facial forgery patterns through vision-language matching.

		\vspace{-1em}
\subsection{Influences of the placement and number of VIM}
To explore VIM in greater depth, we delve into the influence of the placement and number of VIM. In Table~\ref{vimm}, the performance of our model attains the maximum when two VIMs are placed in LE and LD, respectively. Besides, as Table~\ref{vimdep} shows, our method with the VIM placed between MHA and FF in LE, and after MHA and FF in LD, achieves the best performance among models with VIMs at different locations, which emphasizes that the placement of VIMs plays a critical role in maximizing the model effectiveness. We argue that this placement strategically optimizes the interaction between the MHA and FF layers, ensuring that the model can effectively capture long-range vision-language modelling dependencies and enhance both local and global features across vision and language modalities.

\begin{table}[t]
	\caption{Impacts of the number of transformer blocks in MVE. We tested models on FSLSD and FaceSwapper, after training on DiffFace. \label{tabtransb}}
	\setlength{\tabcolsep}{4.2 mm}{
		\begin{tabular}{ccccc}
			\toprule
			\multicolumn{1}{c}{\multirow{2}[3]{*}{\shortstack{Transformer Block}}} & \multicolumn{2}{c}{FSLSD} & \multicolumn{2}{c}{FaceSwapper} \\\cmidrule{2-5}      & ACC   & \cellcolor[gray]{0.9}AUC & ACC   & \cellcolor[gray]{0.9}AUC    \\
			\midrule
			B=0 & 50.86& \cellcolor[gray]{0.9}60.97  & 71.53 & \cellcolor[gray]{0.9}94.78  \\B=1  &   51.97    &    \cellcolor[gray]{0.9}62.45   &    73.68   &   \cellcolor[gray]{0.9}96.09   \\B=2  &   52.90   &    \cellcolor[gray]{0.9}62.88  &    73.49   &   \cellcolor[gray]{0.9}96.80   \\B=3 &     54.21  &     \cellcolor[gray]{0.9}64.96  &     76.17   & \cellcolor[gray]{0.9}97.26    \\
			B=4 & \textbf{58.97}   & \textbf{\cellcolor[gray]{0.9}69.02}   & \textbf{77.31} & \textbf{\cellcolor[gray]{0.9}99.95} \\
			B=5 & 55.76   & \cellcolor[gray]{0.9}65.99   & 76.28& \cellcolor[gray]{0.9}99.90 \\
			\bottomrule
		\end{tabular}
	}
		\vspace{-2em}
\end{table}

\begin{table}[t!]
	\centering
	\caption{Ablation study of the placement and number of VIM. † denotes the VIM before MHA and FF. * means the VIM between MHA and FF. ‡ is the VIM after MHA and FF. First column symbol: various VIM positions in LE; Second column symbol: various VIM positions in LD.\label{vimdep}}
	\setlength{\tabcolsep}{1.2mm}{
		\small
\begin{tabular}{rccrrrrrr}
	\midrule
	\multirow{2}[3]{*}{} & \multicolumn{2}{c}{\multirow{2}[3]{*}{Method}} & \multicolumn{3}{c}{FSLSD} & \multicolumn{3}{c}{LatDiff} \\
	\cmidrule{4-9}      & \multicolumn{2}{c}{} & \multicolumn{1}{c}{ACC} & \multicolumn{1}{c}{AUC} & \multicolumn{1}{c}{\cellcolor[gray]{0.9}MIoU} & \multicolumn{1}{c}{ACC} & \multicolumn{1}{c}{AUC} & \multicolumn{1}{c}{\cellcolor[gray]{0.9}MIoU} \\
	\midrule
	\multicolumn{1}{c}{\multirow{3}[2]{*}{LE}} & \multicolumn{2}{c}{Ours w/VIM†*} & \multicolumn{1}{c}{55.12} & \multicolumn{1}{c}{64.86} & \multicolumn{1}{c}{\cellcolor[gray]{0.9}58.42 } & \multicolumn{1}{c}{83.98} & \multicolumn{1}{c}{90.55} & \multicolumn{1}{c}{\cellcolor[gray]{0.9}68.06} \\
	& \multicolumn{2}{c}{Ours w/VIM**} & \multicolumn{1}{c}{57.99} & \multicolumn{1}{c}{65.48} & \multicolumn{1}{c}{\cellcolor[gray]{0.9}63.79 } & \multicolumn{1}{c}{85.67} & \multicolumn{1}{c}{92.96} & \multicolumn{1}{c}{\cellcolor[gray]{0.9}70.72} \\
	& \multicolumn{2}{c}{Ours w/VIM‡*} & \multicolumn{1}{c}{56.82} & \multicolumn{1}{c}{63.55} & \multicolumn{1}{c}{\cellcolor[gray]{0.9}59.91 } & \multicolumn{1}{c}{84.45} & \multicolumn{1}{c}{92.78} & \multicolumn{1}{c}{\cellcolor[gray]{0.9}69.26} \\
	\midrule
	\multicolumn{1}{c}{\multirow{3}[2]{*}{LD}} & \multicolumn{2}{c}{Ours w/VIM*†} & \multicolumn{1}{c}{55.79} & \multicolumn{1}{c}{66.15} & \multicolumn{1}{c}{\cellcolor[gray]{0.9}58.81 } & \multicolumn{1}{c}{85.63} & \multicolumn{1}{c}{91.92} & \multicolumn{1}{c}{\cellcolor[gray]{0.9}68.05} \\
	& \multicolumn{2}{c}{Ours w/VIM**} & \multicolumn{1}{c}{56.42} & \multicolumn{1}{c}{67.34} & \multicolumn{1}{c}{\cellcolor[gray]{0.9}60.36 } & \multicolumn{1}{c}{84.37} & \multicolumn{1}{c}{90.32} & \multicolumn{1}{c}{\cellcolor[gray]{0.9}69.92} \\
	& \multicolumn{2}{c}{\textbf{Ours w/VIM*‡}} & \multicolumn{1}{c}{\textbf{58.97}} & \multicolumn{1}{c}{\textbf{69.02}} & \multicolumn{1}{c}{\textbf{\cellcolor[gray]{0.9}65.04}} & \multicolumn{1}{c}{\textbf{86.64}} & \multicolumn{1}{c}{\textbf{93.60}} & \multicolumn{1}{c}{\textbf{\cellcolor[gray]{0.9}71.68}} \\
	\midrule
	& \multicolumn{2}{c}{Ours w/VIM‡‡} & \multicolumn{1}{c}{54.92} & \multicolumn{1}{c}{59.96} & \multicolumn{1}{c}{\cellcolor[gray]{0.9}58.94} & \multicolumn{1}{c}{79.81} & \multicolumn{1}{c}{90.17} & \multicolumn{1}{c}{\cellcolor[gray]{0.9}67.09} \\
	\midrule
	& \multicolumn{2}{c}{} &       &       &       &       &       &  \\
\end{tabular}%

	}
	\vspace{-2em}
\end{table}

\begin{table}[t!]
	\centering
	\caption{Impacts of different language prompts.  \label{flp}}
	\setlength{\tabcolsep}{1.0mm}{
		\small
\begin{tabular}{rrrrrrrr}
	\midrule
	\multicolumn{2}{c}{\multirow{2}[3]{*}{Method}} & \multicolumn{3}{c}{FSLSD} & \multicolumn{3}{c}{LatDiff} \\
	\cmidrule{3-8}\multicolumn{2}{c}{} & \multicolumn{1}{c}{ACC} & \multicolumn{1}{c}{AUC} & \multicolumn{1}{c}{\cellcolor[gray]{0.9}MIoU} & \multicolumn{1}{c}{ACC} & \multicolumn{1}{c}{AUC} & \multicolumn{1}{c}{\cellcolor[gray]{0.9}MIoU} \\
	\midrule
	\multicolumn{2}{l}{Ours w/L1} & \multicolumn{1}{c}{53.43} & \multicolumn{1}{c}{58.80} & \multicolumn{1}{c}{\cellcolor[gray]{0.9}54.80 } & \multicolumn{1}{c}{82.43} & \multicolumn{1}{c}{89.55} & \multicolumn{1}{c}{\cellcolor[gray]{0.9}67.96} \\
	\multicolumn{2}{l}{Ours w/L1+L2} &  55.62     &  60.93     &  \cellcolor[gray]{0.9}58.18   &    83.25   &   91.76    &\cellcolor[gray]{0.9}69.02  \\
	\multicolumn{2}{l}{Ours w/L1+L2+L3} &    56.79   &    64.97   &  \cellcolor[gray]{0.9}62.41     &  85.09     &    92.89   & \cellcolor[gray]{0.9}70.53 \\
	\multicolumn{2}{l}{Ours w/L1+L2+L3+L4} & \multicolumn{1}{c}{\textbf{58.97}} & \multicolumn{1}{c}{\textbf{69.02}} & \multicolumn{1}{c}{\cellcolor[gray]{0.9}\textbf{65.04} } & \multicolumn{1}{c}{\textbf{86.64}} & \multicolumn{1}{c}{\textbf{93.60}} & \multicolumn{1}{c}{\cellcolor[gray]{0.9}\textbf{71.68}} \\
	\midrule
	\multicolumn{2}{l}{Ours w/o L1} & \multicolumn{1}{c}{57.21} & \multicolumn{1}{c}{68.09} & \multicolumn{1}{c}{\cellcolor[gray]{0.9}63.76 } & \multicolumn{1}{c}{85.08} & \multicolumn{1}{c}{92.09} & \multicolumn{1}{c}{\cellcolor[gray]{0.9}70.41} \\
	\multicolumn{2}{l}{Ours w/o L2} & \multicolumn{1}{c}{55.63} & \multicolumn{1}{c}{63.74} & \multicolumn{1}{c}{\cellcolor[gray]{0.9}61.08 } & \multicolumn{1}{c}{83.91} & \multicolumn{1}{c}{90.76} & \multicolumn{1}{c}{\cellcolor[gray]{0.9}68.93} \\
	\multicolumn{2}{l}{Ours w/o L3} & \multicolumn{1}{c}{56.37} & \multicolumn{1}{c}{66.72} & \multicolumn{1}{c}{\cellcolor[gray]{0.9}62.30 } & \multicolumn{1}{c}{84.99} & \multicolumn{1}{c}{91.02} & \multicolumn{1}{c}{\cellcolor[gray]{0.9}69.37} \\
	\multicolumn{2}{l}{Ours w/o L4} & \multicolumn{1}{c}{55.99} & \multicolumn{1}{c}{65.40} & \multicolumn{1}{c}{\cellcolor[gray]{0.9}61.79} & \multicolumn{1}{c}{84.67} & \multicolumn{1}{c}{91.98} & \multicolumn{1}{c}{\cellcolor[gray]{0.9}69.04} \\
	\midrule
\end{tabular}%

	}
\end{table}

\begin{table}[t!]
	\centering
	\caption{ Effects of different fusions in VIM. We tested models on FSLSD and LatDiff, after training on DiffFace.\label{fusion}}
	\setlength{\tabcolsep}{1.0mm}{
		\small
	\begin{tabular}{rrcccccc}

		\midrule
		\multicolumn{2}{c}{\multirow{2}[3]{*}{Interaction }} & \multicolumn{3}{c}{FSLSD} & \multicolumn{3}{c}{LatDiff} \\
		\cmidrule{3-8}\multicolumn{2}{c}{} & ACC   & AUC   &\cellcolor[gray]{0.9} MIoU  & ACC   & AUC   & \cellcolor[gray]{0.9}MIoU \\
		\midrule
		\multicolumn{2}{c}{Concatenation} & 54.07   & 64.09 & \cellcolor[gray]{0.9}64.58  & 85.71 & 92.64   & \cellcolor[gray]{0.9}69.10 \\
		\multicolumn{2}{c}{Summation} & 55.96    &65.78 &\cellcolor[gray]{0.9}64.31  & 84.53 & 90.79 & \cellcolor[gray]{0.9}68.42 \\
		\multicolumn{2}{c}{Cross-Attention} & \textbf{58.97} & \textbf{69.02} & \cellcolor[gray]{0.9}\textbf{65.04}  & \textbf{86.64} & \textbf{93.60} & \cellcolor[gray]{0.9}\textbf{71.68} \\
		\midrule
	\end{tabular}%
	}
\end{table}

\begin{table}[t!]
	\centering
	\caption{ Impacts of various keys and values in VIM.\label{kv}}
	\setlength{\tabcolsep}{0.5mm}{
		\small
		\begin{tabular}{ccccccccrr}
				\midrule
			\multicolumn{2}{c}{\multirow{2}[3]{*}{Interaction }} & \multicolumn{3}{c}{FSLSD} & \multicolumn{3}{c}{LatDiff} & \multicolumn{1}{c}{\multirow{2}[3]{*}{\shortstack{Params\\(M)} }} & \multicolumn{1}{c}{\multirow{2}[3]{*}{\shortstack{FLOPs\\ (G)}}} \\
			\cmidrule{3-8}\multicolumn{2}{c}{} & ACC   & AUC   & \cellcolor[gray]{0.9}MIoU  & ACC   & AUC   & \cellcolor[gray]{0.9}MIoU  &       &  \\
			\midrule
			\multicolumn{2}{c}{KV w/all} & 53.88    & 65.20 & \cellcolor[gray]{0.9}62.94  & 85.08 & 92.57   & \cellcolor[gray]{0.9}70.03 &   141.00    & 175.71 \\
			\multicolumn{2}{c}{KV w/patch} & 57.64    & 64.89 & \cellcolor[gray]{0.9}60.76  & 84.97 & 90.25 & \cellcolor[gray]{0.9}68.19 & 141.00      & 175.70 \\
			\multicolumn{2}{c}{KV w/cls} & \textbf{58.97} & \textbf{69.02} & \cellcolor[gray]{0.9}\textbf{65.04}  & \textbf{86.64} & \textbf{93.60} & \cellcolor[gray]{0.9}\textbf{71.68} &    141.00   & 173.04 \\
			\bottomrule
		\end{tabular}%
		
	}
\end{table}

\begin{table}[t!]
	\centering
	\caption{Effects of various number of transformer blocks in LE and LD. We tested models on FSLSD and LatDiff, after training on DiffFace.  \label{leld}}
	\setlength{\tabcolsep}{1.2mm}{
		\small
\begin{tabular}{rccrrrrrr}
	\cmidrule{2-9}      & \multicolumn{2}{c}{Transformer Block} & \multicolumn{3}{c}{FSLSD} & \multicolumn{3}{c}{LatDiff} \\
	\cmidrule{2-9}      & LE    & LD    & \multicolumn{1}{c}{ACC} & \multicolumn{1}{c}{AUC} & \multicolumn{1}{c}{\cellcolor[gray]{0.9}MIoU} & \multicolumn{1}{c}{ACC} & \multicolumn{1}{c}{AUC} & \multicolumn{1}{c}{\cellcolor[gray]{0.9}MIoU} \\
	\cmidrule{2-9}     
	& 10    & 6     & \multicolumn{1}{c}{53.43} & \multicolumn{1}{c}{58.80} & \multicolumn{1}{c}{\cellcolor[gray]{0.9}48.99} & \multicolumn{1}{c}{85.43} & \multicolumn{1}{c}{90.55} & \multicolumn{1}{c}{\cellcolor[gray]{0.9}66.92} \\
	& 11    & 6     &  55.69     &   62.74    &  \cellcolor[gray]{0.9}53.91    &  83.86     &  91.98     &\cellcolor[gray]{0.9}68.40 \\
	& 12    & 6     & \multicolumn{1}{c}{56.96} & \multicolumn{1}{c}{63.48 } & \multicolumn{1}{c}{\cellcolor[gray]{0.9}54.11} & \multicolumn{1}{c}{85.95} & \multicolumn{1}{c}{92.17} & \multicolumn{1}{c}{\cellcolor[gray]{0.9}69.61}  \\
	& 13    & 6     & \multicolumn{1}{c}{54.32} & \multicolumn{1}{c}{61.07 } & \multicolumn{1}{c}{\cellcolor[gray]{0.9}52.96} & \multicolumn{1}{c}{84.91} & \multicolumn{1}{c}{92.06} & \multicolumn{1}{c}{\cellcolor[gray]{0.9}67.83} \\
	\cmidrule{2-9}   
	& 12   & 5     & \multicolumn{1}{c}{50.00} & \multicolumn{1}{c}{62.18} & \multicolumn{1}{c}{\cellcolor[gray]{0.9}55.00 } & \multicolumn{1}{c}{79.56} & \multicolumn{1}{c}{89.63} & \multicolumn{1}{c}{\cellcolor[gray]{0.9}62.60} \\
	& 12    & 6     & \multicolumn{1}{c}{53.43} & \multicolumn{1}{c}{65.80} & \multicolumn{1}{c}{\cellcolor[gray]{0.9}60.99 } & \multicolumn{1}{c}{82.43} & \multicolumn{1}{c}{92.55} & \multicolumn{1}{c}{\cellcolor[gray]{0.9}66.90} \\
	& 12    & 7     & \multicolumn{1}{c}{\textbf{58.97}} & \multicolumn{1}{c}{\textbf{69.02}} & \multicolumn{1}{c}{\cellcolor[gray]{0.9}\textbf{65.04} } & \multicolumn{1}{c}{\textbf{86.64}} & \multicolumn{1}{c}{\textbf{93.60}} & \multicolumn{1}{c}{\cellcolor[gray]{0.9}\textbf{71.68}} \\
	& 12    & 8     &   55.46    &  66.32     &   \cellcolor[gray]{0.9}58.93  &    85.70   & 91.19      & \cellcolor[gray]{0.9}68.56 \\
	\cmidrule{2-9}\end{tabular}%

	}
	\vspace{-2em}
\end{table}

\begin{figure*}[t]
	\centering
	\includegraphics[width=\linewidth]{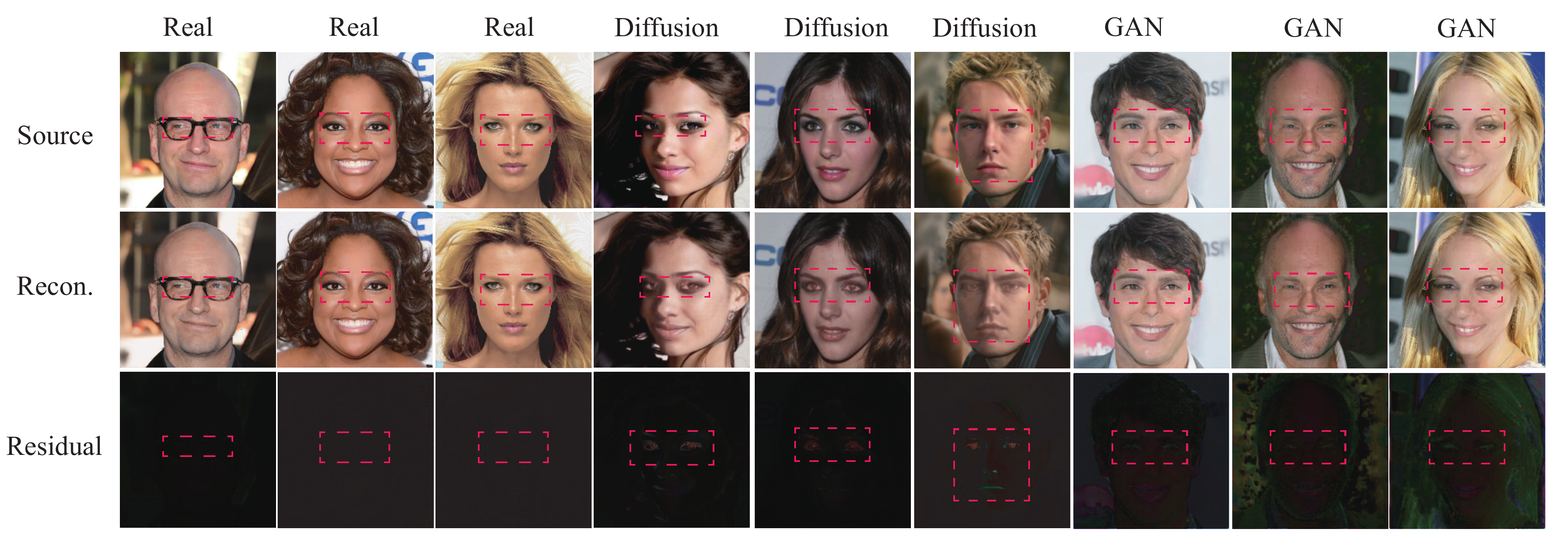} 
		\vspace{-1em}
	\caption{The visualization of the residual between the source image and the reconstructed one. The first to third rows represent the source image, reconstructed image, and the residual, respectively.}\label{residual} 
		\vspace{-1em}
\end{figure*}

\begin{figure}[t]
	\centering
	\includegraphics[width=\linewidth]{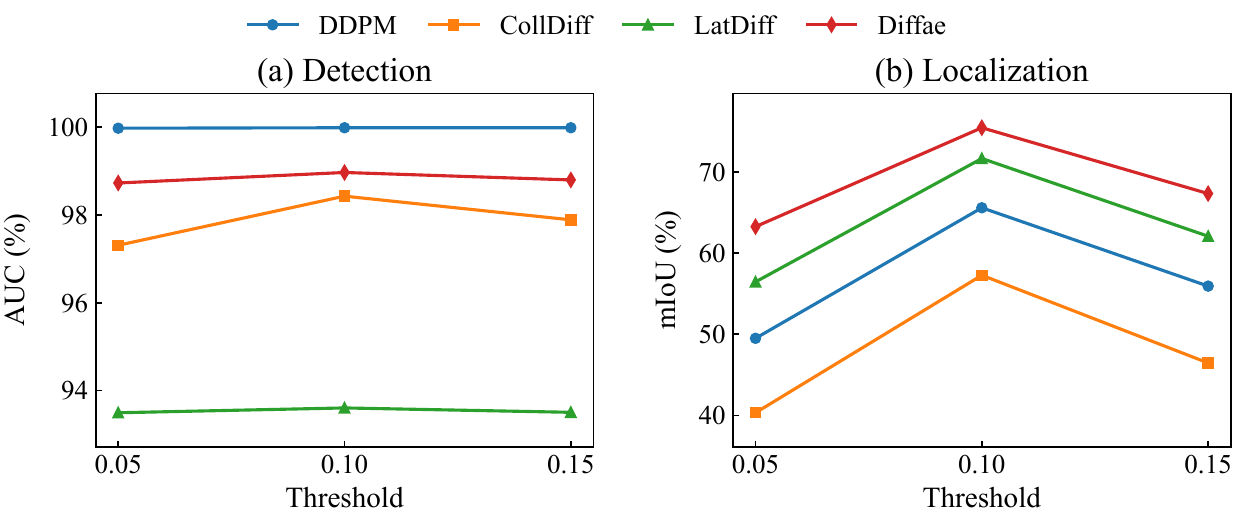} 
	\caption{Effect of the mask threshold on performance across different generators.}\label{threshold} 
	\vspace{-2em}
\end{figure}

\subsection{Effects of different interaction details in VIM}
For VIM, we employed the class token as the key and value. Next, we varied them to observe the effect on performance and efficiency. As Table~\ref{kv} shows, the FLOPs of the model are increased by about 2.7G when all tokens are included. Unlike the quadratic cost incurred when all tokens are used as keys and values, the attention map generation becomes linear, leading to a more efficient process. This demonstrates the advantage of VIM in improving efficiency. We further studied the impact of various fusion methods, such as concatenation, summation, and cross-attention. In Table~\ref{fusion}, among all compared strategies, the proposed cross-attention schemes achieve the best AUC and MIoU, demonstrating the effectiveness of the various global and fine-grained fusion between vision and language features.

		\vspace{-1.5em}
\subsection{Impacts of different language prompts}
To evaluate the model's sensitivity to variations in text prompt design, we conduct an ablation study including varying prompt granularity and missing forgery-type cues. We assess the performance of our method on FSLSD and LatDiff after training using DiffFace by gradually adding hierarchical texts. In Table~\ref{flp}, we observe that the performance of our model tends to improve with the increase of fine-grained text prompts. Specifically, both the AUC and MIoU of MFVLR are improved on FSLSD and LatDiff as text prompt granularity from levels 1 to 4 is added, validating that hierarchical text prompts guide the model to mine more general and discriminative forgery cues, to enhance the advancement of DFFDL. In addition, we assess the performance of our method using the cross-generator protocol by missing forgery-type hierarchical texts. We note that the model exhibits strong sensitivity to the level 2 text prompt and the level 4 text prompt. In detail, about 6\% and 4\% decrease of AUC could be achieved by missing the level 2 or level 4 text prompts, respectively, while the AUC of our model without level 3 prompts or level 1 prompts is reduced by about 2.3\% and 1\%, respectively. This indicates that the two levels of prompts can provide valuable semantic information to facilitate DFFDL.
		\vspace{-1em}
\subsection{Effects of language decoder}
 To analyze the effects of the language decoder, we investigated the performance of CLIP and our MFVLR method on DDPM, LatDiff, and Diffae after training on CollDiff. The results of model ablations are shown in Fig.~\ref{ldablation}. We add a text decoder to CLIP aligned with its text encoder. As shown in Fig.~\ref{ldablation} (a), we observed that the performance of the CLIP significantly grows with the addition of a text decoder. Likewise, Fig.~\ref{ldablation} (b) displays that the results of our MFVLR model show an increasing trend due to the introduction of the language decoder. In detail, the AUC of CLIP and our model is increased by about 4\% on DDPM, respectively. Besides, as Fig.~\ref{ldablation} (c) shows, the mIoU of  MFVLR is improved considerably on the images generated by diffusion models like DDPM, LatDiff, Diffae, and DiffFace, since the language decoder is added. We argued that language reconstruction offers more abundant semantic information, to push the model to capture more comprehensive and discriminative visual forgery traces, which facilitates the development of DFFDL.

		\vspace{-1em}
\subsection{Impacts of various number of blocks in LE and LD}

To delve into the impact of varying the number of blocks in LE and LD, we increase the number of transformer blocks in LE from 10 to 13, while six blocks in LD are involved. In Table~\ref{leld}, the performance in common increases with the growth of the number of transformer blocks. Both the AUC and MIoU achieve the maximum when twelve blocks are utilized and begin to decrease thereafter. We further raise the number of blocks in LD from 5 to 8, when twelve blocks in LE are used. We notice that the performance reaches a maximum when seven blocks in LD are employed and then decreases when the number of blocks equals eight. This suggests that there is an optimal number of transformer blocks beyond which adding more blocks does not lead to better performance. Additionally, excessive blocks may cause the model to overfit, as it starts to memorize the training distribution and does not generalize well to unseen generators, ultimately reducing its effectiveness.

\begin{figure}[t]
	\centering
	\includegraphics[width=\linewidth]{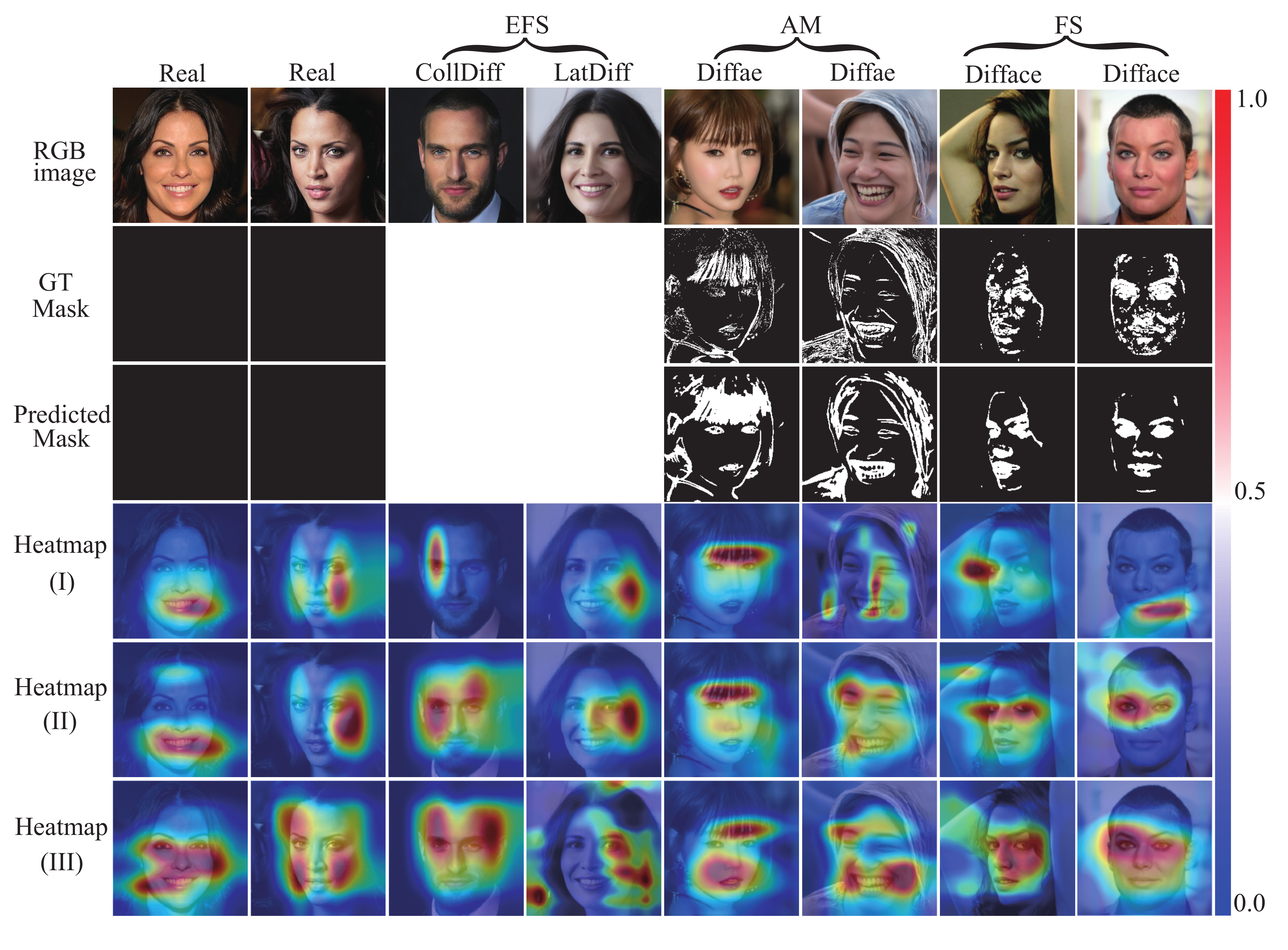} 
	\caption{ The heatmap visualization of various modules in MFVLR. GT mask means the ground truth mask. For the heatmap, the hotter (red color) a position is, the more forgery traces are captured by the models.}\label{figheat}
		\vspace{-1.5em}
\end{figure}
		\vspace{-1.5em}
		
		\subsection{Impacts of various localization mask thresholds } To study the sensitivity of different generators to the threshold of the forgery localization mask, we train our model using DiffFace by varying the threshold from 0.05, 0.10, to 0.15 and evaluate the localization and detection performance of our model on diffusion-based generators, including CollDiff, LatDiff, DDPM, and Diffae. In Fig~\ref{threshold}, the AUC of our model remains stable across different thresholds under the cross-generator evaluation, indicating that threshold selection has little impact on detection. By contrast,  the mIoU of our model is much more sensitive to the threshold, which shows that it mainly influences localization performance of models. In Fig~\ref{threshold} (b), the localization performance generally improves with the increase of the threshold. Among all thresholds, the mIoU attains its maximum at 0.1 and then decreases. The mIoU of our model fluctuates relatively little across various generators under the threshold of 0.1. We believe that the ground truth mask often contains both clear forgery regions and weak noise areas around boundaries or in the background. When the threshold is too low, the noisy regions are retained, which leads to more false positives and poorer mask quality. A large threshold may remove not only noisy responses but also useful manipulated regions, which may cause incomplete masks and reduce localization accuracy. A proper threshold can suppress noise while preserving the main manipulated regions, resulting in better localization performance.
\vspace{-1em}
\section{Visualization}\label{sec6}
{\bfseries\setlength\parindent{0em} Visualization of residual images.}
To intuitively observe the images reconstructed by our model and the corresponding residuals, we visualized the source images from various generative models, the corresponding reconstructed images generated by our model, and the residuals. In Fig~\ref{residual}, we can observe that there are significant distinct between the real facial residual images and fake ones. In addition to the residual inconsistency between authentic and manipulated images, we also noticed that residuals in facial images generated by diffusion are evident visually. However, they are rarely visible in GAN-synthesized ones. Therefore, residuals could be regarded as valuable prior information, to facilitate the DFFDL. We argued that pristine face images tend to contain rich details and texture information, which are often captured and restored during reconstruction. Nevertheless, face images created by diffusion models may not be able to retain the high-frequency information of the original real one, such as texture, which increases the difficulty of reconstruction. By contrast, forgery face images generated by GAN preserve more high-frequency information of the source real images than those created by diffusion \cite{mfclip}, which is helpful for reconstruction, so the residual is relatively small.

\begin{figure}[t!]
	\centering
	\includegraphics[width=\linewidth]{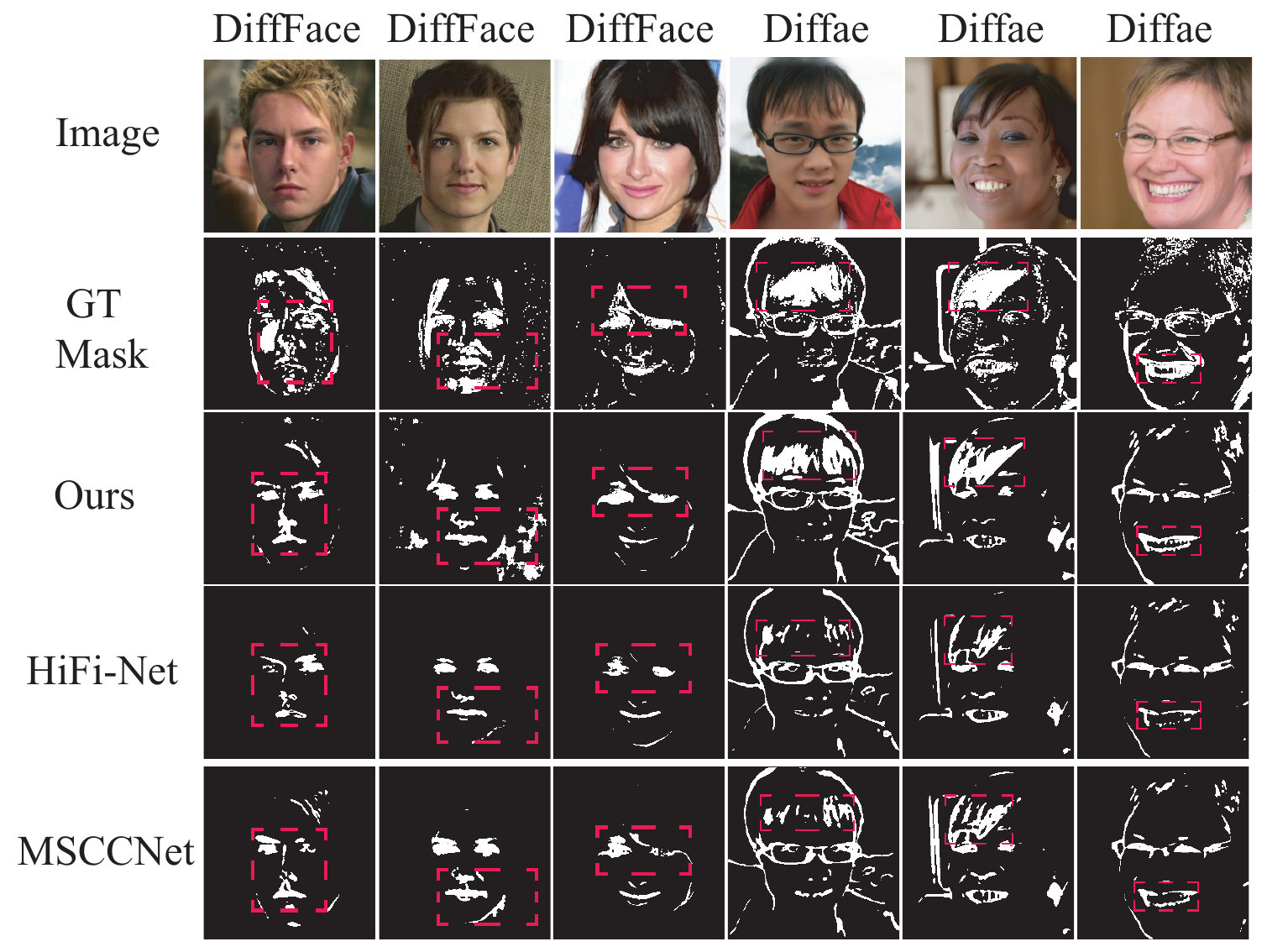} 
	\caption{ The visualization of the manipulation localization of various FFDL models. GT mask means the ground truth mask.  }\label{figmask}
\vspace{-2em}
\end{figure}

{\bfseries\setlength\parindent{0em} Visualization of forgery localization.}
To further study the influence of various modules in MFVLR, we showed the heatmap of various settings in Fig.~\ref{figheat}. Each column displays a face yielded by various generators. The second and third rows show the ground truth mask and predicted mask, respectively. The fourth to sixth rows illustrate the heatmap of three networks: (I) MFVLR only with MVE; (II) MFVLR with MVE and VD; (III) MFVLR consisting of the MVE, VD, and FLT. Compared to (I), the VD module (II) can explore the long-range forgery patterns. FLT (III) further captures more precise and comprehensive manipulated traces than (II), demonstrating that language-guided face forgery representation learning could boost the DFFDL. In detail, in the fifth column, we added the bangs to the real image through the attribute-manipulated diffusion model Diffae, and we observed that our model could, to a large extent, recognize the manipulated regions. Furthermore, compared with the heatmap visualization which tends to be employed to help the detector conduct manipulation localization, FFDL can achieve more precise and pixel-level forgery localization. In addition, FFDL effectively highlights the manipulated regions, and provides clear and detailed visual evidence of facial image tampering. Specifically, it is noticed that although the facial forgery patterns are successfully mined and visualized through heatmaps, the authentic and forged faces are indistinguishable merely based on them. By contrast, the predicted mask could display distinguishable and comprehensive forgery regions between real and fake facial images.

To demonstrate the superiority of our MFVLR method for manipulation localization, we visualized the predicted mask generated by various FFDL models, including HiFi-Net and MSCCNet. As illustrated in Fig~\ref{figmask}, each column displays a fake face generated by various diffusion models. The first and second rows denote the RGB images and ground truth mask, respectively. The second to fifth rows represent the predicted mask yielded by our MFVLR method, HiFi-Net, and MSCCNet, respectively. Compared with the latest FFDL methods, our model captures more comprehensive forgery traces and achieves more accurate face manipulation localization.

\vspace{-1em}
\section{Conclusion}\label{sec7}

We design an innovative MFVLR approach to facilitate the development of generalizable DFFDL. First, we observe that there are significant inconsistencies between the real and fake residuals. Thus, we design the MVE module with the residual encoder to mine the discriminative and fine-grained residual forgery traces, and integrate them with the global facial image forgery features. Second, we devise the vision decoder to reconstruct the image appearance and achieve pixel-level forgery predictions for images, which further boosts the learning of manipulated representations. Finally, we propose an FLT module to extract fine-grained and general class-aware language embeddings, to enhance the learning of face forgery representations through vision-language contrastive learning. Besides, we design a novel plug-and-play VIM module to conduct the vision-guided language representation learning, which could be integrated into any VLM like CLIP, to improve the generalization to face images generated using diffusion models with only a slight increase in the number of parameters.

{\bfseries\setlength\parindent{0em} Limitations.} Although our model has attained the generalizable DFFDL, we may need to improve model generalization to unseen face forgery images synthesized by state-of-the-art diffusion models, and decrease computational costs. In the future, we intend to combine the proposed MFVLR method with the multi-modal large language model, to achieve visual question answering and face manipulation detection and localization. Besides, our method could be applied to detect faces generated by other AIGC methods such as GANs, VAEs or flow models, which may require slight adjustments or refinements in future applications using the specific nature of artifacts produced by different generative models.
\vspace{-1.2em}
\bibliographystyle{IEEEtran}
\bibliography{mfvlr}
\vspace{-2em}
\vspace{-5em}
\begin{IEEEbiography}[{\includegraphics[width=1in,clip,keepaspectratio]{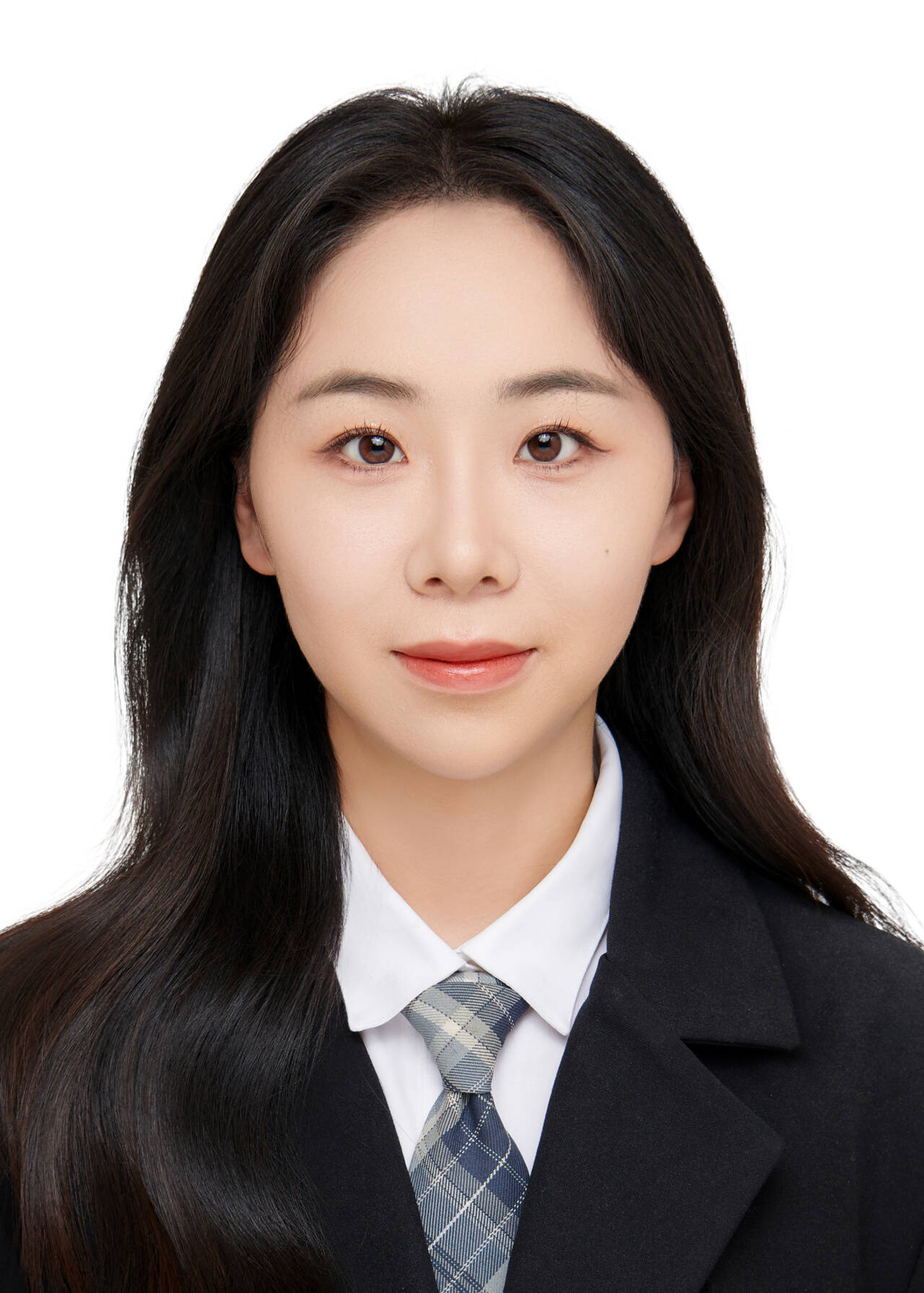}}]{Yaning Zhang} received the double bachelor’s degree in Internet of Things Engineering and English and the M.S. degree in Computer Applied Technology from Qilu University of Technology (Shandong Academy of Sciences), Jinan, China, in 2020 and 2023, respectively, where she is currently pursuing the Ph.D. degree. Her research interests include computer vision, artificial intelligence, multimedia forensics, and face forgery detection.
\end{IEEEbiography}
\vspace{-3.2em}
\begin{IEEEbiography}[{\includegraphics[width=1in,clip,keepaspectratio]{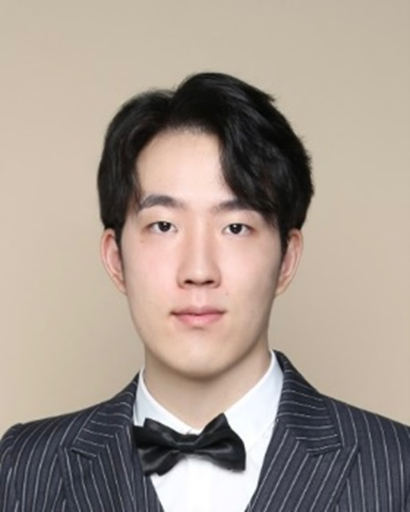}}]{Tianyi Wang} (Member, IEEE) received the double major B.S. degree in Computer Science and Applied and Computational Mathematical Sciences from the University of Washington, Seattle, USA, in 2018. After that, he received the Ph.D. degree in Computer Science, under the supervision of Dr. Kam Pui Chow, from the University of Hong Kong, Hong Kong, in 2023. He is currently a Research Fellow at School of Computing, National University of Singapore, Singapore. His major research interests include multimedia forensics, face forgery detection, misinformation detection, generative artificial intelligence, and computer vision.
\end{IEEEbiography}
\vspace{-3.5em}
\begin{IEEEbiography}[{\includegraphics[width=1in,clip,keepaspectratio]{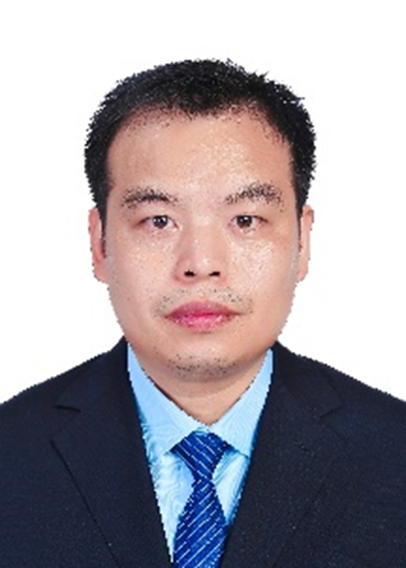}}]{Zan Gao}  (Senior Member, IEEE) received his Ph.D degree from Beijing University of Posts and Telecommunications in 2011. He is currently a full Professor with the Shandong Artificial Intelligence Institute, Qilu University of Technology (Shandong Academy of Sciences). From Sep. 2009 to Sep. 2010, he worded in the School of Computer Science, Carnegie Mellon University, USA. From July 2016 to Jan 2017, he worked in the School of Computing of National University of Singapore. His research interests include artificial intelligence, multimedia analysis and retrieval, and machine learning. He has authored over 100 scientific papers in international conferences and journals including TPAMI, TIP, TNNLS, TMM, TCYBE, CVPR, ACM MM, WWW, SIGIR and AAAI.
\end{IEEEbiography}
\vspace{-3.5em}

\begin{IEEEbiography}[{\includegraphics[width=1in,clip,keepaspectratio]{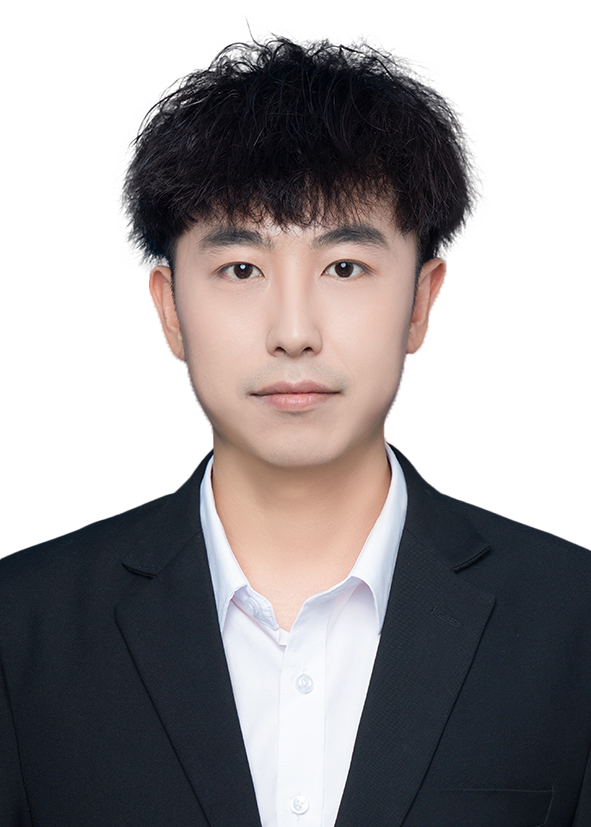}}]{Yibo Zhao} is pursuing his PhD degree in the School of Computer Science and Technology, Tianjin University of Technology. He received his Bachelor's degree from the Tianjin University of Technology in 2018. His research interests include multimedia analysis and retrieval, machine learning.
\end{IEEEbiography}
\vspace{-2.5em}
\begin{IEEEbiography}[{\includegraphics[width=1in,clip,keepaspectratio]{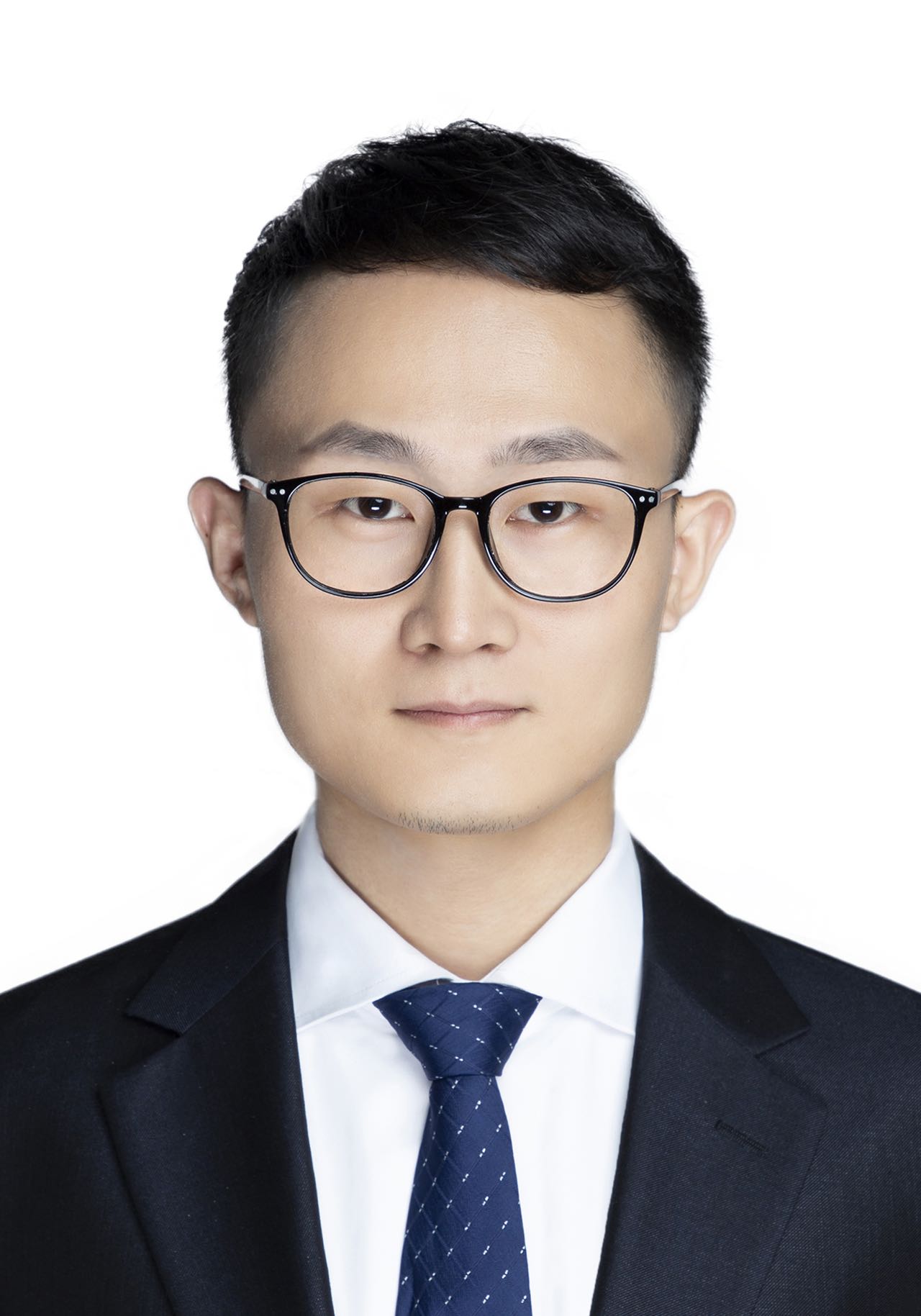}}]{Chunjie Ma} received the Ph.D. degree in Electronic Science and Technology from the Beijing University of Technology in 2023. He is currently a Research Assistant with the Shandong Artificial Intelligence Institute, Qilu University of Technology. His research interests include computer vision, X-ray image prohibited object detection, tamper detection, etc.
\end{IEEEbiography}
\vspace{-3em}
\begin{IEEEbiography}[{\includegraphics[width=1in,clip,keepaspectratio]{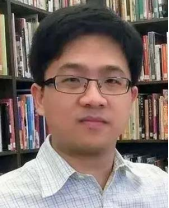}}]{Meng Wang} (Fellow, IEEE)  received the BE and PhD degrees in the Special Class for the Gifted Young and the Department of Electronic Engineering and Information Science from the University of Science and Technology of China (USTC), Hefei, China, respectively. He is a professor at the Hefei University of Technology, China. His current research interests include multimedia content analysis, search, mining, recommendation, and large-scale computing. He received the best paper awards successively from the 17th and 18th ACM International Conference on Multimedia, the best paper award from the 16th International Multimedia Modeling Conference, the best paper award from the 4th International Conference on Internet Multimedia Computing and Service, and the best demo award from the 20th ACM International Conference on Multimedia.
\end{IEEEbiography}

\end{document}